\pdfoutput=1
\documentclass[
pdflatex,sn-basic,Numbered]{sn-jnl}

\usepackage{graphicx}%
\usepackage{multirow}%
\usepackage{amsmath,amssymb,amsfonts}%
\usepackage{amsthm}%
\usepackage{mathrsfs}%
\usepackage[title]{appendix}%
\usepackage{xcolor}%
\usepackage{textcomp}%
\usepackage{manyfoot}%
\usepackage{booktabs}%
\usepackage{listings}%

\usepackage[T1]{fontenc}
\usepackage{lmodern}     %
\usepackage{microtype}
\usepackage{array}       %
\usepackage{longtable}   %
\usepackage{tikz}
\usepackage{fontawesome5}
\usepackage{enumitem}
\usepackage{etoolbox}    %
\usepackage{placeins}    %
\usepackage[nameinlink,capitalize]{cleveref} 

\usetikzlibrary{arrows.meta,positioning,calc,fit,shapes.geometric,shapes.misc}

\tikzset{
  iconnode/.style={
    rectangle,
    rounded corners=2pt,
    draw,
    thick,
    align=center,
    minimum width=3.3cm,
    minimum height=1.2cm,
    inner sep=4pt,
    fill=gray!10
  },
  processbox/.style={
    rectangle,
    rounded corners=2pt,
    draw,
    thick,
    align=center,
    inner sep=5pt,
    fill=gray!10
  },
  flow/.style={-Latex,thick}
}

\newcommand{\toolname}{EpiLens}
\newcommand{\numpaper}{2{,}297}%
\newcommand{\numpapers}{\numpaper}

\AtBeginEnvironment{table}{\let\small\relax}
\AtBeginEnvironment{longtable}{\def\baselinestretch{1}\tablebodyfont\let\small\relax}

\newcommand{\sdpad}{\phantom{0}\phantom{\scriptstyle 0}}
\makeatletter
\@ifundefined{UrlBreaks}{}{\g@addto@macro\UrlBreaks{\do\/\do\-\do\_\do\.}}
\makeatother
\ifdefined\Urlmuskip \Urlmuskip=0mu plus 1mu\relax \fi

\newif\ifshownew\shownewfalse%

\newif\ifshowrev\showrevfalse%

\newbool{publicversion}
\booltrue{publicversion}

\newcommand{\hide}[1]{%
    \ifbool{publicversion}{%
    }{%
        \textcolor{red}{#1}%
    }%
}

\begin{document}

\title{\toolname: Lightweight RAG for Epidemiological Knowledge Acquisition}

\author*[1]{\fnm{Vincenzo} \sur{Perri}}\email{vincenzo.perri@isi.it}

\author[1]{\fnm{Ciro} \sur{Cattuto}}\email{ciro.cattuto@isi.it}

\author[1]{\fnm{Daniela} \sur{Paolotti}}\email{daniela.paolotti@isi.it}

\affil*[1]{\orgname{ISI Foundation}, \orgaddress{\city{Turin}, \country{Italy}}}

\abstract{To prepare for future outbreaks, researchers need to know what studies have been done and what data they use.
Mapping a large literature by hand is slow, while large language models alone can misstate or conflate sources.
We present \toolname{}, a lightweight, open-source retrieval-augmented generation package that applies a fixed retrieval template and output schema to every paper in a corpus and links outputs to source passages.
We build a corpus of \numpapers{} papers from the sources of a multi-pathogen parameter study and evaluate \toolname's paper-classification workflow against 214 manually labelled papers.
Run locally with an open-weight model, so that licensed full texts never leave our hardware, the classifier is comparable to supervised baselines trained on the manual labels, and ahead of them on geography, despite the lack of task-specific training data.
The study's early extraction sheets cite published sources for 327 estimates.
We link each estimate to a cited document or to corpus studies included by a review.
Then, we use the classifier's labels to compare the papers associated with different parameters and pathogens.
\toolname{} labels more papers linked to reproduction numbers than fatality risks as model-based (53\% versus 15\%, compared with 62\% versus 24\% in the manual labels). 
The gap is consistent with the frequent use of transmission models to estimate reproduction numbers and largely persists within pathogens.
}


\maketitle

\section{Introduction}
\label{sec:introduction}

If a new pandemic started tomorrow, we would again face basic questions.
Which data streams are available and where?
How have they been used in past outbreaks?
Which analyses depended on non-traditional data such as mobility traces, economic data, or social media activity, and under what conditions were those data shared?
What kinds of papers produce the parameter estimates we would use, such as reproduction numbers or fatality risks, and how often are they model-based or based on simulated data?
Recent work in digital epidemiology and non-traditional data has shown the potential and the limitations of these sources for crisis response \cite{chafetz_data4covid,bolt2024unconventional,fallatah2024digital}.
Yet, answering such questions still requires the manual screening of hundreds (if not thousands) of papers.

At the same time, large language models (LLMs) have advanced to the point where they can support a wide range of knowledge-intensive tasks, including search-like question answering, summarisation, and information extraction in scientific and technical domains \cite{chen2025ai4research,gokdemir2025hiperrag,annepaka2025large}.
Nevertheless, these models are prone to hallucination: they produce fluent but non-factual statements or conflate sources, which undermines their reliability as stand-alone knowledge repositories \cite{huang2025survey}.
Retrieval-augmented generation (RAG) addresses this issue by coupling LLMs with explicit document retrieval, so that generation is conditioned on retrieved evidence and not just on the model's parameters \cite{lewis2020retrieval,arslan2024survey}.
For epidemiology, where provenance, transparency, and reviewability are essential, such architectures offer a natural way to leverage LLMs' strengths while maintaining stricter epistemic standards.
Finally, a practical obstacle remains in the use of commercial LLMs: some publishers do not allow the full texts they license to be sent to external AI services.
Therefore, analyses have to run where the documents are held, on models the analyst operates.

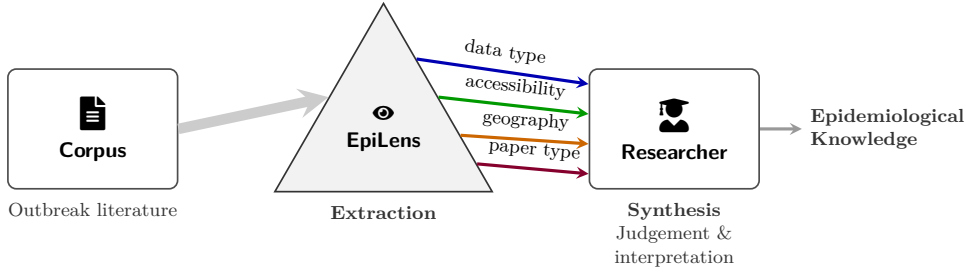
\begin{figure}[t]
  \centering
  \resizebox{\columnwidth}{!}{%

\begin{tikzpicture}[
  font=\sffamily\small,
  >=stealth,
  processbox/.style={
    draw=black!70, thick, rounded corners=4pt, fill=white,
    minimum height=2.0cm, text width=2.6cm, align=center
  },
  flow/.style={->, line width=1.5pt}
]

\node[
  regular polygon,
  regular polygon sides=3,
  draw=black!80,
  thick,
  fill=gray!10,
  align=center,
  minimum size=3.6cm
] (epilens) {\faIcon{eye}\\[0.2em]\normalsize\textbf{\toolname{}}};

\node[processbox, left=2.2cm of epilens] (corpus) {{\Large\faIcon{file-alt}}\\[0.4em]\normalsize\textbf{Corpus}};

\node[processbox, right=2.2cm of epilens] (researcher) {{\Large\faIcon{user-graduate}}\\[0.4em]\normalsize\textbf{Researcher}};

\node[below=0.1cm of corpus, text width=3.2cm, align=center, font=\small, color=black!80] {Outbreak literature}; %

\node[below=0.1cm of epilens, text width=3.2cm, align=center, font=\small, color=black!80] {\textbf{Extraction}}; %

\node[below=0.1cm of researcher, text width=3.2cm, align=center, font=\small, color=black!80] {\textbf{Synthesis}\\Judgement \& interpretation};

\coordinate (prism_in) at ($(epilens.corner 2)!0.5!(epilens.corner 1)$);
\draw[->, line width=5pt, draw=gray!40] (corpus.east) -- node[above, font=\footnotesize\bfseries, text=black!60] {} (prism_in); %

\coordinate (r1) at ($(epilens.corner 1)!0.3!(epilens.corner 3)$);
\coordinate (r2) at ($(epilens.corner 1)!0.5!(epilens.corner 3)$);
\coordinate (r3) at ($(epilens.corner 1)!0.7!(epilens.corner 3)$);
\coordinate (r4) at ($(epilens.corner 1)!0.85!(epilens.corner 3)$);

\coordinate (dest1) at ($(researcher.west) + (0, 0.75)$);
\coordinate (dest2) at ($(researcher.west) + (0, 0.25)$);
\coordinate (dest3) at ($(researcher.west) + (0, -0.25)$);
\coordinate (dest4) at ($(researcher.west) + (0, -0.75)$);

\draw[flow, draw=blue!70!black]   (r1) -- node[above, sloped, font=\small, yshift=1pt] {data type} (dest1);
\draw[flow, draw=green!60!black]  (r2) -- node[above, sloped, font=\small, yshift=1pt] {accessibility} (dest2);
\draw[flow, draw=orange!80!black] (r3) -- node[above, sloped, font=\small, yshift=1pt] {geography} (dest3);
\draw[flow, draw=purple!70!black] (r4) -- node[above, sloped, font=\small, yshift=1pt] {paper type} (dest4);

\draw[->, line width=1.5pt, draw=black!40] (researcher.east) -- ++(0.7,0) node[right, align=left, font=\small\bfseries, text=black!80] {Epidemiological\\Knowledge};

\end{tikzpicture}

} %
  \caption{\textbf{Division of labour between \toolname{} and the researcher.}
    \toolname{} processes outbreak literature into structured, user-configurable dimensions (left), as illustrated by examples in this paper.
    By automating extraction and provenance, the system allows researchers to shift from manual sorting to higher-level interpretation (right).
}
  \label{fig:prism_overview}
\end{figure}

To address these challenges, this work introduces  \toolname{}, a Python package that enables scalable, structured analysis of unstructured outbreak literature by extracting comparable information across a corpus using user-configurable dimensions.
The system enables researchers to map data use, availability, and analysis types efficiently, supporting rapid, evidence-grounded assessment during emergencies.
This paper's aim is not to automate epidemiological judgement.
Instead, we focus on reallocating human effort.
RAG-based tooling can handle the mechanical processing of research literature at scale (search, extraction, structured labelling), while researchers focus on intentionality, domain expertise, and interpretation.
This assistance can be particularly relevant during emergencies, when time pressure makes it difficult to maintain both breadth and depth in literature surveillance.
\Cref{fig:prism_overview} illustrates this principle.

We argue that epidemiology can benefit from \emph{AI-assisted knowledge acquisition}, and we develop \toolname{}, an open-source RAG package designed to support this shift.
Prior work already establishes the strengths of RAG in grounding model outputs, as well as its vulnerabilities to low-quality corpora, retrieval errors, and context misuse \cite{lewis2020retrieval,fan2024survey,arslan2024survey}.
Therefore, we do not seek to re-establish these conclusions, but focus on the design of workflows that apply RAG systematically across a corpus.
\toolname{} provides \emph{structured workflows}: reusable analyses that fix the retrieval template and constrain the model's output with an explicit schema, then apply that same template identically to every paper in a corpus.
This design supports consistent, inspectable corpus-wide analysis: each paper is processed with the same template, outputs are checked against the schema, and retrieved passages are retained for review.
\toolname{} runs these workflows entirely on local hardware with open-weight models, so they can be applied to licensed literature after a few installation steps; hosted models remain an option where licences allow.
We then use a multi-pathogen corpus as a case study to show what becomes possible when such workflows are applied to questions of data use, availability, and geographic coverage, and to the source literature associated with compiled parameter estimates.

\subsection{Contributions}

The contributions of this paper are:

\begin{itemize}
  \item \textbf{A Python package for structured, reusable analysis workflows over epidemiological corpora.}
  \toolname{} turns an analyst-defined retrieval template and output schema into a workflow that applies consistently across papers and adapts to new tasks.
  It supports local or remote inference and links each output to its supporting passages.
  \item \textbf{A reusable worked example: the Paper Classifier.} We develop and evaluate a classification workflow across four label axes with different evidence demands. The evaluation shows the accuracy attainable without task-specific training labels, where body-text retrieval is most useful, and where accuracy remains limited.
  \item \textbf{A corpus-scale case study in digital epidemiology.}
  We provide a multi-axis description of the review-seeded corpus of \numpapers\ papers and a source-level comparison of the papers associated with parameter estimates from the seed study by Ward et al.~\citep{ward2025epidemiology}, by parameter and pathogen.
\end{itemize}

\Cref{sec:background} positions the work, and \Cref{sec:methods} describes \toolname{}, its two workflows, and the evaluation setup.
\Cref{sec:results} evaluates the Paper Classifier on a manually labelled sample, applies it to the full corpus, and links the corpus papers to the parameter estimates of the seed study.
\Cref{sec:discussion} discusses how retrieval-grounded workflows may fit into the knowledge-acquisition toolkit of digital epidemiology and sets out the limitations. The appendices hold the detail behind each step: system components, an extraction workflow, and inference cost (\Cref{app:system_details}); corpus construction and the licences of the evaluation papers (\Cref{app:corpus_pipeline,app:licences}); the labelling protocol (\Cref{app:labeling_protocol}); full metrics, the fully local deployment and error diagnostics (\Cref{app:full_tables}); the baselines and statistical tests (\Cref{app:baselines}); and the linkage of estimates to corpus papers (\Cref{app:provenance}).

\section{Background}
\label{sec:background}

This work sits at the intersection of digital epidemiology, RAG, and automated literature processing.
From digital epidemiology, it takes the motivation to understand how diverse data sources and methods have been used across outbreaks and to turn that understanding into a component of preparedness.
From RAG, it takes the architectural pattern of grounding LLM outputs in retrieved evidence and the associated evaluation concerns.
From literature processing tools, it inherits an emphasis on machine-readable and extensible corpora and references.
By combining these elements in an open, lightweight package focused on epidemiological corpora, \toolname{} aims to make AI-assisted knowledge acquisition a part of digital epidemiology.

\subsection{Digital Epidemiology and Data-Centric Preparedness}

Digital epidemiology developed as a response to the increasing availability of digital data describing human behaviour, mobility, and health, and to the need for more timely and granular information on disease dynamics.
Early work framed digital epidemiology as the use of digitally collected data and tools to study patterns of health and disease, emphasising both its possibilities and the technical and computational challenges it introduces \citep{salathe2012digital}.
Subsequent reviews have documented how digital epidemiology expanded in scope, driven by mobile devices, online services, and novel sensing technologies, and how it began to complement traditional surveillance by covering gaps in timeliness or spatial coverage \citep{park2018digital,fallatah2024digital}.
The COVID-19 pandemic accelerated the use of varied data sources in epidemiology, including structured and unstructured data collected inside and outside health systems \citep{mesquita2025epidemiological}.

Alongside this expansion came a need to better understand not only which data exist but also how they have been used across pathogens, regions, and modelling tasks.
Work on non-traditional data during COVID-19 illustrates this point.
The Data4COVID19 Review \citep{chafetz_data4covid} and related efforts \citep{bolt2024unconventional} catalogued how mobility, financial transactions, social media activity, and other unconventional sources supported decision-making, while further highlighting difficulties around integration, governance, and sustainability.
More broadly, recent reviews argue that digital epidemiology will only fulfil its promise if the biases and shortcomings of new data sources are explicitly recognised and if methods and data practices are adapted accordingly \citep{mesquita2025epidemiological}.
In this context, being able to quickly map which data have been used for which analyses, under which assumptions, becomes a component of preparedness in its own right.

\subsection{Large language models and RAG}

Large language models can summarise, answer questions, and extract information at scale \citep{zhao2023survey}, but are prone to hallucinating plausible but incorrect statements when used as stand-alone knowledge sources \citep{huang2025survey}.
Grounding conclusions in verifiable evidence is a basic requirement across the empirical sciences.
In epidemiology and public health, the cost of failing to do so is especially high.
A fabricated or misattributed finding can propagate into surveillance summaries, resource allocation, and clinical or policy decisions.
During an outbreak, these decisions are made under time pressure, with little opportunity to catch an error before it has consequences.
An unverifiable model output is therefore not simply an inconvenience in this setting, but a direct risk.
RAG tackles this by conditioning generation on passages retrieved from an external corpus rather than using parametric memory alone \citep{lewis2020retrieval}.

RAG has also been applied to the analysis of scientific literature \citep{gokdemir2025hiperrag}.
Recent surveys of RAG architectures detail the integration of retrieval and generation for knowledge-intensive tasks, including open-domain question answering and summarisation, and analyse common design choices and trade-offs in these workflows \citep{fan2024survey,arslan2024survey}.
PaperQA2 and OpenScholar employ RAG to synthesise cited answers to questions concerning scientific literature \citep{skarlinski2024paperqa2,asai2024openscholar}.
In contexts more closely related to this work, LLMs have been evaluated across stages of systematic reviews, from screening to data extraction \citep{gartlehner2024dataextraction,khraisha2024gpt4}, as well as in epidemiological reviews of WHO priority pathogens \citep{padarha2026agentslr}.
These studies investigate whether a model can replicate the decisions made in a given review. 
Agreement varies depending on the specific task and the evaluation metrics used \citep{khraisha2024gpt4}. 
The epidemiological benchmark identifies structured extraction as the primary bottleneck, with current models remaining unreliable without human oversight \citep{padarha2026agentslr}.
SciDaSynth uses RAG to extract structured, source-linked tables from scientific PDFs for interactive validation \citep{wang2025scidasynth}.
Here, we evaluate fixed-template classification across an epidemiological corpus, including a fully local setup, against manual labels and classification baselines.
Ease of use is also a primary design objective. Researchers can begin with a predefined workflow and adapt its retrieval template and output schema without constructing the underlying RAG pipeline.

Finally, RAG has also begun to be adopted in healthcare and biomedicine, where factual accuracy and provenance are critical.
Surveys document applications in medical question answering, guideline querying, and electronic health record summarisation \citep{abo2025survey,gargari2025enhancing,liu2025improving}.
The challenge is adapting these ideas to the specific needs of epidemiology, where questions often concern data availability, study design, and modelling assumptions rather than clinical recommendations alone.

\subsection{Approaches to classifying a literature at scale}
\label{sec:design_space}

Classification is our main example of an \toolname{} workflow.
In this context, it tackles the need to organise a body of literature too large to screen by hand into analytically useful categories, without turning each new research question into a new annotation project.
Methods for this differ along two requirements that are easy to conflate: i) whether the category scheme is specified by the analyst or discovered from the corpus, and ii) whether the method needs manually labelled training examples before it can be applied.
Supervised classifiers learn an analyst-specified scheme, but only from manually labelled examples, typically hundreds per version of the scheme, which must be relabelled whenever the scheme changes.
Active-learning screening tools such as ASReview \citep{vandeschoot2021asreview} reduce this labelling cost by learning from reviewer decisions during screening and prioritising likely relevant records, but they typically learn a single inclusion decision rather than an analyst-defined, multi-axis scheme.
Off-the-shelf zero-shot classifiers apply analyst-specified categories without task-specific training labels; the NLI and prototype baselines tested here score the title and abstract without retrieving passages from the paper's body.
Topic models need no labels, but they discover categories from the corpus and can at most be steered towards a scheme with seed terms.
The Paper Classifier combines an analyst-specified scheme, revisable as text and applied without task-specific training labels, with evidence retrieved from the full document.
Recent general-purpose decision models such as Jev \citep{typesafe2026jev,deusser2026jev} extend the zero-shot family: given analyst-written options and their descriptions, they return a probability for each without task-specific training.
They decide from the text they are given, so they complement evidence retrieval rather than replace it.
We lay out this design space, instantiate each family with standard methods, and summarise the head-to-head comparison in \Cref{sec:baselines_summary}, developing it in full in \Cref{app:baselines}.

\section{Methods}
\label{sec:methods}
This section describes the system design of \toolname{}, our RAG package for epidemiological literature analysis.
The system offloads mechanical text-processing work (retrieval, evidence filtering, extraction, and coarse categorisation) while keeping domain judgement and interpretation with the researcher.
This division of labour is consistent throughout \toolname's design and deployment.
\toolname{} supports a family of \emph{structured workflows}: automated analyses that produce structured, machine-readable outputs based on user-defined templates and schemas.
In this paper, we develop two such workflows: the Paper Classifier, which is the target of our quantitative evaluation, and the Precision Miner, an extraction workflow whose output we illustrate on three corpus papers (\Cref{app:system:workflows}).

\subsection{System overview}
\toolname{} implements a standard three-stage RAG pipeline: ingestion and indexing, retrieval, and generation.
Ingestion parses papers, segments them into chunks, embeds them, and stores them in a searchable index.
Retrieval selects the most relevant chunks for a query or template, optionally filtered by paper metadata.
Generation prompts an LLM to produce an output that is grounded in the retrieved evidence.
We accompany every output with structured provenance that links each extracted fact or label back to the source paper and the exact text snippets used as evidence.
\toolname{} follows a local-first design: the analyst controls parsing, storage, retrieval, and task logic, while embedding and generation can run locally or via external APIs, and the package can also use a remote vector database.
This local layer assembles and structures relevant chunks and enforces schemas, treating the LLM as a stateless computation step that cannot access the corpus or any context beyond what it is given.

\begin{figure}[t]
  \centering
  \definecolor{colCorpus}{HTML}{D6E8F7}   %
\definecolor{colQuery} {HTML}{D5EDD5}   %
\definecolor{colDB}    {HTML}{FDEBD0}   %
\definecolor{colLogic} {HTML}{EDE0F5}   %
\definecolor{colLLM}   {HTML}{FADADD}   %
\definecolor{colOut}   {HTML}{D5F0EA}   %

\definecolor{bdrCorpus}{HTML}{2E86C1}
\definecolor{bdrQuery} {HTML}{1E8449}
\definecolor{bdrDB}    {HTML}{CA6F1E}
\definecolor{bdrLogic} {HTML}{7D3C98}
\definecolor{bdrLLM}   {HTML}{C0392B}
\definecolor{bdrOut}   {HTML}{148F77}

\definecolor{arrowSolid}{HTML}{1C2833}
\definecolor{arrowDash} {HTML}{95A5A6}

\resizebox{\columnwidth}{!}{%
\begin{tikzpicture}[
  font=\normalsize,
  node distance=1.0cm and 1.5cm
]
\tikzset{
  dbnode/.style={
    cylinder,
    shape border rotate=90,
    aspect=0.25,
    draw=bdrDB, thick,
    fill=colDB,
    minimum height=1.35cm,
    minimum width=2.4cm,
    align=center,
    inner sep=4pt
  }
}

\node[processbox, text width=2.4cm,
      fill=colCorpus, draw=bdrCorpus]
  (corpus) {\faIcon{file-alt}\\\textbf{Corpus}\\(papers)};

\node[processbox, text width=2.4cm, below=1.1cm of corpus,
      fill=colQuery, draw=bdrQuery]
  (query) {\faIcon{question-circle}\\\textbf{Query}};

\node[dbnode, right=2.0cm of query]
  (vdb) {\faIcon{database}\\\textbf{Vector DB}\\(index)};

\node[processbox, text width=2.4cm, right=2.2cm of vdb,
      fill=colCorpus, draw=bdrCorpus]
  (chunks) {\faIcon{file-alt}\\\textbf{Relevant\\chunks}};

\node[processbox, text width=2.4cm,
      above=1.2cm of chunks, xshift=-1.4cm,
      fill=colLogic, draw=bdrLogic]
  (tmpl)
    {\faIcon{search}\\\textbf{Templates}};

\node[processbox, text width=2.4cm, right=0.4cm of tmpl,
      fill=colLogic, draw=bdrLogic]
  (instr)
    {\faIcon{clipboard}\\\textbf{Instructions}\\(\& schemas)};

\node[processbox, text width=2.4cm, right=0.6cm of instr,
      fill=colLLM, draw=bdrLLM]
  (llm) {\faIcon{robot}\\\textbf{LLM}};

\node[processbox, text width=2.4cm, below=1.0cm of llm,
      fill=colOut, draw=bdrOut]
  (out) {\faIcon{reply}\\\textbf{Structured output}\\+ provenance};

\draw[flow, dashed, color=arrowDash]
  (corpus.east) -- node[above, xshift=15pt, font=\normalsize] {embedding}
  (vdb.north west);
\draw[flow, dashed, color=arrowDash]
  (vdb) edge[loop above] node[above, font=\normalsize] {indexing} (vdb);

\draw[flow, color=arrowSolid]
  (query.east) -- node[above, font=\normalsize] {embedding} (vdb.west);
\draw[flow, color=arrowSolid]
  (vdb) edge[loop below] node[below, font=\normalsize] {search} (vdb);
\draw[flow, color=arrowSolid]
  (vdb.east) -- node[above, font=\normalsize] {top-$k$ } (chunks.west);
\draw[flow, color=arrowSolid]
  (chunks.east) -- (llm.south west);
\draw[flow, color=arrowSolid]
  (tmpl.south) to (vdb.north east);
\draw[flow, color=arrowSolid]
  (instr.east) to (llm.west);
\draw[flow, color=arrowSolid]
  (llm.south) -- (out.north);

\node[draw, rounded corners=2pt, thick, inner sep=6pt,
      fit=(corpus)(query)(vdb)(chunks)(tmpl)(instr),
      label={[font=\normalsize]above:Local indexing, retrieval, and task logic}]
  (localbox) {};

\node[rounded corners=2pt, inner sep=6pt,
      fit=(llm)(out),
      label={[font=\normalsize, align=center]above:LLM backend\\(local or remote)}]
  (backendbox) {};

\end{tikzpicture}
} %
  \caption{\textbf{RAG workflow used in \toolname{}.}
  Dashed arrows indicate one-time ingestion and indexing of the corpus.
  Solid arrows indicate per-task interrogation (query embedding, search, and generation) with structured outputs linked to source chunks.
  }
  \label{fig:rag_workflow}
\end{figure}
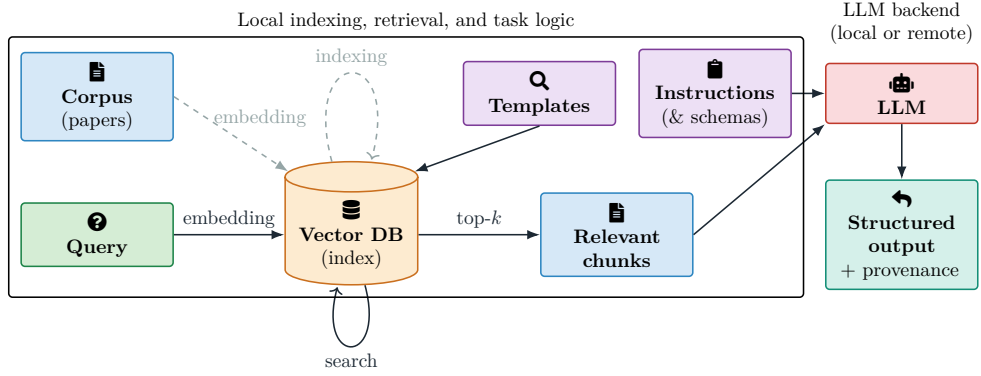

\Cref{fig:rag_workflow} illustrates the ingestion (indexing) and interrogation (querying) branches of this pipeline.
Additional implementation details, including component interfaces and backend choices, are provided in \Cref{app:system_details}.

\toolname{} is deliberately lightweight, by which we mean three things.
First, it depends only on standard, swappable libraries rather than on a general-purpose RAG framework.
The local CLI parses PDFs with \texttt{pdfminer.six} by default, with \texttt{unstructured}~\citep{Unstructured} or GROBID~\citep{grobid} available as alternatives.
Parsed documents are held in memory or, optionally, in a MongoDB database.
Embeddings use a \texttt{sentence-transformers} model~\citep{reimers2019sentence} or a provider embedding API (our main configuration embeds locally with \texttt{BAAI/bge-m3}; the served reference uses Gemini embeddings).
Vector storage uses a local file index, FAISS~\citep{johnson2019billion}, or a Qdrant server~\citep{qdrant}.
Schema definition and validation use \texttt{pydantic}~\citep{colvin2024pydantic}, and generation uses a provider API (OpenAI, Google, Anthropic, OpenRouter) or a local model.
Any layer can be swapped without touching the others.
Second, the core orchestration can run on commodity hardware such as a laptop.
External embedding, generation, or vector-database services are optional deployment choices rather than architectural requirements.
Setup is minimal: local-first defaults run out of the box, with optional upgrades (GROBID parsing, a Qdrant server, a Streamlit UI) for users who want to scale up.
Third, cost is bounded: because each prompt carries a fixed retrieval budget regardless of document length, per-paper token usage is nearly flat in the length of the paper, which keeps corpus-scale application affordable (\Cref{app:cost}).
General RAG toolkits such as LangChain~\citep{langchain}, LlamaIndex~\citep{llamaindex}, and Haystack~\citep{haystack} expose low-level primitives for assembling arbitrary pipelines and maximise flexibility.
\toolname{} offers higher-level building blocks, from ad-hoc retrieval-augmented question answering to structured workflows, trading breadth of customisation for out-of-the-box usability.

\subsection{Structured workflows for systematic corpus characterisation}
Unlike one-off question answering, structured workflows produce structured, machine-readable outputs rather than free-text answers.
Running RAG over each document of a corpus raises two problems that one-off question answering does not.
First, the analysis must be \emph{identical} across papers so that the per-paper results are comparable and can be aggregated.
Second, the outputs must be \emph{structured and validated} so that results can be computed across thousands of papers.
Our workflow resolves both through two artefacts that the user writes once: a \emph{template} and a \emph{schema}.

The \emph{template} fixes how every paper is interrogated, through two parts with distinct jobs.
Its \emph{retrieval queries} fix \emph{what evidence is gathered}: a fixed set of search phrases (or per-category exemplars) is run against each paper, so the same kind of passage is surfaced from every document rather than left to whatever a one-off query happens to return.
Its \emph{prompt instructions} fix \emph{what to do with that evidence}: the task and the criteria for performing it, such as the operational definition of each label, held identical across papers so the same judgement is applied everywhere.
Together, they ensure every paper is asked the same question over the same kind of evidence.

The \emph{schema} fixes the \emph{shape of the answer}: the allowed fields, their types, and their permitted values, expressed as a constrained JSON structure.
It also acts as a validation gate: every generated output is checked against it and rejected if it does not conform.
This is what makes the results usable at scale: each one is guaranteed to be well-formed structured data ready for aggregation, not prose to be re-read and re-parsed by hand.
As shown in \Cref{fig:rag_workflow}, the template governs the query and instruction components of the interrogation branch, and the schema governs the structure of the generator's output, turning an interactive RAG session into a pipeline that can run unattended over an entire corpus.

All workflows share core ingestion, retrieval, generation, validation, and provenance components, but are configured differently for each task.
The Paper Classifier performs category-specific evidence aggregation and returns labels from a fixed vocabulary.
The Precision Miner pools evidence from flat queries, can filter by section, and returns a variable-length list of items.
We evaluate the classifier quantitatively and show the miner's output on three corpus papers in \Cref{app:system:workflows}.

\subsubsection{Paper Classifier}

For evaluation, the Paper Classifier labels each paper by data type, data accessibility, geographical coverage, and paper type (see \Cref{app:labeling_protocol}).
The difficulty is that the category scheme is a moving target: it depends on the question and changes as the question evolves, with new pathogens, a shifting geographic focus, or a revised coding scheme.
This is the setting that the usual tools handle poorly (\Cref{sec:design_space}): supervised classifiers need hundreds of labelled examples per scheme version, and re-labelling whenever it changes, while unsupervised methods cannot be told which scheme to recover and instead surface what the vocabulary signals rather than what the question requires.
A RAG-based classifier sidesteps both: the analyst specifies each category in natural language and revises it by editing text, with no task-specific training labels, and retrieval grounds each decision in the paper's own evidence.
The trade-off becomes an engineering one (retrieval quality, prompt design, model choice) rather than a data-availability one.
Mechanically, the classifier retrieves evidence for each candidate label using exemplar-guided queries and asks the LLM to discriminate among the competing categories, producing a standardised classification that supports a bird's-eye view of the literature.
Component and workflow details are provided in \Cref{app:system_details}.

\subsubsection{Precision Miner}

Some questions need the concrete items themselves rather than a label: the datasets and repositories a field uses, or the sources behind its parameter values.
The Precision Miner is the \emph{extraction} workflow for such questions: it pools evidence from several retrieval queries per paper and returns a variable-length list of items, each with a name, an optional URL or identifier, a short explanation, and the source sentence as provenance.
Its output is a per-paper worklist to verify rather than a distribution to tally.
\Cref{app:system:workflows} describes it and shows its output on three corpus papers.

\subsection{Evaluation setup}
\label{sec:evaluation_setup}
We evaluate the Paper Classifier on four axes: \emph{data type}, \emph{data accessibility}, \emph{geographical coverage}, and \emph{paper type} (\Cref{app:labeling_protocol}).
The first three are set-based (multi-label), whereas \emph{paper type} is a single primary label.
Retrieval has a fixed budget of $k=15$ chunks per label axis: for each axis, the system embeds every template paragraph, retrieves its $k$ nearest chunks from the target paper, pools these candidates across paragraphs, removes duplicates, and keeps the $k$ highest-scoring unique chunks as the evidence context passed to the generator with the task instructions.

We compare three generator backends, with configurations that differ by model, serving mode (local or remote), temperature, and, for the local generator, retrieval index.
The main configuration is fully local: we self-host the open-weight \texttt{qwen2.5:32b-instruct}~\citep{qwen2025qwen25} on our own hardware, embed the corpus on-device with \texttt{BAAI/bge-m3}~\citep{chen2024bgem3}, and evaluate it at $T{=}0$ and $T{=}1$.
For reference, two models are accessed through remotely served inference: Google's proprietary \texttt{gemini-2.5-flash} and \texttt{gemini-2.5-pro}~\citep{comanici2025gemini25}, each evaluated at $T{=}0$ and $T{=}1$.
All remotely served configurations retrieve with \texttt{gemini-embedding-001}~\citep{lee2025geminiembedding}, and a \emph{partially local} configuration combines the local generator with the same Gemini-embedded index.
Because some publishers do not allow their papers to be sent to external AI services, every configuration that uses a remote service is evaluated only on the openly licensed papers of the evaluation sample (licence screen in \Cref{app:licences}).
For each configuration, we repeat the classification task $R{=}5$ times, holding the prompts and post-processing fixed throughout.

\section{Results}
\label{sec:results}

After describing the corpus (\Cref{sec:corpus}), we evaluate the Paper Classifier on a manually labelled sample (\Cref{sec:classification_labeled}) using the settings of \Cref{sec:evaluation_setup}.
The main results use fully local deployment on all $214$ papers: we report accuracy and run-to-run stability, test retrieval's contribution via ablation, and compare the classifier with external baselines.
Served models are reported as a reference on the papers selected by the licence screen of \Cref{app:licences} (\Cref{sec:served_reference}).
We then use the local classifier to describe the full corpus and the papers associated with the seed study's parameter estimates (\Cref{sec:full_corpus_distributions,sec:provenance}).

\subsection{Case-study corpus}
\label{sec:corpus}
We evaluate \toolname{} on a purpose-built corpus of epidemiological literature.
The corpus was seeded from the sources of a recent review of epidemiological parameters for pathogens with pandemic potential \citep{ward2025epidemiology}, which we refer to as the \emph{seed study}.
Its authors shared the data extraction sheets of an early version of the study (close to its first preprint) with us  \citep{ward2025preprint}.

Once identifiers are resolved, the sheets name $138$ published sources: reviews, studies, and other documents (\Cref{app:provenance}).
We retained the directly cited studies and other documents, then added the studies retrieved from the source reviews.
The final corpus contains \numpaper{} papers and excludes the source reviews themselves.
A separate, semi-automated pipeline identified included studies (with LLM assistance for reviews with an open licence), resolved bibliographic records and DOIs, retrieved documents, and ingested them.
Its archived records show what entered the corpus and what remained unresolved or unretrieved, although reconstructing the exact document collection still depends on access to the documents (\Cref{app:corpus_pipeline}).
We retain the links between corpus papers and the seed study’s sources to define the labelled-sample strata (\Cref{sec:classification_labeled}) and to analyse the papers associated with its parameter estimates (\Cref{sec:provenance}).

\subsection{Classification on the manually labelled sample}
\label{sec:classification_labeled}
We evaluate the classifier on $214$ papers sampled from the \numpaper{}-paper corpus.
The sampling groups are the source reviews that contributed papers to the corpus and, as one additional group, the corpus documents cited directly by the seed study (\Cref{sec:corpus}).
From each group of $n$ papers, we randomly select $\max(\lfloor 0.1n\rfloor,1)$ papers.
The one-paper minimum samples the smallest reviews at a higher rate than larger ones.
Only $14$ sampled papers come from reviews with fewer than ten papers.

One annotator labelled each sampled paper on all four axes using a written protocol that defines the labels and sets rules for ambiguous cases (\Cref{app:labeling_protocol}).
We discuss the limitation of relying on a single annotator in \Cref{sec:limitations_future_work}.
The annotations were completed before the annotator saw any model output.
The classifier prompts use the same label definitions, aligning the instructions given to the model with those the annotator followed.

We specified the classifier prompts, retrieval templates, and configuration from the protocol before evaluating on the $214$ papers and made no changes in response to results on this set.
The sample therefore serves as a held-out evaluation set for \toolname{}.
Where a baseline required tuning of NLI wording, topic count, or supervised model configuration, we report the strongest setting observed on these same papers (\Cref{app:baselines:zero_shot,app:baselines:guided_topics,app:baselines:grid}), which favours the baselines.

We report the unweighted mean per-paper Jaccard similarity ($J$) on each axis, with a $95\%$ bootstrap interval over papers.
For each paper, $J$ is the number of labels shared by the prediction and manual annotation divided by the number in either set.
For paper type, which has one label per paper, this is ordinary accuracy.
We report micro- and macro-averaged precision, recall, F\textsubscript{1}, and subset accuracy in \Cref{app:full_tables}.
Because papers from the smallest reviews were sampled at a higher rate, we checked whether unequal sampling affects the reported averages.
Weighting each paper by the inverse of its selection probability changes the mean $J$ of the fully local \toolname{} at $T{=}0$ by at most $0.009$ on any axis (effective sample size $207$).

As set out in \Cref{sec:evaluation_setup}, served models are reported only on the $122$ openly licensed papers (\Cref{app:licences}).
On these papers we score every method, so that comparisons with the served models are paired.
The subset is not a random sample.
It over-represents open-access journals and public-health agencies, and it is somewhat easier on geography and data type, where the majority floor rises from $0.353$ to $0.416$ and from $0.843$ to $0.870$.

\subsubsection{Performance and stability of \toolname{}}
\label{sec:epilens_performance}

We first evaluate the fully local workflow on the $214$ manually labelled papers. \Cref{tab:local_main} reports mean per-paper $J$ with bootstrap intervals, the majority floor, and stability across five runs.

\begin{table}[!htbp]
\centering
\small
\setlength{\tabcolsep}{3pt}
\caption{\textbf{Accuracy and stability of the fully local workflow.}}
\label{tab:local_main}
\begin{tabular}{l c | c c | c c}
\toprule
& \textbf{Floor} & \multicolumn{2}{c|}{\textbf{Jaccard $J$}} & \multicolumn{2}{c}{\textbf{Stable papers (\%)}} \\
\cmidrule(lr){2-2}\cmidrule(lr){3-4}\cmidrule(lr){5-6}
\textbf{Task} & Majority & $T{=}0$ & $T{=}1$ & $T{=}0$ & $T{=}1$ \\
\midrule
Geography          & \textit{0.353} & $0.869_{\,0.82\text{--}0.91}$ & $0.869_{\,0.83\text{--}0.90}$ & 100.0 & 80.4 \\
Data access.       & \textit{0.472} & $0.539_{\,0.48\text{--}0.59}$ & $0.516_{\,0.47\text{--}0.56}$ & 100.0 & 17.8 \\
Data type          & \textit{0.843} & $0.861_{\,0.82\text{--}0.90}$ & $0.866_{\,0.83\text{--}0.90}$ & 100.0 & 79.4 \\
Paper type         & \textit{0.593} & $0.794_{\,0.73\text{--}0.84}$ & $0.775_{\,0.72\text{--}0.82}$ & 100.0 & 87.4 \\
\bottomrule
\end{tabular}
\footnotetext{Agreement with the manual labels on all $214$ evaluation papers for \toolname{} with \texttt{qwen2.5:32b-instruct} served by Ollama on our own hardware and a \texttt{BAAI/bge-m3} index built locally, so no stage of the classification contacts a remote service. $J$ is the mean over $R{=}5$ runs. The subscript is a $95\%$ bootstrap interval over papers (BCa, $B{=}10{,}000$), computed after averaging each paper's score over the runs. \emph{Stable papers} is the share of papers whose complete label set is identical in all five runs. The \emph{Floor} column gives the leave-one-out majority label-set reference. Paper type is a single-label task. The remaining tasks are multi-label.}
\end{table}

At $T{=}0$, the workflow scores above the majority floor on every axis, with the largest margins on \emph{geography} ($0.869$ versus $0.353$) and \emph{paper type} ($0.794$ versus $0.593$).
For each paper, the floor predicts the most frequent complete label set among the other manual annotations. It uses the evaluation sample's label frequencies but no paper content. \toolname{} is not fitted to these labels, although pretraining and prompt wording may favour some categories; the floor cannot isolate the contribution of any such preference to its $J$.

On \emph{data type}, the workflow reaches $J{=}0.861$ against an already high floor of $0.843$, reflecting the prevalence of \emph{Traditional}. Mean $J$ gives limited information about less common labels; \Cref{app:per_label_f1_all_configs} reports per-label F\textsubscript{1} on the $122$ openly licensed papers, where several labels have small support.
\emph{Data accessibility} is the most difficult axis ($0.539$ against $0.472$). Moving from $T{=}0$ to $T{=}1$ changes mean $J$ by at most $0.023$ on any axis.

Mean agreement does not show whether repeated runs assign the same labels to each paper. We define \emph{stability} as the share of papers whose complete predicted label set is identical across all five runs.
At $T{=}0$, every paper is stable on every axis. At $T{=}1$, stability falls to $79$--$87\%$ on geography, data type, and paper type, and to $17.8\%$ on data accessibility.
Mean $J$ nevertheless changes little between runs (standard deviation at most $0.027$; \Cref{app:tab:qwen_metrics_local}): gains and losses across papers largely offset each other, and a changed prediction often alters only part of a label set.

\subsubsection{Served models on the openly licensed papers}
\label{sec:served_reference}

\Cref{tab:qwen_backend_open} compares the local workflow with the two served backends on the $122$ openly licensed papers, all at $T{=}0$.

\begin{table}[!htbp]
\centering
\small
\setlength{\tabcolsep}{3pt}
\caption{\textbf{Self-hosted and served backends of the same workflow.}}
\label{tab:qwen_backend_open}
\begin{tabular}{l c | c c | c c}
\toprule
& \textbf{Floor} & \multicolumn{2}{c}{\textbf{Qwen2.5-32B} (self-hosted)} & \multicolumn{2}{c}{\textbf{Served backends}} \\
\cmidrule(lr){2-2}\cmidrule(lr){3-4}\cmidrule(lr){5-6}
\textbf{Task} & Majority & \emph{fully local} & \emph{partially local} & \texttt{flash} & \texttt{pro} \\
\midrule
Geography           & \textit{0.416} & $0.892_{\pm 0.000}$ & $0.890_{\pm 0.000}$ & $0.890_{\pm 0.009}$ & $0.936_{\pm 0.001}$ \\
Data access.        & \textit{0.474} & $0.536_{\pm 0.000}$ & $0.553_{\pm 0.000}$ & $0.642_{\pm 0.021}\sdpad$ & $0.778_{\pm 0.028}\sdpad$ \\
Data type           & \textit{0.870} & $0.907_{\pm 0.000}$ & $0.929_{\pm 0.000}$ & $0.934_{\pm 0.006}$ & $0.968_{\pm 0.011}\sdpad$ \\
Paper type          & \textit{0.615} & $0.795_{\pm 0.000}$ & $0.811_{\pm 0.000}$ & $0.843_{\pm 0.007}$ & $0.877_{\pm 0.016}\sdpad$ \\
\bottomrule
\end{tabular}
\footnotetext{Sample-averaged Jaccard on the $122$ openly licensed evaluation papers (\Cref{app:licences}), as mean with the standard
deviation across $R{=}5$ repeated runs shown as a subscript. \textbf{Every column is at
$T{=}0$}. Both \textbf{Qwen2.5-32B} columns generate with
\texttt{qwen2.5:32b-instruct} run locally through Ollama on our own hardware, and differ
only in how the retrieval index is embedded: \emph{fully local} embeds with
\texttt{BAAI/bge-m3} on the same machine, so no stage of the pipeline contacts a remote
service, whereas \emph{partially local} reuses the Gemini-embedded index shared with the
served-backend runs. Everything else is held fixed across all columns (dense retrieval,
\texttt{workflow\_top\_k}${=}15$, grobid parses, identical prompts, schema-constrained
decoding). The \emph{Floor} column is the deterministic majority-class baseline.
The two Gemini backends, \texttt{flash} and
\texttt{pro}, are the served reference. Both
self-hosted configurations return \emph{identical} predictions on all four axes across all
five repeats, so their SD is exactly zero (\Cref{app:qwen_local}).}
\end{table}

\texttt{gemini-2.5-pro} has the highest mean on every axis.
On these papers the local workflow matches \texttt{gemini-2.5-flash} on \emph{geography} ($0.892$ against $0.890$), is lower on \emph{data type} ($0.907$ against $0.934$) and \emph{paper type} ($0.795$ against $0.843$--$0.877$), and furthest behind on \emph{data accessibility} ($0.536$, against $0.642$ for Flash and $0.778$ for Pro), the axis that benefits most from a stronger model.
Retrieving from a Gemini-embedded index instead of the local one changes the local workflow's $J$ by at most $0.022$.

Unlike the local workflow, both served backends change some labels even at $T{=}0$ (\Cref{tab:stability_all_configs_open}).
Flash and Pro are stable on $84$--$97\%$ of papers on geography, data type, and paper type, but on $61$--$66\%$ on data accessibility.
Variation under nominally deterministic settings has also been observed elsewhere \citep{atil2024nondeterminism}.
At $T{=}0$ decoding is nominally greedy, picking the most probable token, but the probabilities are recomputed for every request, and a floating-point sum depends on the order in which its terms are added \citep{pytorch_numerical_accuracy,yuan2025numerical}.
On a shared inference server, that order changes with the batch size a request is grouped into, which depends on other users' traffic.
A small difference can then reverse a near tie between two tokens, and every later token follows from the changed prefix \citep{he2025nondeterminism}.
With kernels whose results do not depend on the batch, \citet{he2025nondeterminism} obtained identical completions in $1{,}000$ repeats, at some cost in speed \citep[see also][]{gond2026llm42}.
We cannot observe how the served models batch our requests, so this is the likely but untested explanation for their label changes.

\subsubsection{Ablation of body-text access and selection}
\label{sec:ablation}

Aggregate performance does not reveal which components of the workflow are driving its gains.
Therefore, we compare conditions that, within each generator, use the same model, prompt, and post-processing and differ only in the evidence supplied:

\begin{enumerate}[leftmargin=*, label=\arabic*.]

    \item \textbf{No body text} (\texttt{metadata\_llm}):
    The model receives only the paper's title and abstract.

    \item \textbf{Random body retrieval} (\texttt{random\_chunk\_llm}):
    the prompt additionally receives $k$ body-text chunks sampled uniformly
    at random from the paper.

    \item \textbf{Targeted retrieval} (full \toolname{}):
    The same number of body-text chunks is selected by semantic similarity
    to the template paragraphs.

    \item \textbf{Whole body} (\texttt{full\_body\_llm}):
    The prompt receives every body-text chunk of the paper, in document order,
    with no selection.

\end{enumerate}

The contrast between step~1 and step~2 measures the contribution of access to body text, while the contrast between step~2 and step~3 measures the additional contribution of semantic targeting at a fixed evidence budget.
Step~4 tests whether selecting passages improves on supplying the whole parsed body.

\begin{figure}[!htbp]
    \centering
    \includegraphics[width=\linewidth]{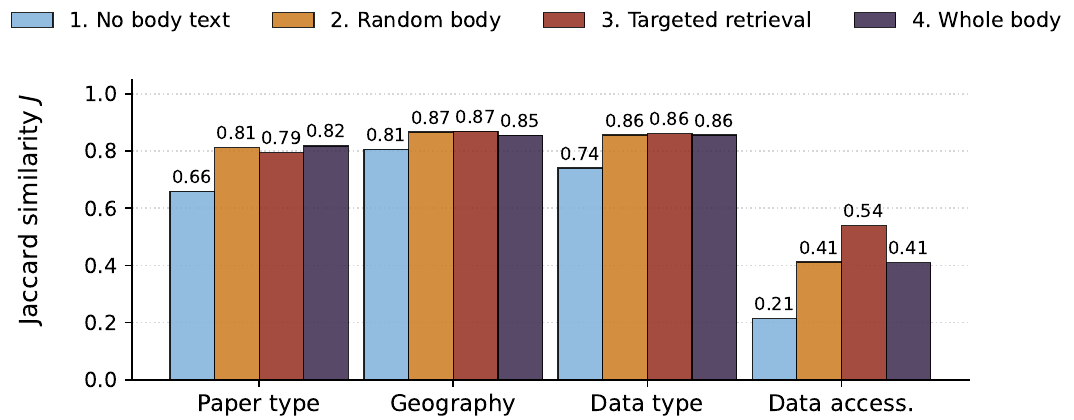}

    \caption{\textbf{Ablation results for the fully local workflow.}
    \texttt{qwen2.5:32b-instruct} at $T{=}0$ on all $214$ evaluation papers, with the model, prompt, and post-processing fixed.
    The four conditions are title and abstract only, title and abstract plus $k{=}15$ randomly selected body-text chunks, title and abstract plus $k{=}15$ semantically retrieved chunks, and title, abstract, and the whole parsed body.
    Comparing conditions 1 and 2 shows the effect of adding body text.
    Comparing conditions 2 and 3 shows the effect of targeted selection with the same number of chunks.
    \Cref{fig:ablation_open} repeats the ablation with \texttt{gemini-2.5-flash} on the openly licensed papers.}
    \label{fig:ablation}
\end{figure}

The largest and most consistent gain comes from access to body text, even when the passages are selected at random (\Cref{fig:ablation}).
Moving from title and abstract alone to random body-text retrieval increases $J$ by ${+}0.06$ on \emph{geography}, ${+}0.20$ on \emph{data accessibility}, ${+}0.12$ on \emph{data type}, and ${+}0.15$ on \emph{paper type}.
Papers in this corpus have $33$ chunks on average (\Cref{app:cost}), so the $15$ randomly selected chunks already expose the model to much of the body.
Nevertheless, the starting point differs by task.
Title and abstract alone already give $J{=}0.807$ on \emph{geography}, $0.741$ on \emph{data type}, and $0.659$ on \emph{paper type}, but only $0.213$ on \emph{data accessibility}, indicating that the latter depends much more strongly on information beyond the abstract.

Semantic targeting contributes additional value more selectively.
Its largest effect is on \emph{data accessibility}, where replacing random chunks with targeted ones raises $J$ by ${+}0.13$ ($0.412{\to}0.539$).
On the other three axes, the effect is small or absent.
In the paired paper-level resampling check of \Cref{app:baselines:statistics}, the gain from adding random body-text chunks meets the within-task Holm-adjusted threshold on every axis, and the gain from targeted chunks meets it only on data accessibility.
Supplying the whole parsed body instead of the $15$ targeted chunks brings no clear gain on any axis, despite about twice the input tokens (\Cref{app:cost}).
On \emph{data accessibility}, $J$ falls from $0.539$ with targeted chunks to $0.410$ with the whole body (pointwise $95\%$ BCa interval for the difference, $0.05$--$0.20$), the only one of the four comparisons that meets the threshold.
Selecting passages can therefore help on this task even when the whole parsed body fits in the model's context window.
With \texttt{gemini-2.5-flash} on the openly licensed papers, the decomposition is similar: body text helps most on data accessibility, paper type, and data type, and targeting mainly on data accessibility (\Cref{fig:ablation_open}).

\subsubsection{Comparison with classification baselines}
\label{sec:baselines_summary}

We compare \toolname{} with alternative workflows that a practitioner might use to organise a literature corpus.
Standard supervised classifiers pair a paper representation with a classifier head trained on labelled examples.
Here, we use TF-IDF\,+\,linear SVM for a lexical setup and frozen ModernBERT with an MLP head for a long-context setup.
Zero-shot NLI instead uses a pretrained language encoder to score how strongly a paper supports each verbalised category, without task-specific fitting.
Guided and semi-supervised BERTopic discover corpus-specific clusters rather than applying a fixed taxonomy, so their topics require subsequent interpretation and alignment.
SetFit provides a distinct supervised route, contrastively fine-tuning an MPNet encoder on pairs of labelled examples before fitting a classification head.
We also include two prototype approaches, which encode each category description (or, for geography, its name) in sparse TF-IDF or dense SPECTER space and assign papers by similarity, with label-free and threshold-calibrated variants.
Finally, as in the preceding results, a majority prior that always predicts the most prevalent label set provides a no-text reference floor.
The abstract-only \texttt{metadata\_llm} row is kept only as the scope-matched \toolname{} condition analysed in the ablation above.
All baseline rows in \Cref{tab:baseline_comparison_local} run on our own hardware.
The methods also differ in practical text scope, recorded by \emph{Body?} in \Cref{tab:baseline_comparison_local}.
The prototype, NLI, and SetFit baselines use title and abstract, whereas BERTopic, ModernBERT, and full \toolname{} use parsed body text.
The wider supervised grid tests both scopes wherever possible.
Body text is not uniformly beneficial: some representation--head combinations decline slightly, perhaps because additional text dilutes task-relevant cues already concentrated in the abstract, although the best full-body configuration raises the supervised ceiling on every axis.

Creating a sufficiently broad labelled set is a substantial additional step for a large or diverse corpus, and a changed coding scheme generally requires renewed labelling and model fitting.
All rows are scored on the same papers with the same metric.
Where a baseline needs tuning, we report its strongest setting, which favours the baselines because \toolname{} was fixed before evaluation (\Cref{sec:classification_labeled}).
The notes to \Cref{tab:baseline_comparison_local} define its columns and its upper-bound rows, and \Cref{app:baselines} gives the configurations, tuning, complete supervised grid, and statistical comparisons.

\begin{table}[!htbp]
\centering
\small
\setlength{\tabcolsep}{4pt}
\caption{\textbf{Performance and requirements of the baseline approaches.}}
\label{tab:baseline_comparison_local}
\begin{tabular}{@{}>{\raggedright\arraybackslash}p{3.9cm}@{\hskip 0.5em}cc@{\hskip 0.5em}cccc@{}}
\toprule
\textbf{Method} & \multicolumn{2}{c}{\textbf{Requirements}} & \multicolumn{4}{c}{\textbf{Jaccard by task}} \\
\cmidrule(lr){2-3}\cmidrule(lr){4-7}
 & \textbf{Body?} & \begin{tabular}[b]{@{}c@{}}\textbf{Task}\\\textbf{sup.}\end{tabular} & \begin{tabular}[b]{@{}c@{}}\textbf{Paper}\\\textbf{type}\end{tabular} & \begin{tabular}[b]{@{}c@{}}\textbf{Geo-}\\\textbf{graphy}\end{tabular} & \begin{tabular}[b]{@{}c@{}}\textbf{Data}\\\textbf{type}\end{tabular} & \begin{tabular}[b]{@{}c@{}}\textbf{Data}\\\textbf{access.}\end{tabular} \\
\midrule
    \addlinespace[2pt]\multicolumn{7}{l}{\textbf{Reference}} \\
    \quad Majority prior (floor) & No & Weak & 0.593 & 0.353 & 0.843 & 0.472 \\
    \midrule
    \addlinespace[2pt]\multicolumn{7}{l}{\textbf{Zero-shot (no training labels)}} \\
    \quad Prototype (TF-IDF) & No & None & 0.407 & 0.283 & 0.315 & 0.217 \\
    \quad Prototype (SPECTER) & No & None & 0.126 & 0.165 & 0.515 & 0.360 \\
    \quad Zero-shot NLI (DeBERTa) & No & None & 0.565 & 0.709 & 0.018 & 0.354 \\
    \midrule
    \addlinespace[2pt]\multicolumn{7}{l}{\textbf{Topic models (corpus-derived, label-guided)}} \\
    \quad Guided BERTopic & Yes & Weak & 0.621 & 0.576 & 0.843 & 0.472 \\
    \quad Semi-supervised BERTopic & Yes & Weak & 0.626 & 0.581 & 0.843 & 0.477 \\
    \midrule
    \addlinespace[2pt]\multicolumn{7}{l}{\textbf{Supervised (title + abstract)}} \\
    \quad TF-IDF + Linear SVM & No & Manual & 0.713 & 0.592 & 0.854 & 0.533 \\
    \quad SetFit (MPNet) & No & Manual & 0.734 & 0.599 & 0.864 & 0.574 \\
    \quad Prototype (TF-IDF) & No & Manual & 0.407 & 0.339 & 0.315 & 0.247 \\
    \quad Prototype (SPECTER) & No & Manual & 0.126 & 0.179 & 0.672 & 0.363 \\
    \midrule
    \addlinespace[2pt]\multicolumn{7}{l}{\textbf{Supervised (full body)}} \\
    \quad ModernBERT (MLP head) & Yes & Manual & 0.785 & 0.652 & 0.864 & 0.577 \\
    \quad \emph{Best supervised} (ceiling) & Yes & Manual & 0.785 & 0.771 & 0.898 & 0.619 \\
    \midrule
    \addlinespace[2pt]\multicolumn{7}{l}{\textbf{LLM, self-hosted (abstract-only ablation and full \toolname{})}} \\
    \quad Abstract only (\texttt{metadata\_llm}) & No & None & 0.659 & 0.807 & 0.741 & 0.213 \\
    \addlinespace[4pt]
    \quad \textbf{\toolname{}} (qwen2.5:32b-instruct) & Yes & None & \begin{tabular}[t]{@{}c@{}}$0.794$\\[2pt]{\footnotesize 0.73--0.84}\end{tabular} & \begin{tabular}[t]{@{}c@{}}$0.869$\\[2pt]{\footnotesize 0.82--0.91}\end{tabular} & \begin{tabular}[t]{@{}c@{}}$0.861$\\[2pt]{\footnotesize 0.82--0.90}\end{tabular} & \begin{tabular}[t]{@{}c@{}}$0.539$\\[2pt]{\footnotesize 0.48--0.59}\end{tabular} \\
\bottomrule
\end{tabular}
\footnotetext{Sample-averaged Jaccard similarity $J$ on the fixed $214$-paper evaluation set.
\textbf{Body?} indicates whether parsed body text is required. \emph{No} generally means title and abstract,
while the majority floor uses no text. \textbf{Task sup.} records the task-specific input used to construct or
fit the method: \emph{None} uses no labelled papers, \emph{Weak} uses a class prior, seeds, or partial paper
labels, and \emph{Manual} uses a fully manually labelled training set. All scores use the same manual labels.
Prototype rows contrast uncalibrated similarity with threshold calibration. BERTopic uses the best topic
count and oracle cluster-to-label alignment. \emph{Best supervised} is the per-task maximum over the
supervised grid. Both are favourable descriptive upper bounds rather than single deployable configurations.
For \toolname{}, the mean is over $R{=}5$ runs at $T{=}0$, with its $95\%$ bootstrap interval over papers beneath it, as in \Cref{tab:local_main}. Other rows are
deterministic evaluations or cross-validated estimates. Scores are rounded to three decimals. Paired tests
are reported in \Cref{tab:significance_local}. Full method specifications are given in \Cref{app:baselines}.}
\end{table}

In this comparison, which is deliberately favourable to the baselines, the fully local workflow, lacking task-specific training labels, has the highest point estimate on \emph{geography} and \emph{paper type}, including the retrospective supervised ceiling.
On \emph{data type} it matches SetFit and ModernBERT, and on \emph{data accessibility} it is slightly below them ($0.539$ against $0.574$ and $0.577$); on both axes the supervised ceiling is higher.
In paired tests, its advantage on geography meets the within-task Holm-adjusted threshold against every baseline, and on paper type against TF-IDF\,+\,SVM; on data type and data accessibility the differences from the supervised baselines are small and not significant (\Cref{tab:significance_local}).
On the $122$ openly licensed papers, Pro has the highest point estimate on every axis, widening the margin over the baselines, and Flash is above every baseline (\Cref{tab:baseline_comparison_open}).
With this smaller sample, Flash's advantage over the supervised baselines meets the threshold against all of them on geography, against TF-IDF\,+\,SVM on data accessibility, on paper type only against the leave-one-out variant of TF-IDF\,+\,SVM, and against none on data type (\Cref{tab:significance_open}).

\subsection{Corpus-level analysis}
\label{sec:full_corpus_distributions}

As mentioned in \Cref{sec:corpus}, we assembled the corpus from sources cited for epidemiological parameter estimates across several pathogens.
After evaluating the classifier on a manually labelled sample, we turn to the literature those sources gathered.
The overall label distribution across \numpaper{} papers provides a reference for comparing papers by parameter and pathogen.
Both analyses use the fully local workflow at $T{=}0$, so that no paper leaves our hardware.
We report corpus shares only for paper type, geography, and data type, where it agrees best with the manual labels (\Cref{sec:epilens_performance}).

Across the full corpus, the model labels $69.4\%$ of papers as empirical and $18.9\%$ as inference, with $44.2\%$ covering Asia and $33.8\%$ Africa.
It labels $94.5\%$ as using traditional data and $11.0\%$ as presenting simulated data.
On the labelled sample, the same model labels more papers empirical than the annotator ($70\%$ against $59\%$) and detects model-based and simulated-data papers less often (sensitivity $0.65$ and $0.68$; \Cref{app:provenance}), so these shares probably understate model-based work.
Detailed distributions appear in \Cref{fig:corpus_scale_distributions_local}.

\subsection{Papers associated with parameter estimates}
\label{sec:provenance}

\begin{figure}[!htbp]
  \centering
  \includegraphics[width=0.72\textwidth]{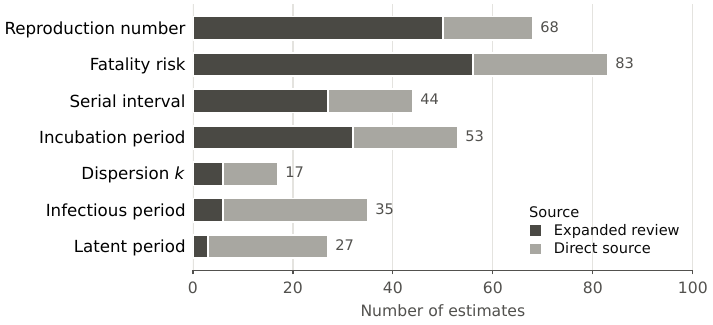}
  \caption{\textbf{Sources of parameter estimates.}
  Estimates of each parameter, by source: a review whose included studies are in our corpus, or a single study or other document in our corpus.
  Parameters are ordered by the share of their estimates taken from reviews.}
  \label{fig:provenance}
\end{figure}

We now turn from the full corpus to the papers associated with the seed study's parameter estimates.
We make these comparisons using the source links retained during corpus construction (\Cref{sec:corpus}).
The privately shared early extraction sheets contain $346$ estimate rows; our source-link audit finds that $327$ cite a published source (\Cref{app:provenance}).
We exclude the other $19$ from this analysis.
For each of the $327$, we link the estimate to its cited corpus document or, if it cites a review, to the retrieved studies included in that review.
We refer to these as the \emph{papers behind} an estimate.
When an estimate cites a review, the link does not identify which of the retrieved included studies informed the particular value.
We describe those studies as a group associated with the estimate, without assigning the value to any one study (\Cref{app:provenance}).
Of those estimates, $180$ cite reviews and $147$ cite single studies or other documents (\Cref{fig:provenance}).

\subsubsection{Parameter}
\begin{figure}[!htbp]
  \centering
  \includegraphics[width=\textwidth]{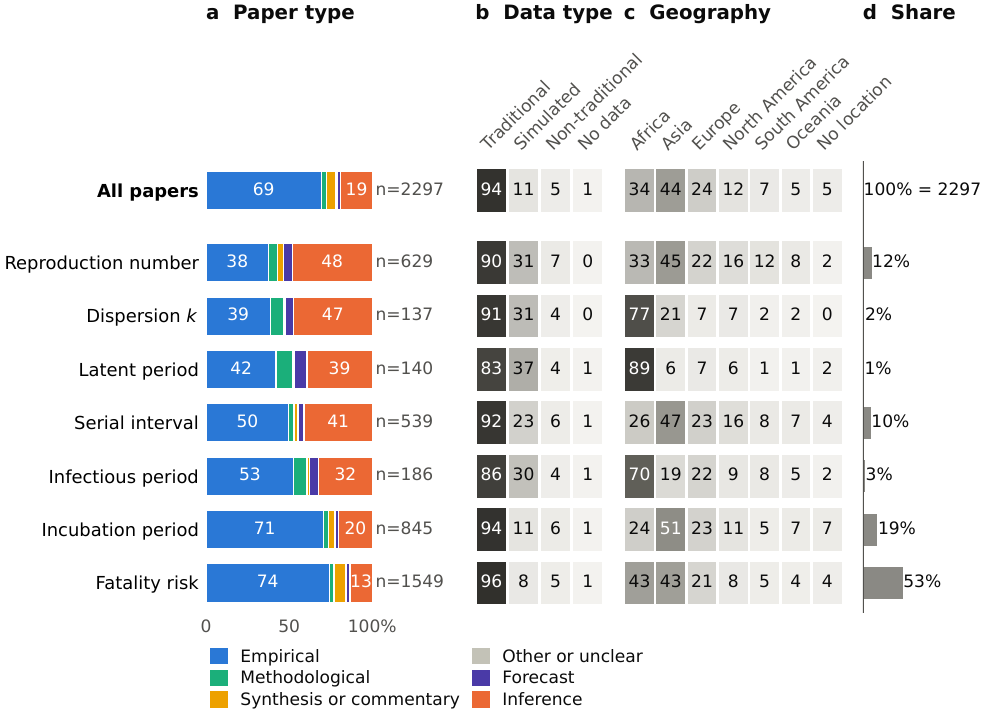}
  \caption{\textbf{Profiles of associated papers by parameter.}
  The first row describes all \numpaper{} corpus papers, and each other row the papers behind the estimates of one parameter, as labelled by the fully local \toolname{} (model outputs, not estimates).
  Rows are ordered by the share of model-based papers, and $n$ is the number of papers.
  (a) Percentage of the papers with each paper type.
  Each paper has one, and model-based papers are those labelled forecast or inference.
  (b, c) Percentage of papers by data type and continent.
  A paper can have several.
  ``No data'' means no empirical data, and ``no location'' means irrelevant or unclear.
  (d) Fractional allocation of the corpus: each paper has total weight one, divided equally among the pathogen--parameter pairs whose estimates it lies behind, so that the row weights sum to $100\%$.}
  \label{fig:provenance_parameters}
\end{figure}

\Cref{fig:provenance_parameters} compares the profiles of papers associated with each parameter.
Its first row shows all \numpaper{} corpus papers for reference.
The profiles use the same paper-type, data-type, and geographic labels as the corpus overview, from the fully local workflow.
For this comparison, we group papers labelled inference or forecast under \emph{model-based}, including statistical distribution fitting and transmission modelling, and use the synthetic-data label for \emph{simulated data} (\Cref{app:labeling_protocol}).
The model-based and simulated-data shares are tabulated in \Cref{tab:underlying_papers}.

We focus first on reproduction numbers and fatality risks, the parameters with the most estimates ($68$ and $83$).
In the manually labelled sample (\Cref{sec:classification_labeled}), $62\%$ of the $61$ papers behind reproduction numbers are model-based ($49\%$ inference, $13\%$ forecast), against $24\%$ of the $144$ behind fatality risks ($20\%$ inference, $4\%$ forecast).
A higher model-based share for reproduction numbers is plausible given the role of model fitting in their estimation.
Across all associated corpus papers, \toolname{}'s outputs show the same ordering: $53\%$ of the $629$ papers behind reproduction numbers are model-based ($48\%$ inference, $5\%$ forecast), against $15\%$ of the $1{,}549$ behind fatality risks ($13\%$ inference, $2\%$ forecast).

The papers behind each parameter span different pathogens, so the gap could simply reflect the mix of pathogens behind each parameter.
To check this, we compare the two sets within pathogens: within SARS-CoV-2, MERS-CoV, Ebola, and Lassa, the reproduction-number papers are still more often model-based, by $7$--$28$ percentage points, whereas for mpox the two sets hardly differ.
Standardised to the pathogen mix of either parameter, the gap is smaller ($16$--$18$ percentage points, against $38$ overall), so it is partly, but not only, a matter of which pathogens lie behind each parameter (\Cref{tab:paper_profile_comparisons}).

\subsubsection{Pathogen}

\begin{figure}[!htbp]
  \centering
  \includegraphics[width=\textwidth]{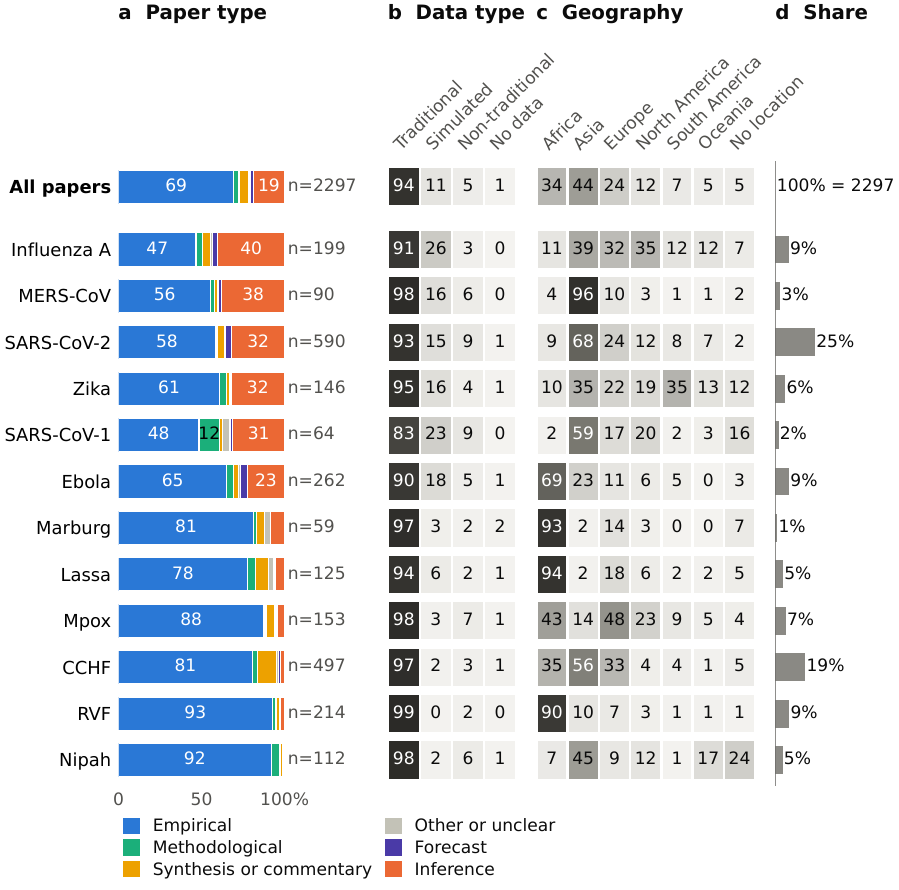}
  \caption{\textbf{Papers linked to cited sources by pathogen.}
  The first row describes all \numpaper{} corpus papers.
  Each other row describes papers linked to sources cited for that pathogen across all parameters in the seed study.
  The panels show the same measures as in \Cref{fig:provenance_parameters}.
  Pathogens with at least $10$ associated papers are shown.
  Influenza A subtypes are merged here and listed separately in \Cref{tab:underlying_papers}.
  CCHF: Crimean--Congo haemorrhagic fever.
  RVF: Rift Valley fever.}
  \label{fig:provenance_pathogens}
\end{figure}

We now compare the corpus papers linked to the seed study's cited sources across pathogens.
\Cref{fig:provenance_pathogens} shows their paper type, data type, and geographic profiles, pooling papers across parameters for each pathogen.

In the model's labels, the source sets fall into two groups. Six (influenza A, SARS-CoV-1, SARS-CoV-2, MERS-CoV, Zika, and Ebola) contain larger shares of inference papers and of papers using simulated data than the corpus as a whole, with influenza A and MERS-CoV highest on inference ($40\%$ and $38\%$, against $19\%$). The other six (Marburg, Lassa, mpox, CCHF, RVF, and Nipah) are predominantly empirical ($78$--$93\%$ of papers).
In the geography panel, nearly all papers linked to MERS-CoV cover Asia, and nearly all those linked to Marburg and RVF cover Africa.
Because each row pools the papers behind all of a pathogen's estimates, these profiles also reflect which parameters the seed study estimated for each pathogen (\Cref{app:provenance}).

\section{Discussion}
\label{sec:discussion}

In this work, we present \toolname{}, a RAG package for organising epidemiological literature at a scale that manual review cannot reach.
It treats the generator model as a linguistic engine over retrieved passages and retains the intermediate artefacts behind each output (the retrieved passages, the prompt, and the schema-validated label or extraction with a back-link to its source chunks), so that any aggregate result can be traced to the evidence that produced it rather than discarded once the answer is returned.

Our evaluation shows both the effectiveness of the retrieval-augmented workflow and how its performance depends on the target analytical task.
In the broad corpus-organising axes of \emph{geography}, \emph{data type}, and \emph{paper type}, the full RAG workflow, run entirely on local hardware, achieves high agreement with the manual annotations (on data type, little above the majority baseline) and is comparable to supervised classifiers without requiring task-specific training labels, ahead of them on geography (\Cref{app:baselines}).
Performance is lower on \emph{data accessibility}, which emerges consistently as the most difficult of the four axes: the local workflow agrees least with the manual labels there, and the served models gain most on this axis.

The ablation analysis clarifies where these gains come from (\Cref{sec:ablation}).
Across both the self-hosted \texttt{qwen2.5:32b-instruct} and, on the openly licensed papers, the served \texttt{gemini-2.5-flash}, access to body text supplies the largest and most consistent improvement over title- and abstract-only classification.
Semantic targeting adds a further, task-dependent gain: it matters most for \emph{data accessibility} and little for \emph{geography}.
This pattern aligns with access-related evidence being comparatively sparse and localised, while place-related cues are more widespread throughout a paper.
Supplying the whole body showed no clear accuracy advantage over selected passages despite about twice the input tokens.
On \emph{data accessibility}, it diminished accuracy for the self-hosted generator.
The pattern recurs across two generators, while the overall performance level remains model-dependent.

More broadly, the workflow fills a different role from standard supervised classification.
An analyst can define or revise a coding scheme in natural language, apply it directly to full document evidence, and retain provenance for each classification.
This makes the approach especially useful when taxonomies are evolving, labelled examples are limited, or the relevant evidence is not consistently available in titles and abstracts.
The absence of training labels also makes the workflow complementary to established approaches: an annotation-free pass can map a corpus, identify difficult cases for expert inspection, or provide candidate labels for subsequent supervised modelling or manual curation.
Running locally, in turn, makes such analyses possible on licensed literature that cannot be sent to external AI services (\Cref{sec:evaluation_setup}), at the accuracy reported here.
Once an axis has been validated on a labelled sample, the same specification can then be applied across the full corpus to produce structured distributions while retaining a path back to the papers and passages that support them.
Applied to our corpus, the labels describe both the literature as a whole and the paper sets linked to the seed study's parameter estimates (\Cref{sec:full_corpus_distributions,sec:provenance}).
In this setting, domain expertise shifts from manually coding every document toward defining operational categories, validating their reliability, and auditing the evidence behind decisions.

\subsection{Significance for Digital Epidemiology and Preparedness}

Pandemic preparedness depends in part on having a transparent, up-to-date map of what has been studied, on which data, and in which settings.
\toolname{} facilitates maintaining such a map, while retaining the link between high-level summaries and the underlying documents that support them.
A recurring concern in digital epidemiology is not only whether data exist, but how they have been used across pathogens, regions, and modelling tasks.
The corpus-level analyses outlined in \Cref{sec:full_corpus_distributions} begin to make this usage legible in a structured, updateable form.
The four dimensions describe complementary features of the literature.
Data type records the kinds of evidence used.
Data accessibility records stated access routes or whether printed values permit limited reconstruction.
Geographical coverage locates the evidence, and paper type records the principal contribution.
Together, they support retrospective assessment and prospective planning by making explicit the corpus's mix of study types, data sources, and geographic settings.
Aggregate agreement with manual labels on three of these four dimensions (\Cref{sec:epilens_performance}) suggests that such a characterisation can be obtained with limited manual coding and updated as new studies appear.
The fourth dimension, \emph{data accessibility}, is both the hardest for the workflow and arguably the most directly preparedness-relevant: knowing whether the data behind past analyses were made available and reusable is a prerequisite for any rapid re-analysis when a new threat emerges.
Because this axis has lower agreement, and lower stability under sampling and with the served models, improving it is a priority.
We examine why it is hard, and how to address it, in \Cref{sec:limitations_future_work}.

\subsection{Limitations and Future Work}
\label{sec:limitations_future_work}

RAG reduces unsupported generations by conditioning outputs on retrieved passages, but errors can still occur when relevant evidence is missed, incomplete, or incorrectly interpreted.

\paragraph{Evaluation scope.}

The present evaluation assumes that the coding scheme can be specified independently of the corpus being evaluated.
Our four axes were defined in advance using explicit operational criteria and externally anchored distinctions, including the United Nations M49 standard for geography, with Northern America, Central America, and the Caribbean merged into North America \citep{un_m49}, and the traditional/non-traditional data distinction from previous work~\citep{chafetz_data4covid}.
This lets us evaluate the resulting fixed scheme directly on the annotated sample.
Other applications may instead derive or improve their categories after inspecting the corpus or preliminary outputs.
In that setting, the workflow's ability to revise categories in natural language is still useful, but the evaluation design should separate a development set used to refine the scheme from a held-out set used only for final performance estimation.

\paragraph{Corpus and annotation.}

Some limitations arise from the case-study corpus rather than from \toolname{} itself.
The corpus was assembled through a semi-automated pipeline (\Cref{sec:corpus,app:corpus_pipeline}).
This process is explicitly best-effort and logs exceptions such as inaccessible PDFs, incomplete references, and non-machine-readable supplements, but downstream retrieval and classification can only operate on material that has been correctly identified, parsed, and indexed.
At the same time, \toolname{} can provide useful downstream warning signals, for example when an unexpected paper-type label suggests that a spurious document has entered the corpus.
Another limitation is that a single annotator produced the evaluation labels following the explicit coding protocol in \Cref{app:labeling_protocol}.
Operational definitions and tie-breaking rules reduce ambiguity, but they do not allow us to report inter-annotator agreement.
A multi-annotator evaluation with adjudicated labels is therefore an important next step.

\paragraph{Licences and served models.}
Because served models were run only on the $122$ openly licensed papers (\Cref{sec:evaluation_setup}), and the corpus is described with the fully local workflow, this subset is not a random sample of the evaluation papers, so the served models' scores indicate their performance rather than measure it on the whole sample, and comparisons on it have less statistical power.
The fully local workflow is also less accurate than \texttt{gemini-2.5-pro} on every axis and than \texttt{gemini-2.5-flash} on all but geography, particularly on data accessibility.
On the manually labelled sample, it detects model-based and simulated-data papers less often than the annotator (sensitivity $0.65$ and $0.68$, specificity at least $0.97$), so the shares of these labels reported for the corpus and for the papers associated with parameter estimates (\Cref{sec:full_corpus_distributions,sec:provenance}) are probably lower than manual labelling would give.
On the openly licensed papers, the served models identify inference and forecast papers more reliably (F$_1$ $0.775$--$0.935$, against $0.364$--$0.623$; \Cref{app:tab:per_label_f1_open}), so the under-detection lies mainly with the local generator.
The same criterion applies to corpus construction: Gemini read only the source reviews with an open licence, and we extracted the included studies from the others by hand (\Cref{app:corpus_pipeline}).

\paragraph{Tracing estimates to papers.}
Our study links estimates to their cited sources.
For an estimate citing a review, the associated set contains its retrieved included studies, which may include related infections.
The link does not identify which studies contributed to the reported value.
Parameter-specific study lists, where reviews provide them, would allow more precise links, as the \texttt{epireview} data illustrate \citep{epireview}.

The corpus contains $61\%$ of the reported inclusions in expanded source reviews with available counts, with coverage varying across reviews (\Cref{app:provenance}).
Restricting the reproduction-number--fatality-risk comparison to better-covered reviews preserves its direction but changes the individual label shares.
The profiles also use model-assigned labels.
The manual labels show the same ordering for the main contrasts, while the small labelled sets behind some pathogen--parameter comparisons limit how precisely their differences can be assessed.

\paragraph{Evidence retrieval and verification.}

The ablation points to evidence selection as one important target for improvement.
Relevant statements may be missed because chunking separates them from their context, the retrieval query does not match the terminology used in the paper, or ranking places the decisive passage below the cut-off.
This is most visible for \emph{data accessibility}, which gains the most from targeted retrieval and also benefits most from the stronger generator (\Cref{sec:served_reference,sec:ablation}).
Improving this axis therefore calls for attention to both evidence selection and subsequent interpretation.
Promising directions are hybrid lexical and dense retrieval, section-aware retrieval targeting locations such as Methods and Data Availability, stronger reranking, and direct measurement of retrieval recall separately from downstream classification performance.
The accessibility axis also illustrates an important scope distinction: our protocol evaluates what the paper \emph{reports} about access to its data, not whether the stated access mechanism still works.
A natural extension is to add a verification stage that checks resolvable links, accession identifiers, repositories, or request mechanisms after extracting the reported access condition.

\paragraph{Reliability and repeatability.}

Repeated runs can flag difficult cases only when outputs vary.
In the fully local deployment at $T{=}0$, repeats are identical, so run-to-run disagreement carries no signal.
At $T{=}1$, and for the served models, disagreements concentrate on data accessibility (\Cref{tab:stability_all_configs_open}).
For the deterministic local configuration, comparable signals can come from sampling at $T{=}1$ or from disagreement between models.
The two served models, for example, largely share their errors on three of the four tasks ($81$--$92\%$ containment at $T{=}0$ on the openly licensed papers, \Cref{app:containment}).
Neither signal should be interpreted as calibrated confidence without further validation.

\paragraph{Extensions.}

Finally, a package like \toolname{} benefits from the continuing improvements in language models and retrieval components.
Stronger locally runnable models may narrow the remaining gap to the served models, while task-specific fine-tuning of smaller models could provide a practical option when labelled examples are available and a classification scheme has stabilised.
Better embedding models, rerankers, and structured decoding methods can likewise be incorporated.
More agentic approaches could also be useful when kept deliberately constrained, for example by triggering a second retrieval step, checking an external identifier, or requesting additional evidence only for papers whose initial classification is ambiguous or unstable.
Such selective escalation is preferable here to long autonomous agent chains, since it preserves the predictable cost, provenance, and inspectability that motivate the framework in the first place.

\subsection{Ethical Aspects}
\label{sec:ethical_considerations}

Epidemiological evidence informs surveillance, resource allocation, and public-health policy.
Errors in its organisation or interpretation may therefore have serious consequences, particularly during emergencies.
We therefore state \toolname{}'s intended role, the risks related to its use, and the safeguards needed to reduce them.
As with any analytical tool used in a consequential setting, its outputs should be interpreted within clearly defined conditions of use and by people able to assess them.

\toolname{} acts as a filter through which evidence is viewed.
Its classifications can affect which studies are surfaced, inspected, prioritised, or treated as absent, so systematic classification errors could distort the resulting map of the literature.
Their practical consequences depend on how the labels are used.
When labels only prioritise papers for expert review, false positives mainly add verification work.

When labels are used to exclude papers or stop screening, false negatives can alter the apparent evidence base by removing relevant studies from view.
Work on ML-assisted screening therefore treats the risk of omitted studies as something that must be assessed and communicated~\citep{callaghan2024computer}.
Such omissions are harder to detect and correct during outbreaks, when decisions are urgent, scientific information is uncertain and changing, and time and resources are limited.
If they are also unevenly distributed throughout regions or languages, the resulting map may reinforce existing under-representation.
WHO raises a related distributional concern for scientific research supported by large multimodal models~\citep{world2024ethics}.

Accordingly, \toolname{} should support evidence triage and corpus mapping rather than replace systematic screening, critical appraisal, or expert judgement.
In this study, the manually labelled evaluation and repeated runs quantify performance and stability for the case-study corpus, while provenance supports inspection of individual outputs.
Applications to a new corpus or coding scheme should therefore be validated on relevant labelled examples, with closer review of unstable outputs and of performance across relevant corpus strata where sample sizes permit.
Domain experts should verify outputs before they inform consequential research or preparedness decisions, and the level of review should reflect the intended downstream use.
Performance should be reassessed after material changes to the corpus, taxonomy, retrieval pipeline, or underlying model.
Users should also check whether the terms under which they hold the documents allow sending them to a remote model; a fully local deployment avoids the question.

\section{Conclusions}
\label{sec:conclusion}

We presented \toolname{}, a RAG package that applies analyst-defined questions identically across a corpus of epidemiological papers, returns structured, validated outputs, and keeps a path from each output back to its evidence.
Without task-specific training labels and running entirely on local hardware, the Paper Classifier outperforms the zero-shot approaches tested and is comparable to supervised classifiers trained on manual labels, ahead of them on geography; the strongest served model scores higher where licences allow its use.
Agreement with manual labels is strongest for geography, data type, and paper type, while data accessibility remains more difficult.
The ablations show that finding relevant evidence in full texts is an important part of this performance.

We also applied the classifier to \numpapers{} outbreak-related papers assembled from the cited sources of a multi-pathogen study of epidemiological parameters.
The labels describe this corpus as a whole.
Linking them to the study's cited sources shows that the kinds of papers associated with its parameter estimates differ across parameters and pathogens.

More broadly, we argue for treating tools such as \toolname{} as part of epidemiology's analytical infrastructure.
Their function is not to replace expert review, but to shift expert effort from manually processing every paper toward defining meaningful questions, validating how reliably they can be answered, and interrogating the evidence behind the resulting map of the literature.
Because the analytical scheme, retrieval components, and language model remain separable, the same package can incorporate improvements in models and retrieval without sacrificing provenance or inspectability.
Because it runs on local hardware, it can also be applied to literature whose licences do not allow sending full texts to external AI services.
As corpora grow and models improve, such tools can help ensure that the knowledge accumulated across past epidemics remains reachable and actionable when the next threat emerges.

\backmatter

\bmhead{Abbreviations}
AI: artificial intelligence.
API: application programming interface.
BCa: bias-corrected and accelerated.
CCHF: Crimean--Congo haemorrhagic fever.
COVID-19: coronavirus disease 2019.
CV: cross-validation.
DOI: digital object identifier.
JSON: JavaScript Object Notation.
LDA: latent Dirichlet allocation.
LLM: large language model.
LOO: leave-one-out.
LSA: latent semantic analysis.
MERS(-CoV): Middle East respiratory syndrome (coronavirus).
ML: machine learning.
MLP: multilayer perceptron.
NLI: natural language inference.
NMF: non-negative matrix factorisation.
NTD: non-traditional data.
PCR: polymerase chain reaction.
PDF: Portable Document Format.
PLSA: probabilistic latent semantic analysis.
RAG: retrieval-augmented generation.
RVF: Rift Valley fever.
SARS: severe acute respiratory syndrome.
SARS-CoV-1, SARS-CoV-2: severe acute respiratory syndrome coronavirus 1 and 2.
SD: standard deviation.
SVM: support vector machine.
TEI: Text Encoding Initiative.
TF-IDF: term frequency--inverse document frequency.
UI: user interface.
URL: uniform resource locator.
WHO: World Health Organization.
XML: Extensible Markup Language.

\section*{Declarations}

\bmhead{Availability of data and materials}
The datasets supporting the conclusions of this article will be made available on Zenodo~\citep{epilens_reproducibility}.
The reproduction package \citep{epilens_reproducibility} contains the corpus bibliography, the \(214\)-paper manually labelled evaluation set, per-paper and aggregate outputs, and a per-estimate file recording pathogen, parameter, source author, year and DOI, and links to corpus papers. Its scripts regenerate the reported tables and figures from the released inputs.
Predictions of the fully local workflow will be released for all papers; predictions of the served models will be released only for the \(122\) openly licensed evaluation papers (\Cref{app:licences}).

The software is available as follows.\\
Project name: \toolname{}.\\
Project home page: \url{https://github.com/VinsRR/EpiLens} \citep{epilens_software}; also distributed on PyPI as \texttt{epilens}.\\
Operating system(s): platform independent.\\
Programming language: Python (version 3.10 or later).\\
Other requirements: the Python dependencies declared in the package; optionally a GROBID server for parsing, a MongoDB server for document storage, a Qdrant server for vector storage, and a local model server such as Ollama.\\
License: GNU Affero General Public License v3.0 (AGPL-3.0-only).\\
Any restrictions to use by non-academics: none beyond the licence terms.

The article PDFs and full-text index cannot be redistributed because of publisher copyright.
The seed study's extraction sheets \citep{ward2025preprint} were shared privately and are not redistributed. The released source links contain no parameter values.

\bmhead{Competing interests}
The authors declare that they have no competing interests.

\bmhead{Funding}
This project has received funding from the European Union's Horizon Europe research and innovation programme under grant agreement No 101095619 (ESCAPE).

\bmhead{Authors' contributions}
VP designed and implemented \toolname{}, built the corpus, produced the manual labels, ran the experiments, and wrote the manuscript. CC and DP supervised the study. All authors approved the manuscript.

\bmhead{Use of AI tools in preparing this work}
Apart from the models evaluated in this study, we used OpenAI Codex to write and test parts of the analysis code and to revise parts of the text.
The authors reviewed all such output and take full responsibility for the content.

\bmhead{Acknowledgements}
We thank Jack Ward and W. John Edmunds for sharing the seed study's early data extraction sheets and approving release of the derived source links.

\bmhead{Ethics approval and consent to participate}
Not applicable.

\bmhead{Consent for publication}
Not applicable.

\begin{appendices}
\counterwithin*{table}{section}
\counterwithin*{figure}{section}
\renewcommand*{\theHtable}{\thesection.\arabic{table}}
\renewcommand*{\theHfigure}{\thesection.\arabic{figure}}

\section{System and workflow details}
\label{app:system_details}
This appendix adds component-level detail to the system summarised in \Cref{sec:methods}.
Here we specify the components and backends used at each stage of that pipeline.

\subsection{Pipeline components and backends}
Each pipeline stage is handled by distinct, interchangeable components that a researcher can swap to suit their setup.

\subsubsection{Phase 1: Ingestion and Indexing}
This phase transforms a collection of scientific papers into a searchable index.
It involves parsing, chunking, embedding, and storing the content.

\paragraph{Document Parsing.}
The local \toolname{} CLI defaults to \texttt{pdfminer.six} for PDFs and uses Unstructured~\cite{Unstructured} for other formats.
For academic papers, it can instead use GROBID~\cite{grobid}, which extracts plain text, rich metadata (title, authors, journal), structured sections (abstract, methods, results), and a bibliography.
All experiments in this paper use GROBID.
Parsed content is stored in memory or in MongoDB; our experiments use MongoDB.

\paragraph{Text Chunking.}
To create focused, contextually relevant units for retrieval, we segment the parsed text sections into smaller chunks.
\toolname{} provides many of the standard strategies, including splitting by paragraph, sentence, or fixed-size overlapping windows.
In all experiments, we split by paragraph.

\paragraph{Embedding.}
Each text chunk is mapped to a high-dimensional vector using an embedding model.
\toolname{} is model-agnostic and supports a range of providers, including open-source sentence-transformer models from Hugging Face, as well as API-based models from OpenAI, Google (Gemini), and local models served via Ollama.
This grants flexibility in regulating performance, computational cost, and privacy.

\paragraph{Vector Storage.}
The generated embeddings and their associated metadata (e.g., paper ID, section title, text) are stored in a vector database.
This database provides the core functionality for efficient similarity search.
\toolname{} includes swappable backends for \texttt{Qdrant}, \texttt{FAISS}, and a simple file-based storage system, enabling deployment in a range of environments from local machines to server-based setups.

\subsubsection{Phase 2: Retrieval}
The retrieval phase calls the vector database to find the most relevant indexed text chunks in response to a user query.

\paragraph{Query Processing.}
\toolname{} supports multiple retrieval modes.
The primary mode is \emph{semantic search}, where a user query is embedded using the same model as the document chunks, and the vector database is searched for the "nearest" chunk embeddings.
For the experiments in the main text, we used \texttt{Qdrant}.
A \emph{keyword search} is also available for simple term matching.

\paragraph{Similarity Search and Filtering.}
The core retrieval operation is a nearest-neighbor search performed by the vector database.
Results can be filtered by metadata, for instance by restricting the search to a specific paper ID or a set of paper sections (e.g., \texttt{methods}).
For the experiments presented in the main text, we used cosine similarity as the distance metric.
The standard RAG approach uses the user query as a reference for finding similar chunks.
We additionally use a template-based approach (see~\cref{app:system:workflows}).

\subsubsection{Phase 3: Generation}
In the final phase, an LLM synthesises a coherent answer from the retrieved evidence.

\paragraph{Prompt Engineering.}
The system compiles the retrieved text chunks into a carefully structured prompt, together with the original user query.
The prompt explicitly instructs the LLM to base its answer strictly on the provided evidence and to avoid using its internal knowledge.
The only exception is geographic: the model may use its own knowledge to map a stated location to its continent (\Cref{app:geo_rules}).
For the workflows discussed in \cref{app:system:workflows}, we provide a schema to guide the model to generate JSON output with specific characteristics.

\paragraph{Answer Synthesis.}
\toolname{}'s \texttt{Generator} component sends the prompt to a local or remote LLM through a provider client.
This lets users select a model that fits their performance, privacy, and cost requirements.

\paragraph{Provenance and Inspectability.}
A core principle of \toolname{} is to preserve transparency.
Every generated answer is accompanied by a list of evidence items, each linking back to a precise text chunk from a source paper and containing the source excerpt, the paper ID, and the section.
This makes it possible to inspect the path from query to answer and check the output against the cited source passages.

\subsection{High-Level Analytical Workflows in \toolname{}}
\label{app:system:workflows}

The structured-workflow contract (a fixed retrieval \emph{template} plus a validating JSON \emph{schema}, implemented with Pydantic~\citep{colvin2024pydantic} and applied identically across a corpus) is defined in \Cref{sec:methods}.
Here we record the implementation specifics of the two instantiations used in this paper, the \textbf{Precision Miner} (extraction) and the \textbf{Paper Classifier} (classification).
They reuse the same ingestion, retriever, generator, validation, and provenance components, but use workflow-specific evidence aggregation, prompts, schemas, and parsing.

\subsubsection{The Precision Miner for Structured Extraction}

The Precision Miner is the extraction instantiation of the pattern.
It finds and lists specific, concrete pieces of information in a paper, such as the data sources used, links to supplementary materials, or key bibliographic references.

\paragraph{Methodology.}
The miner's implementation of the pattern is geared towards maximising the discovery of relevant facts within a document.
\begin{itemize}
    \item \textbf{Retrieval phase (multi-query aggregation):} The miner is configured with several queries designed to find relevant passages (for example, ``What datasets were used in this study?'' or ``Where can supplementary data be found?'').
    It searches each paper separately for every query, can restrict candidates by section, and combines the results into one deduplicated, score-ranked evidence pool.
    The multi-query strategy is intended to broaden candidate coverage without guaranteeing recall.
    \item \textbf{Generation phase (the LLM as an extractor):} The consolidated evidence pool is formatted into a prompt and sent to the LLM, which is given the role of an information extractor: it scans the snippets, identifies every instance of the target information (for example, dataset names and URLs), and organises them into a list.
    The JSON schema requires an object containing this list, where each item has fields such as \texttt{name}, \texttt{url}, and \texttt{explanation}, with the source sentence retained as provenance.
\end{itemize}

\paragraph{Worked outputs.}
\label{app:precision_miner_outputs}
We ran the \texttt{find\_data\_sources} variant of the miner over the corpus index on three corpus papers from different pathogen families, to show what an extraction workflow returns rather than to benchmark its accuracy.
\Cref{tab:precision_miner} lists the result: for each paper, a variable-length list of concrete data sources, each with a link or accession where the paper supplies one.
The recovered sources are heterogeneous, from public code and data repositories (GitHub, Zenodo, GenBank) to surveillance systems, health records, reports from the World Health Organization, and the cited studies behind model parameters.
Other variants ship with the package to find supplementary materials or important references.
The table is limited to paper identifiers, source names, and links or accessions: the miner also returns source sentences and model explanations for auditing a run, but these can reproduce publisher text and are not redistributed.

\begin{table}[!htbp]
  \centering
  \small
  \setlength{\tabcolsep}{4pt}
  \caption{\textbf{Output of a second structured workflow: the data-source Precision Miner.}}
  \label{tab:precision_miner}
  \begin{tabular}{@{}>{\raggedright\arraybackslash}p{0.21\textwidth} p{\dimexpr 0.79\textwidth-2\tabcolsep\relax}@{}}
  \toprule
  \textbf{Paper} & \textbf{Data sources extracted (name and link or accession where given)} \\
  \midrule
  Lauer et al.\ \citep{lauer2020incubation} \newline \emph{COVID-19 incubation} &
  \begin{itemize}[topsep=1pt,itemsep=1pt,leftmargin=1.3em,label=\textbullet]
    \item News and public health reports
    \item Johns Hopkins Center for Systems Science and Engineering
    \item HopkinsIDD/ncov\_incubation GitHub repository ({\footnotesize\url{https://github.com/HopkinsIDD/ncov_incubation}})
    \item Zenodo record 3692048 ({\footnotesize\url{https://zenodo.org/record/3692048}})
  \end{itemize} \\
  \midrule
  Althaus \citep{althaus2014estimating} \newline \emph{Ebola $R_0$} &
  \begin{itemize}[topsep=1pt,itemsep=1pt,leftmargin=1.3em,label=\textbullet]
    \item World Health Organization (WHO)
    \item Chowell et al. (study)
  \end{itemize} \\
  \midrule
  Nakoune et al.\ \citep{nakoune2017nosocomial} \newline \emph{Mpox, CAR} &
  \begin{itemize}[topsep=1pt,itemsep=1pt,leftmargin=1.3em,label=\textbullet]
    \item DDBJ/EMBL/GenBank database
    \item KP849469, A27L gene (reference sequence)
    \item Standardized tools for data collection (Ministry of Health and World Health Organization)
    \item Consultation registers, health records, and hospitalization forms
    \item Institut Pasteur de Bangui (IPB) reports of suspected cases
  \end{itemize} \\
  \bottomrule
  \end{tabular}
  \footnotetext{Rights-safe output of the \texttt{find\_data\_sources} miner on three corpus papers from different pathogen families. The released summary contains identifiers, data-source names, and links or accessions. It deliberately omits source sentences, retrieved text, prompts, and model traces.}
\end{table}

\subsubsection{The Paper Classifier for Automated Categorisation}

The Paper Classifier instantiates this pattern for classification.
It is designed to automatically assign categorical labels to an entire paper along predefined axes, such as study design, data accessibility, or geographic focus.

\paragraph{Methodology.}
The classifier is built to gather comparative evidence and facilitate judgement.
\begin{itemize}
    \item \textbf{Retrieval Phase (Evidence Gathering by Category):} Unlike the miner, the classifier's templates are \emph{exemplars}: canonical text snippets that represent each of the possible categories in a classification scheme (e.g., a sentence describing open data access is an exemplar for the \texttt{OPEN} category).
    Each exemplar retrieves its nearest chunks from the paper, and each distinct chunk is assigned to the category whose exemplar it matches best.
    The $k$ highest-scoring chunks across all categories are passed to the model grouped by category, as a structured ``dossier'' of evidence for the competing labels.
    A label whose exemplars match only weakly may receive no chunk.
    \item \textbf{Generation Phase (The LLM as a Classifier):} This comparative dossier is presented to the LLM.
    Its task is not to extract items, but to weigh the evidence for the competing categories and make a final, reasoned decision about which label (or labels) best applies to the paper as a whole.
    The JSON schema requires one or more valid labels and a textual justification.
    A confidence score is optional.
    Schema validation makes the output structured, while the retained passages make the decision inspectable.
\end{itemize}

\subsection{Token usage and projected inference cost}
\label{app:cost}

This section estimates the token cost of running the \toolname{} Paper Classifier over a corpus.
A classification pass has two components: (i) a \emph{one-time} embedding stage that indexes the corpus for retrieval, and (ii) a \emph{per-call} generation stage that classifies each paper on each axis.
We report token usage rather than monetary cost, letting readers convert these figures using the pricing model of their chosen provider.
Embedding costs were estimated from corpus statistics, as described below.

\paragraph{Generation: token usage per call.}
\Cref{tab:cost_tokens} reports the mean token usage of one \toolname{} classification call (one paper, one axis).
Input is dominated by the fifteen retrieved chunks (mean $760$ characters, approximately $190$ tokens each) together with the task instructions.
The visible JSON output is small (about $150$ tokens), but \texttt{gemini-2.5-flash} is a reasoning model, so most billable output consists of internal reasoning tokens (about $1{,}200$ tokens).
Overall, a call averages approximately $3{,}500$ input and $1{,}400$ output tokens, measured on the $122$ openly licensed evaluation papers (\Cref{app:licences}).
The whole-body ablation used $7{,}665$ input tokens per call, against $3{,}510$ for the fixed-chunk proxy, a factor of $2.2$ (\Cref{sec:ablation}).

\begin{table}[t]
  \centering
  \caption{\textbf{Mean token usage per classification case in the cost proxy.}}
  \label{tab:cost_tokens}
  \begin{tabular}{l r r r r}
    \toprule
     & & \multicolumn{2}{c}{\textbf{Output (billable)}} & \\
    \cmidrule(lr){3-4}
    \textbf{Axis} & \textbf{Input} & \textbf{Reasoning} & \textbf{Visible} & \textbf{Total} \\
    \midrule
    Paper type           & $4{,}102$ & $1{,}598$ & $152$ & $5{,}852$ \\
    Geography            & $3{,}293$ & $904$ & $165$ & $4{,}362$ \\
    Data type            & $3{,}151$ & $906$ & $123$ & $4{,}180$ \\
    Data accessibility   & $3{,}495$ & $1{,}470$ & $153$ & $5{,}118$ \\
    \midrule
    All axes (mean) & $3{,}510$ & $1{,}220$ & $148$ & $4{,}878$ \\
    \bottomrule
  \end{tabular}
  \footnotetext{Measured on the random-chunk ablation with \texttt{gemini-2.5-flash}, $T{=}0$, and $k{=}15$, on the openly licensed evaluation papers (\Cref{app:licences}). The targeted workflow uses the same prompts, schemas, generator, and retrieval budget but different evidence selection.}
\end{table}

\paragraph{Generation at corpus scale.}
\toolname{} performs one classification case per paper for each of the four axes.
Averaged across all axes, classifying a single paper requires approximately $20{,}000$ tokens (roughly $14{,}000$ input and $5{,}500$ output tokens).
Scaling the proxy per-case averages to the full corpus of \numpaper{} papers ($9{,}188$ classification cases) gives approximately $32$M input and $13$M output tokens, or about $45$M generation tokens in total (\Cref{tab:cost_projection}).
The $214$-paper evaluation corpus corresponds to approximately $4.2$M generation tokens.
These counts describe the served reference.
Non-reasoning models such as the local \texttt{qwen2.5:32b-instruct} produce far fewer output tokens, and in the fully local workflow the same prompts cost compute time on local hardware rather than billed tokens.

\paragraph{Retrieval: one-time embedding cost.}
Generation is only one component of the overall token budget.
Retrieval additionally requires embedding the corpus, although this cost is incurred only once when constructing the index.
Unlike generation, embedding usage was not metered directly.
Instead, we estimate it from corpus statistics.
Using the observed mean of $33$ retained chunks of about $190$ tokens each gives approximately $6{,}300$ embedded tokens per paper, or roughly $14$M tokens for the complete corpus.
This is an approximation because chunking can overlap text and can omit short fragments.
The cost is amortised across subsequent workflows that reuse the same index.
The queries, a fixed set of short exemplar paragraphs (one or more for each candidate label), are embedded once and reused for every paper, so their cost is negligible.
Overall, corpus indexing contributes an estimated one-time cost of about $14$M embedding tokens, compared with approximately $45$M generation tokens for each complete classification pass (\Cref{tab:cost_projection}).

\begin{table}[t]
  \centering
  \caption{\textbf{Projected token usage of one \toolname{} classification pass over the four axes.}}
  \label{tab:cost_projection}
  \begin{tabular}{l r r r r}
    \toprule
    \textbf{Scope} & \textbf{Cases} & \textbf{Gen. in} & \textbf{Gen. out} & \textbf{Embed (1$\times$)} \\
    \midrule
    Evaluation sample (214 papers) & $856$ & $3.0\,\mathrm{M}$ & $1.2\,\mathrm{M}$ & $1.3\,\mathrm{M}$ \\
    Full corpus (2{,}297 papers) & $9{,}188$ & $32.3\,\mathrm{M}$ & $12.6\,\mathrm{M}$ & $14.5\,\mathrm{M}$ \\
    \bottomrule
  \end{tabular}
  \footnotetext{Projected from the ablation-based per-case proxy in \Cref{tab:cost_tokens}. Generation is incurred per pass. Corpus embedding is a one-time estimate.}
\end{table}

\section{Corpus construction pipeline details}
\label{app:corpus_pipeline}
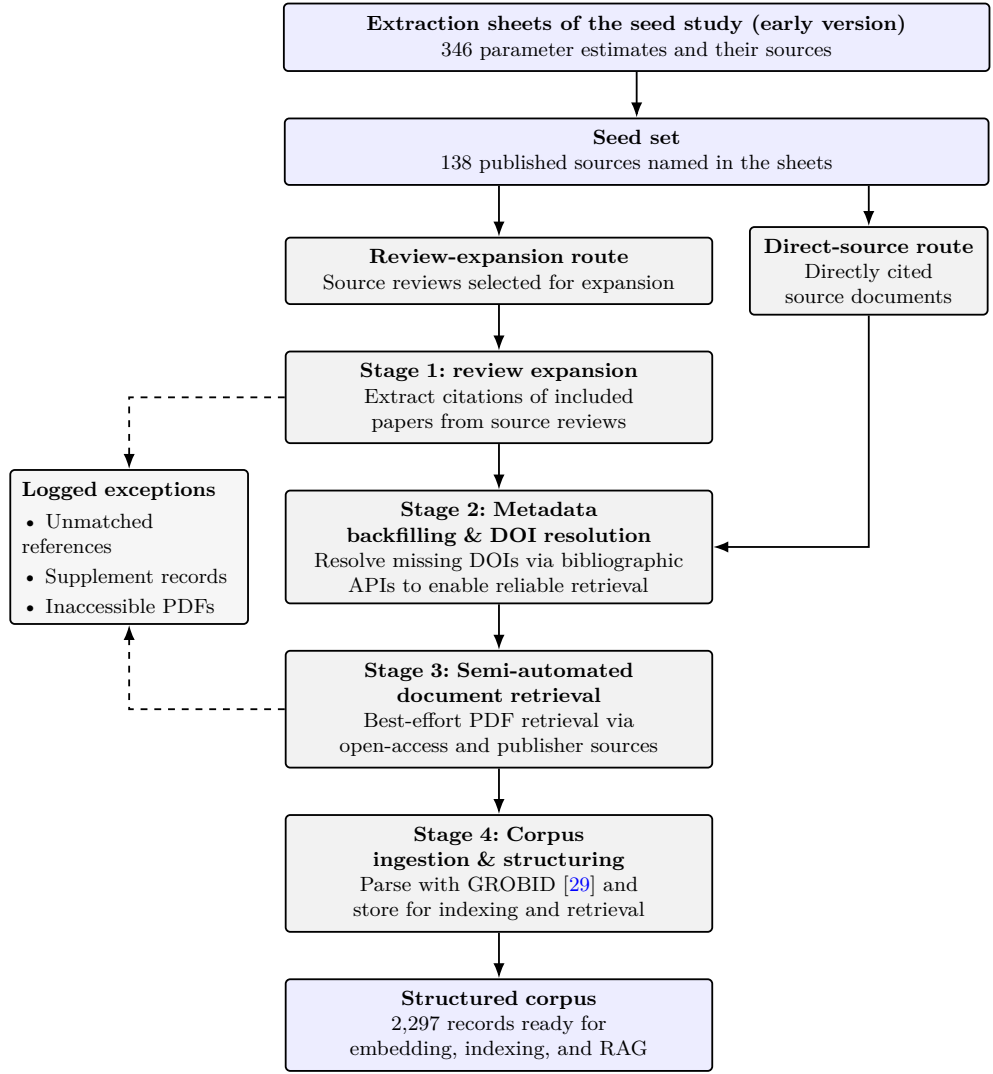
\begin{figure}[t]
  \centering
  \resizebox{\columnwidth}{!}{%
\begin{tikzpicture}[font=\small, node distance=0.7cm and 0.55cm,
  execute at begin node={\hyphenpenalty=10000\exhyphenpenalty=10000}]

\node[processbox, text width=6.2cm] (reviews) {%
  \textbf{Review-expansion route}\\
  Source reviews selected for expansion};

\node[processbox, text width=6.2cm, below=of reviews] (s1) {%
  \textbf{Stage 1: review expansion}\\
  Extract citations of included papers from source reviews};

\node[processbox, text width=6.2cm, below=of s1] (s2) {%
  \textbf{Stage 2: Metadata backfilling \& DOI resolution}\\
  Resolve missing DOIs via bibliographic APIs to enable reliable retrieval};

\node[processbox, text width=6.2cm, below=of s2] (s3) {%
  \textbf{Stage 3: Semi-automated document retrieval}\\
  Best-effort PDF retrieval via open-access and publisher sources};

\node[processbox, text width=6.2cm, below=of s3] (s4) {%
  \textbf{Stage 4: Corpus ingestion \& structuring}\\
  Parse with GROBID~\cite{grobid} and store for indexing and retrieval};

\node[processbox, text width=6.2cm, below=of s4, fill=blue!7] (corpus) {%
  \textbf{Structured corpus}\\
  \numpaper{} records ready for embedding, indexing, and RAG};

\node[processbox, text width=3.3cm, right=of reviews] (direct) {%
  \textbf{Direct-source route}\\
  Directly cited source documents};

\path let \p1 = (reviews.north west), \p2 = (direct.north east), \n1 = {\x2 - \x1} in
  node[processbox, minimum width=\n1, fill=blue!7, anchor=south, yshift=0.7cm]
  (seed) at ($(\p1)!0.5!(\p2)$) {%
  \textbf{Seed set}\\
  138 published sources named in the sheets};

\path let \p1 = (seed.west), \p2 = (seed.east), \n1 = {\x2 - \x1} in
  node[processbox, minimum width=\n1, fill=blue!7, anchor=south, yshift=0.7cm]
  (sheets) at (seed.north) {%
  \textbf{Extraction sheets of the seed study (early version)}\\
  346 parameter estimates and their sources};

\node[processbox, text width=3.3cm, align=left, fill=gray!7, left=of s2] (exc) {%
  \textbf{Logged exceptions}\\[3pt]
  \textbullet\ Unmatched references\\[2pt]
  \textbullet\ Supplement records\\[2pt]
  \textbullet\ Inaccessible PDFs};

\draw[flow] (sheets) -- (seed);
\draw[flow] (seed.south -| reviews.north) -- (reviews.north);
\draw[flow] (seed.south -| direct.north) -- (direct.north);
\draw[flow] (reviews) -- (s1);
\draw[flow] (s1) -- (s2);
\draw[flow] (direct.south) |- (s2.east);
\draw[flow] (s2) -- (s3);
\draw[flow] (s3) -- (s4);
\draw[flow] (s4) -- (corpus);

\draw[flow, dashed] (s1.west) -| (exc.north);
\draw[flow, dashed] (s3.west) -| (exc.south);

\end{tikzpicture}
}%
  \caption{\textbf{Corpus construction pipeline.}
  Source reviews are expanded to include papers; then identifiers are resolved, documents are retrieved, and records are ingested.
  Directly cited source documents bypass review expansion and join the flow at identifier resolution.
  The pipeline tracks unmatched references, supplement-only cases, and inaccessible documents during construction.}
  \label{fig:corpus_pipeline}
\end{figure}

This appendix documents the four-stage corpus construction pipeline summarised in \Cref{sec:corpus}.
Exact reconstruction of the document collection remains access-dependent.
\Cref{fig:corpus_pipeline} illustrates the pipeline.

\subsection{Stage 1: Review Expansion}

The first stage identifies the papers reported as included evidence by each seed source selected for review expansion.
Parsing the complete reference section is insufficient because a review can cite many papers that were not included in its analysis.
We therefore target the subset of references that are explicitly described as included studies in the review text, tables, or appendices.

\paragraph{Methodology.}
Stage~1 identifies the included evidence for each source review by extracting references, followed by rule-based matching.
Extraction is LLM-assisted for the reviews whose stored copy carries an open licence and manual for the others (step~1).
We maintain the extraction and matching code separately from \toolname{}.
For each source review selected for expansion, we execute the following steps.
\begin{enumerate}
    \item \textbf{Candidate extraction.}
    For reviews with an open licence, an LLM reads the review PDF and extracts spans that enumerate included studies.
    This includes narrative ``study characteristics'' descriptions and tables listing included trials or observational studies.
    We use Gemini's document understanding \citep{gemini_files_api,gemini_document_processing}, prompting the model to output the included studies as full bibliographic citations exactly as reported in the review’s reference list.
    We used Gemini only for the source reviews whose stored copy carries an open licence, under the criterion of \Cref{app:licences}.
    For the $20$ reviews without one, we identified the included studies by hand from the review's GROBID parse and tables, assisted by local tools, and matched them to the bibliography as in steps~2 and~3.
    When the included studies are listed in a supplementary file rather than the main text, the model returns the supplement's label (e.g., ``Supplementary Table S1'') and URL instead, and the supplement is processed case by case.
    \item \textbf{Bibliographic parsing and normalisation.}
    In parallel, the review PDF is processed with GROBID to extract structured bibliographic records and reference strings \citep{grobid}.
    \item \textbf{Candidate--bibliography matching.}
    We index bibliography entries by the normalised surname of the first author and link each extracted candidate to one entry.
    A unique exact surname match is selected directly.
    When several entries share that surname, the implementation compares the normalised candidate with each entry's title plus raw reference string using RapidFuzz's \texttt{fuzz.ratio} similarity \citep{rapidfuzz_github}.
    A score of at least $65$ is recorded as title-based disambiguation.
    Below that threshold, the highest-scoring same-surname entry is retained and explicitly marked as an ambiguous best guess.
    If there is no exact surname token, surnames are compared fuzzily and accepted at a threshold of $70$.
    Candidates with no accepted surname match are marked unmatched and excluded from downstream stages.
    This constrains candidates to bibliography entries parsed independently from the source PDF and records the match score and reason for audit.
    \item \textbf{Recorded outputs.}
    The reproduction package releases the final review--paper links in \texttt{review\_membership.tsv} and source-review metadata in \texttt{source\_reviews.tsv}~\citep{epilens_reproducibility}.
    Intermediate extraction runs are not included.
\end{enumerate}

\subsection{Stage 2: Metadata Backfilling and DOI Resolution}

Matched references can lack a DOI or carry incomplete metadata.
Stage~2 retains a DOI parsed by GROBID when present and otherwise attempts to resolve one through Crossref \citep{hendricks2020crossref}.
Directly cited source documents enter the pipeline at this stage.

\paragraph{Methodology.}
For each matched reference without a DOI, the implementation:
\begin{enumerate}
    \item \textbf{Bibliographic query.}
    Queries Crossref with the title, up to two author strings, journal, and, where available, a publication-year filter.

    \item \textbf{Candidate scoring.}
    Ranks the returned records by title similarity using RapidFuzz \texttt{token\_set\_ratio}, dividing its score by $100$ to place it on a $0$--$1$ scale.

    \item \textbf{DOI assignment.}
    Writes the DOI of the best candidate when title similarity is at least $0.85$.
    Otherwise, it retains the unresolved status or query error.
\end{enumerate}

\subsection{Stage 3: Semi-Automated Document Retrieval}

Stage~3 attempted to obtain the documents identified in Stage~2.
The collection that can be recovered depends on access to those documents at the time of retrieval.

\paragraph{Methodology.}

We ran a DOI-driven PDF harvester over the resolved identifiers.
It searched online for full-text copies of each paper, downloaded candidate PDFs, and logged successes and failures.
For unresolved cases, a fallback script searched DOI and PubMed landing pages for PDF links.
We then manually examined failures and suspect files, recovered papers where possible, and verified that the retained files were valid PDFs.
The archived download records also identify documents already obtained through another review.
The released corpus manifest lists the papers retained for analysis \citep{epilens_reproducibility}, with identifiers that readers can use to attempt retrieval under their own access conditions.
Files of directly cited sources were named by year, first author, and parameter.

\subsection{Stage 4: Corpus Ingestion and Structuring}

Stage~4 converts the retrieved documents into the structured representation \toolname{} uses for downstream indexing and RAG workflows.

\paragraph{Methodology.}
We process each document as follows.
\begin{enumerate}
    \item \textbf{Deep parsing.}
    Each PDF is parsed with GROBID into structured TEI XML, including section boundaries, metadata, and a parsed bibliography \citep{grobid}.
    \item \textbf{Corpus database population.}
    We insert the parsed representation into the corpus database (MongoDB) as the source of truth for retrieval and provenance.
\end{enumerate}
The frozen manifest contains \numpaper{} corpus records used by every downstream analysis.
The resulting records are then available for embedding, vector indexing, and query-dependent retrieval.

\subsection{Licences of the evaluation papers}
\label{app:licences}

This section describes our process for selecting papers for served-model experiments based on their licences (\Cref{sec:evaluation_setup}).
While the fully local workflow included all 214 papers, we restricted the served-model experiments based on licensing requirements.
We adopted a conservative criterion: we selected papers for served-model experiments based on the licence of the copy available to us, relying on Creative Commons licences or public-domain/CC0 status rather than making a general determination about compatibility with provider terms.
When necessary, we verified external metadata to confirm licensing information.
Our selection includes papers under Creative Commons licences that contain non-commercial or no-derivatives clauses.
This process resulted in the selection of \(122\) out of \(214\) sampled papers: \(54\) under CC BY, \(1\) under CC BY-SA, \(27\) in the public domain or under CC0, and \(40\) under licences with non-commercial or no-derivatives clauses.
Publisher-specific permissions were not considered open licences. These include Elsevier's user licence (\(3\) papers), research permissions for COVID-19 and mpox content (\(4\)), temporary COVID-19 access granted by other publishers (\(7\)), and free-to-read copies without a stated licence (\(2\)).
These papers are among the \(92\) excluded from all analyses involving a served model.
The reproduction package specifies the licence and version used for each paper.

\section{Labelling protocol}
\label{app:labeling_protocol}
This appendix defines four classification axes: geography, accessibility, data type, and the paper's main research contribution.
The first three describe the \emph{datasets} a paper uses.
The fourth describes the \emph{paper} as a whole.
Making these principles explicit is essential both to define the labels used as ground truth and to preserve consistency between these criteria and the instructions in the LLM prompts, which would otherwise default to unknown and potentially inconsistent standards.
Specifically, some documents may plausibly fall into more than one category, depending on interpretation, or may be sensitive to where boundaries are drawn between categories in the context of our specific classification task.
Potentially ambiguous cases include unclear data availability statements (e.g., claims that data are in the public domain, which may apply only under specific emergency conditions), the appropriate geographic labelling of analyses of contagion events originating in one country but detected in another, and the classification of paper types consistent with the practices of the relevant scientific community (in this case, epidemiology).

For the three data axes (geography, data accessibility, and data type), classification applies to the individual datasets within a paper, not to the paper itself.
Each dataset used to obtain results via the described methodology is assigned a label.
This excludes datasets that, e.g., are only mentioned in a background section.
The final output for each axis is the union of these per-dataset labels, so duplicates are automatically collapsed: if three datasets are assigned labels $A$, $A$, and $B$, the resulting set is $\{A,B\}$.
The fourth axis, paper type, is instead assigned at the level of the whole paper, as described in its own section below.

\subsection{Label set}
We assign labels along four axes, the first three abbreviated GEO, DAVAIL, and \texttt{DTYPE} below:

\begin{enumerate}
  \item \textbf{Geography.}
  A set-valued label representing the continents associated with the epidemiological evidence used in the paper, or whether this information is irrelevant, e.g., in the case of laboratory settings.
  \item \textbf{Data accessibility.}
  A set-valued label recording the access mechanisms stated for the data used in the paper, or whether printed values permit limited in-paper reconstruction.
  \item \textbf{Data type.}
  A set-valued label capturing whether the contribution uses data and the nature of the data used for the contribution, i.e., whether it uses traditional data, non-traditional data~\cite{chafetz_data4covid}, or synthetic data.
  \item \textbf{Paper type.}
  A primary label used in evaluation, with optional secondary labels retained in the stored output, denoting the paper's research-contribution type with respect to epidemiological parameter estimation.

\end{enumerate}

Labels are assigned exclusively based on information provided within the document, disregarding external knowledge or verification.
For instance, if a paper states that its data are available in an online repository, we use that \emph{statement} regardless of whether the repository is still accessible.
This is motivated by our focus on extraction fidelity, not, for example, on verifying data accessibility.
The only exception is the mapping of geographic locations to continents.
Here we allow the model to map a location to its geographic label using its internal knowledge, e.g., mapping a nation to its continent, providing guidance only for the ambiguous cases discussed below.

Geography, data type, and paper type include an \texttt{UNCLEAR} fallback for insufficient evidence, distinct from \texttt{IRRELEVANT} (geography) and \texttt{OTHER} (paper type).

\subsection{Geographic labelling}
\label{app:geo_rules}
This pipeline is intended as a first-stage, high-recall filter for identifying subsets of papers that merit subsequent, more specialised analyses.
Accordingly, we adopt an intentionally inclusive definition of geographic information: we include all locations that are \emph{explicitly tied} to the epidemiological evidence used in the paper, even when multiple plausible ``geographic perspectives'' coexist (e.g., location of exposure vs.\ location of detection/ascertainment).
For instance, when infections are acquired in one country but cases are detected, diagnosed, or reported in another, we include \emph{both} locations in the multi-label GEO output.
We do not include a location mentioned only incidentally: if a paper analyses an outbreak in continent A and merely notes similar patterns in continent B, the correct label is A alone.
More granular labelling (e.g., separating exposure from ascertainment into distinct labels) is feasible, but is beyond our current aim.

There are, however, cases in which location information is considered irrelevant.
If the analysis is conducted entirely in controlled laboratory settings such that geographic or ecological variation is not meaningful (e.g., laboratory-bred animals or strictly controlled experimental conditions), the GEO axis is marked as \texttt{IRRELEVANT}.
The same label applies when the paper relies on purely synthetic or otherwise non-geographic data that are not meaningfully tied to any real-world location.
By contrast, wildlife, farm, market, household, and other field-like settings are treated as geographically meaningful and labelled using the locations the authors state.

Consistent with our general principle of relying only on stated information, we do not infer geographic origin from strain identifiers, sample naming conventions, or external domain knowledge unless the paper explicitly states the geographic origin in the text.

The continent label set is Africa, Asia, Europe, North America, South America, and Oceania (together with \texttt{IRRELEVANT} and \texttt{UNCLEAR}).
Some countries straddle two continents and are intrinsically ambiguous under informal conventions.
For these cases, we fix a single, reproducible country-to-continent assignment based on the United Nations Statistics Division M49 standard (``Standard Country or Area Codes for Statistical Use'', Series M, No.~49) \citep{un_m49}, rather than ad hoc geopolitical or cultural criteria.
M49 regions map directly to our Africa, Asia, Europe, and Oceania labels.
M49 does not define North and South America as separate regions: it divides the Americas into Northern America and Latin America and the Caribbean, with the latter further divided into the Caribbean, Central America, and South America.
We therefore label countries in M49's Northern America, Central America (including Mexico and Panama), and Caribbean as North America, and countries in M49's South America as South America.
For transcontinental countries where the pathogens in scope, those with pandemic potential, have been reported, the annotation prompt uses fixed assignments: Russia to Europe, Egypt to Africa, and Turkey, Kazakhstan, Azerbaijan, Georgia, Armenia, and Cyprus to Asia.

Other sources of ambiguity, such as no-longer-existing political entities (e.g., USSR, Yugoslavia) or politically disputed/recognition-dependent territories (e.g., Kosovo), are largely avoided in this setting because our evaluation targets continent-level labels.
By collapsing geographic mentions onto continents, we anchor the GEO axis to relatively stable, predominantly geographic reference entities, reducing sensitivity to changes in statehood, naming, or recognition.
Finally, for papers covering a global scope, we list all continents.

\subsection{Data accessibility labelling}

\label{app:davail_rules}
DAVAIL records what the paper states about access to each dataset and whether the paper prints enough values to meet the \texttt{REPORTED} re-aggregation rule.
It measures reported accessibility and in-paper reconstructability, not verified access or general dataset reusability.

\begin{description}
  \item[\texttt{OPEN}.]
  The data are publicly available via a persistent identifier (e.g., DOI, accession number), a stable link to a repository, or by being fully released within the paper package (e.g., full line-lists or complete contingency tables) in a form sufficient for a third party to reconstruct the dataset for new analyses.
  Mentions of supplementary material only qualify if they contain the specific, extractable data used for the work (the link/identifier must point to the underlying data supporting the results, not just be a ``supplementary information exists'' pointer).

  \item[\texttt{REPORTED}.]
    The dataset is not fully released, but the paper provides granular, extractable data values (e.g., detailed frequency tables, stratified counts, or partial time-series) that allow a third party to independently recompute at least one primary summary statistic used in the analysis (e.g., a frequency, proportion, or rate) \emph{from the reported values alone}, without requiring access to unreported data or relying on the authors’ interpretive conclusions.
    This label applies based on the richness of the in-paper reporting, regardless of whether the data source is internal or a third party.
    It does not imply access to the underlying dataset or the ability to reproduce the full analysis.

    \item[\texttt{REFERENCED}.]
    The data are credited to a \emph{specifically identifiable third-party source} (e.g., public health authorities, institutional databases, emergency dashboards, named public sources, other papers), but the paper provides neither a stable retrieval path for the source data nor sufficient detail to re-aggregate the data or re-use them as inputs for other analyses without relying on the author's interpretation (i.e., fails the \texttt{REPORTED} re-aggregation criterion).
    This data-thin reporting (limited to derived outputs like effect sizes, coefficients, or fitted model curves) prevents independent verification.
    Vague attributions (e.g., "government data") do not qualify (lean towards \texttt{NOT\_STATED}).

    \item[\texttt{AVAILABLE\_\allowbreak UPON\_\allowbreak REQUEST}.]
    The paper explicitly provides a request-based mechanism, such as contacting the author, a data access committee, or a specific institutional process.

    \item[\texttt{CLOSED}.]
    The paper explicitly states that the data cannot be shared or are restricted due to legal, ethical, or licensing constraints, without providing an actionable access mechanism.

    \item[\texttt{NOT\_STATED}.]
    The paper provides no data access statement, no data repository link, no request mechanism, and no identifiable third-party attribution, and it also fails to provide extractable values sufficient for \texttt{REPORTED}.
    Statements that data were collected/generated, without an access mechanism and without re-aggregatable values, should be labelled \texttt{NOT\_STATED}.
\end{description}

Precedence is applied within each dataset, after which the paper-level output is the union of the per-dataset labels.
For one dataset, use \texttt{OPEN} when a qualifying public route is stated.
Otherwise use \texttt{AVAILABLE\_\allowbreak UPON\_\allowbreak REQUEST} when an actionable request route exists, and \texttt{CLOSED} for an explicit restriction without such a route.
If none applies, distinguish \texttt{REPORTED} from \texttt{REFERENCED} by the re-aggregation criterion, and use \texttt{NOT\_STATED} when neither applies.

\subsubsection{Operational scope}
The protocol is descriptive rather than a compliance assessment: it classifies what a paper states or prints, using categories adapted to the heterogeneous availability statements encountered across journals and communities.
Two rules make the categories operational for both annotation and prompting: \texttt{OPEN} requires a stable retrieval route or a reconstructable in-paper release, and \texttt{REPORTED} requires enough printed values to recompute at least a primary summary statistic.

\subsection{Data type labelling}

The \texttt{DTYPE} labels identify the nature of the data used to support a paper's contribution.
\texttt{DTYPE} is set-valued because a paper can use several data types.
The operational \texttt{NO\_EMPIRICAL\_DATA} output is exclusive and is used only when the contribution relies on neither empirical nor simulated data.
The traditional, health, mobility, economic, and sentiment categories build on the \#Data4COVID19 Review \cite{chafetz_data4covid}.
\texttt{SYNTHETIC} (reported as ``simulated data'' in figures and tables) and \texttt{NO\_EMPIRICAL\_DATA} are extensions introduced for this corpus.
In a nutshell, NTD refers to information digitally captured, mediated, or observed through new instrumentation mechanisms, often privately held and repurposed for uses unrelated to its initial collection.

\begin{description}
\item[\texttt{TRADITIONAL}] when a paper relies on standard epidemiological and public-health data sources collected through established surveillance infrastructure, regardless of digital vs analogue collection methods.
Examples include daily or weekly case counts from a national surveillance system, hospital admissions and deaths, and PCR- or antigen-confirmed test line lists.

\item[\texttt{NON\_TRADITIONAL\_HEALTH}] novel or
non-standard health data sources that emerged outside traditional
public-health surveillance systems, such as: (i) digital patient data from large-scale electronic-health-record repositories, (ii) platform-facilitated self-reports and symptom proxies collected through apps or web platforms, (iii) wastewater surveillance as a community-level proxy for spread, and (iv) wearable or biometric signals.

\item[\texttt{NON\_TRADITIONAL\_MOBILITY}] when the paper uses mobility data, understood as data describing the location of a device relative to people/landmarks in the physical world or to other devices.
This includes mobility signals derived from telecommunications (e.g., call detail records), location data from software development kits embedded in smartphone apps, GPS and vehicle traces, and proximity technologies (e.g., Bluetooth), when used as epidemiological proxies (e.g., population movement patterns, contact opportunities).

\item[\texttt{NON\_TRADITIONAL\_SENTIMENT}] when the paper uses NTD intended to capture how individuals or groups perceive and interpret epidemic-related developments and interventions.
This category includes social-media, crowdsourced, app-based, and online-panel data used to track behaviour or perceptions rather than clinical infections.
Conventional epidemiological surveys remain \texttt{TRADITIONAL}.

\item[\texttt{NON\_TRADITIONAL\_ECONOMIC}] when a paper uses digitally captured or platform-mediated data describing the economic activity of individuals, groups, or organisations.
This category includes credit and debit card transaction data, open procurement/contracting data, and supply chain data.

\item[\texttt{SYNTHETIC}] when simulated data are presented as evidence and support for claims.
We do \emph{not} assign \texttt{SYNTHETIC} when simulations are used solely to illustrate model fit or calibration.
Examples include simulation-based evaluation of policy scenarios, model-based counterfactual projections, and sensitivity analyses where simulated outcomes support the main claim (versus a purely illustrative fit plot, for which we do not assign the label).
The distinction depends on whether what is produced is considered data at all, rather than on whether it is synthetic.

\item[\texttt{NO\_EMPIRICAL\_DATA}] when the paper relies on neither an empirical dataset nor simulated data as evidence.
Examples include a conceptual commentary on containment approaches without data and a theoretical derivation of a model without fitting or empirical evaluation.

\end{description}

\subsection{Paper type labelling}

\texttt{PAPER\_TYPE} classifies each document by the kind of scientific contribution it makes to epidemiological parameter estimation.
Epidemiological research integrates multiple tasks (collecting data, inferring parameters, and making forecasts) within a single workflow.
While usually focused around a main contribution, papers combine different types of contributions.
We handle this by distinguishing between a paper's \emph{primary contribution} (the main scientific claim or deliverable) and any \emph{secondary contributions} (supporting tasks that enable or accompany the main contribution).
We therefore assign a primary label and may assign additional secondary labels as needed.
The same label set is used for both primary and secondary contributions.

\begin{description}
\item[\texttt{EMPIRICAL}.]
The contribution is based on \emph{original empirical data collection}, where claims rely on direct measurement from a clearly defined study population or sample.
Examples: observational cohort studies, case--control studies, cross-sectional surveys, randomised trials, surveillance reports, or data papers describing new datasets.
\item[\texttt{INFERENCE}.]
  The contribution involves \emph{estimating parameters or reconstructing unobserved epidemic processes} by fitting explicit statistical or mechanistic models to observed data.
  Results depend on model structure and assumptions, not on direct measurement alone.
  Examples: papers estimating $R_0$ or generation time from case data, reconstructing transmission chains, performing phylodynamic inference, or estimating seroconversion rates from serological surveys.
    \item[\texttt{FORECAST}.]
    The contribution is \emph{prediction or scenario analysis}: forward projections of epidemic trends, comparisons of intervention scenarios, or counterfactual simulations.
    The outputs are projected future outcomes or policy comparisons under stated assumptions.
    Examples: epidemic projections for policy planning, intervention scenario comparisons, or real-time forecasting submissions.
    \item[\texttt{SYNTHESIS}.]
    The contribution is a \emph{systematic synthesis of existing evidence} using predefined search, selection, and appraisal criteria (e.g., meta-analyses or PRISMA-compliant systematic reviews).
    Examples: quantitative meta-analyses pooling estimates across studies, or systematic reviews synthesising evidence on transmission parameters or intervention effectiveness.
    \item[\texttt{METHODOLOGICAL}.]
    The contribution is the \emph{development, evaluation, or comparison of methods}: new estimators, inference frameworks, identifiability analyses, software tools, or theoretical results about epidemic processes.
    Examples: papers proposing new inference algorithms, proving identifiability results, developing software packages, or obtaining analytical results on epidemic thresholds.
    \item[\texttt{COMMENTARY}.]
    The contribution is \emph{discussion or interpretation} without presenting new empirical estimates, model-based inference, or systematic evidence synthesis.
    Examples include narrative literature reviews, perspective pieces on research priorities, and policy commentaries.
    \end{description}
We use \texttt{OTHER} when a paper does not fit any of the categories above.
In our analysis, we focus on the main contribution, making this a single-label classification task.
Secondary contributions are recorded but not used in the evaluation.

\section{Supplementary results and diagnostics}
\label{app:full_tables}
This appendix provides the detailed quantitative results that support \Cref{sec:results}.
Results for the fully local workflow use all \(214\) manually labelled papers, except in \Cref{app:tab:qwen_metrics_open,app:tab:per_label_f1_open,fig:ablation_open}, which restrict it to the \(122\) openly licensed papers (\Cref{app:licences}) so that it can be compared with the served models on the same papers.
On these \(122\) papers, every method is scored, so comparisons are paired.
Unless stated otherwise, values are mean \(\pm\) SD across \(R{=}5\) repeated runs.

\subsection{The fully local workflow}
\label{app:qwen_local}

We run the Paper Classifier with \texttt{qwen2.5:32b-instruct}, served locally through Ollama \citep{ollama}, with a retrieval index built locally with \texttt{BAAI/bge-m3}, so parsing, embedding, retrieval, and generation in the classification workflow require no remote service.
All other components are those of \Cref{sec:evaluation_setup}: the same GROBID parses, dense retrieval with \texttt{workflow\_top\_k}${=}15$, prompts, schema-constrained decoding, and \(R{=}5\) repeats.
No prompt reached Qwen's context window of $32{,}768$ tokens: the recorded prompts are at most $12{,}090$ tokens long on the evaluation set, $18{,}464$ in the corpus-scale run of \Cref{sec:full_corpus_distributions}, and $22{,}771$ in the whole-body condition of \Cref{sec:ablation}.
\Cref{app:tab:qwen_metrics_local} reports the full metric profile on all \(214\) papers.
For the single-label paper-type task, micro-averaged scores coincide with subset accuracy.
The macro-averaged scores are lower than the corresponding micro-averaged scores, showing sensitivity to uneven class support. \Cref{app:per_label_f1_all_configs} gives per-label F\textsubscript{1} on the $122$ openly licensed papers.
On data type, the majority floor has micro- and macro-averaged F\textsubscript{1} of $0.851$ and $0.138$ on the same $214$ papers, compared with $0.871$ and $0.338$ for the fully local workflow at $T{=}0$. The higher macro score reflects detection of some labels beyond the majority set on this sample, although several labels have too few examples for precise per-label estimates.

The local workflow was exactly repeatable at \(T{=}0\) in our experiments.
Across five independent executions, it returned identical label sets for every paper on all four axes, and the repeats also agree on the generated text, retrieved evidence, and recorded prompt and completion token counts.
Exact repeatability disappears at \(T{=}1\), without a consistent gain in accuracy.
Our runner sent one request at a time, so we computed each request in its own batch, removing what \citet{he2025nondeterminism} identify as the main source of run-to-run variation at $T{=}0$.
Exact repeatability is therefore a property of the deployment tested here rather than a general guarantee of self-hosting: serving concurrent requests, or changing kernels or numerical precision, can reintroduce variation even under greedy decoding.
The experiment shows that controlling sampling and serving end to end was sufficient to achieve exact repeatability on this hardware and software stack, which is useful for an audit trail that may need to be regenerated later.

\begingroup
\scriptsize
\renewcommand{\arraystretch}{0.92}
\setlength{\tabcolsep}{4pt}
\begin{table}[!htbp]
\centering
\caption{\textbf{Full agreement-metric profile of the fully local workflow.}}
\label{app:tab:qwen_metrics_local}
\begin{tabular}{lcc}
\toprule
\textbf{Metric} & $T{=}0.0$ & $T{=}1.0$ \\
\midrule
\addlinespace[2pt]
\multicolumn{3}{l}{\textbf{Geography}}\\
\addlinespace[2pt]
Jaccard           & $0.869_{\pm 0.000}$ & $0.869_{\pm 0.014}\sdpad$ \\
Micro precision   & $0.804_{\pm 0.000}$ & $0.8222_{\pm 0.0075}$ \\
Micro recall      & $0.891_{\pm 0.000}$ & $0.889_{\pm 0.014}\sdpad$ \\
Micro F$_1$       & $0.846_{\pm 0.000}$ & $0.8543_{\pm 0.0086}$ \\
Macro precision   & $0.651_{\pm 0.000}$ & $0.6762_{\pm 0.0089}$ \\
Macro recall      & $0.736_{\pm 0.000}$ & $0.731_{\pm 0.026}\sdpad$ \\
Macro F$_1$       & $0.682_{\pm 0.000}$ & $0.694_{\pm 0.016}\sdpad$ \\
Subset accuracy   & $0.827_{\pm 0.000}$ & $0.833_{\pm 0.013}\sdpad$ \\
\addlinespace[2pt]
\multicolumn{3}{l}{\textbf{Data access.}}\\
\addlinespace[2pt]
Jaccard           & $0.539_{\pm 0.000}$ & $0.516_{\pm 0.027}$ \\
Micro precision   & $0.574_{\pm 0.000}$ & $0.552_{\pm 0.023}$ \\
Micro recall      & $0.629_{\pm 0.000}$ & $0.611_{\pm 0.019}$ \\
Micro F$_1$       & $0.600_{\pm 0.000}$ & $0.580_{\pm 0.020}$ \\
Macro precision   & $0.435_{\pm 0.000}$ & $0.438_{\pm 0.016}$ \\
Macro recall      & $0.491_{\pm 0.000}$ & $0.512_{\pm 0.011}$ \\
Macro F$_1$       & $0.431_{\pm 0.000}$ & $0.442_{\pm 0.016}$ \\
Subset accuracy   & $0.364_{\pm 0.000}$ & $0.328_{\pm 0.044}$ \\
\addlinespace[2pt]
\multicolumn{3}{l}{\textbf{Data type}}\\
\addlinespace[2pt]
Jaccard           & $0.861_{\pm 0.000}$ & $0.8661_{\pm 0.0061}$ \\
Micro precision   & $0.897_{\pm 0.000}$ & $0.8971_{\pm 0.0086}$ \\
Micro recall      & $0.848_{\pm 0.000}$ & $0.8578_{\pm 0.0081}$ \\
Micro F$_1$       & $0.871_{\pm 0.000}$ & $0.8770_{\pm 0.0070}$ \\
Macro precision   & $0.389_{\pm 0.000}$ & $0.479_{\pm 0.034}\sdpad$ \\
Macro recall      & $0.375_{\pm 0.000}$ & $0.464_{\pm 0.060}\sdpad$ \\
Macro F$_1$       & $0.338_{\pm 0.000}$ & $0.390_{\pm 0.035}\sdpad$ \\
Subset accuracy   & $0.790_{\pm 0.000}$ & $0.7925_{\pm 0.0085}$ \\
\addlinespace[2pt]
\multicolumn{3}{l}{\textbf{Paper type}}\\
\addlinespace[2pt]
Jaccard           & $0.794_{\pm 0.000}$ & $0.775_{\pm 0.013}$ \\
Micro precision   & $0.794_{\pm 0.000}$ & $0.775_{\pm 0.013}$ \\
Micro recall      & $0.794_{\pm 0.000}$ & $0.775_{\pm 0.013}$ \\
Micro F$_1$       & $0.794_{\pm 0.000}$ & $0.775_{\pm 0.013}$ \\
Macro precision   & $0.740_{\pm 0.000}$ & $0.631_{\pm 0.033}$ \\
Macro recall      & $0.643_{\pm 0.000}$ & $0.514_{\pm 0.055}$ \\
Macro F$_1$       & $0.615_{\pm 0.000}$ & $0.520_{\pm 0.043}$ \\
Subset accuracy   & $0.794_{\pm 0.000}$ & $0.775_{\pm 0.013}$ \\
\bottomrule
\end{tabular}
\footnotetext{Mean with the standard deviation across $R{=}5$ repeated runs shown as a subscript, on all
$214$ evaluation papers, for \texttt{qwen2.5:32b-instruct} served by Ollama with a \texttt{BAAI/bge-m3}
index built on the same machine. For the single-label paper-type task, micro-averaged scores coincide with subset
accuracy. All $T{=}0$ standard deviations are exactly zero: the five repeats return identical label sets for every
paper.}
\end{table}
\endgroup

\subsection{Served backends on the openly licensed papers}
\label{app:aggregate_metrics_all_configs}
\label{app:open_weight_metrics}

On the \(122\) openly licensed papers, \Cref{tab:metrics_summary_open} gives the mean Jaccard of each served configuration with its bootstrap interval, \Cref{app:tab:all_metrics_all_configs_open} the full set of agreement metrics, and \Cref{tab:stability_all_configs_open} the run-to-run stability of every configuration.
The served configurations are \texttt{gemini-2.5-pro} and \texttt{gemini-2.5-flash} at \(T{=}0\) and \(T{=}1\); both retrieve with \texttt{gemini-embedding-001}.
Temperature has little systematic effect on mean Jaccard for the served models.

\begin{table}[!htbp]
\centering
\small
\setlength{\tabcolsep}{4pt}
\caption{\textbf{Label stability across repeated runs.}}
\label{tab:stability_all_configs_open}
\begin{tabular}{l c c c c | c c c c}
\toprule
& \multicolumn{4}{c}{\textbf{Qwen2.5-32B} (local)} & \multicolumn{2}{c}{\textbf{Gemini pro}} & \multicolumn{2}{c}{\textbf{Gemini flash}} \\
\cmidrule(lr){2-5}
& \multicolumn{2}{c}{\emph{fully local}} & \multicolumn{2}{c}{\emph{partially local}} & & & & \\
\cmidrule(lr){2-3}\cmidrule(lr){4-5}\cmidrule(lr){6-7}\cmidrule(lr){8-9}
\textbf{Task} & $T{=}0$ & $T{=}1$ & $T{=}0$ & $T{=}1$ & $T{=}0$ & $T{=}1$ & $T{=}0$ & $T{=}1$ \\
\midrule
Geography          & 100.0  & 78.7   & 100.0  & 79.5   & 96.7   & 94.3   & 93.4   & 83.6   \\
Data access.       & 100.0  & 17.2   & 100.0  & 21.3   & 60.7   & 48.4   & 66.4   & 33.6   \\
Data type          & 100.0  & 73.8   & 100.0  & 81.1   & 91.8   & 91.0   & 94.3   & 89.3   \\
Paper type         & 100.0  & 89.3   & 100.0  & 86.9   & 84.4   & 82.8   & 88.5   & 73.0   \\
\bottomrule
\end{tabular}
\footnotetext{Percentage of papers whose predicted label set was identical across five repeated runs under each configuration.
A prediction is considered stable if the full label set is unchanged across runs.
 The \textbf{Qwen2.5-32B} columns are the self-hosted backend of \Cref{app:qwen_local}: \emph{fully local} embeds the index locally with \texttt{BAAI/bge-m3}, while \emph{partially local} reuses the Gemini-embedded index. Both are perfectly stable at $T{=}0$, returning an identical label set for every paper across all five repeats. \textbf{Gemini pro} and \textbf{Gemini flash} are \texttt{gemini-2.5-pro} and \texttt{gemini-2.5-flash}.
 The majority floor is omitted here because it is deterministic.}
\end{table}

\begin{table}[!htbp]
\centering
\small
\setlength{\tabcolsep}{4pt}
\caption{\textbf{Performance and requirements of the baseline approaches.}}
\label{tab:baseline_comparison_open}
\begin{tabular}{@{}>{\raggedright\arraybackslash}p{3.9cm}@{\hskip 0.5em}cc@{\hskip 0.5em}cccc@{}}
\toprule
\textbf{Method} & \multicolumn{2}{c}{\textbf{Requirements}} & \multicolumn{4}{c}{\textbf{Jaccard by task}} \\
\cmidrule(lr){2-3}\cmidrule(lr){4-7}
 & \textbf{Body?} & \begin{tabular}[b]{@{}c@{}}\textbf{Task}\\\textbf{sup.}\end{tabular} & \begin{tabular}[b]{@{}c@{}}\textbf{Paper}\\\textbf{type}\end{tabular} & \begin{tabular}[b]{@{}c@{}}\textbf{Geo-}\\\textbf{graphy}\end{tabular} & \begin{tabular}[b]{@{}c@{}}\textbf{Data}\\\textbf{type}\end{tabular} & \begin{tabular}[b]{@{}c@{}}\textbf{Data}\\\textbf{access.}\end{tabular} \\
\midrule
    \addlinespace[2pt]\multicolumn{7}{l}{\textbf{Reference}} \\
    \quad Majority prior (floor) & No & Weak & 0.615 & 0.416 & 0.870 & 0.474 \\
    \midrule
    \addlinespace[2pt]\multicolumn{7}{l}{\textbf{Zero-shot (no training labels)}} \\
    \quad Prototype (TF-IDF) & No & None & 0.451 & 0.246 & 0.310 & 0.206 \\
    \quad Prototype (SPECTER) & No & None & 0.098 & 0.157 & 0.539 & 0.369 \\
    \quad Zero-shot NLI (DeBERTa) & No & None & 0.607 & 0.759 & 0.014 & 0.341 \\
    \midrule
    \addlinespace[2pt]\multicolumn{7}{l}{\textbf{Topic models (corpus-derived, label-guided)}} \\
    \quad Guided BERTopic & Yes & Weak & 0.590 & 0.597 & 0.870 & 0.474 \\
    \quad Semi-supervised BERTopic & Yes & Weak & 0.590 & 0.601 & 0.870 & 0.441 \\
    \midrule
    \addlinespace[2pt]\multicolumn{7}{l}{\textbf{Supervised (title + abstract)}} \\
    \quad TF-IDF + Linear SVM & No & Manual & 0.750 & 0.642 & 0.893 & 0.495 \\
    \quad SetFit (MPNet) & No & Manual & 0.770 & 0.642 & 0.895 & 0.552 \\
    \quad Prototype (TF-IDF) & No & Manual & 0.451 & 0.330 & 0.310 & 0.232 \\
    \quad Prototype (SPECTER) & No & Manual & 0.098 & 0.154 & 0.734 & 0.369 \\
    \midrule
    \addlinespace[2pt]\multicolumn{7}{l}{\textbf{Supervised (full body)}} \\
    \quad ModernBERT (MLP head) & Yes & Manual & 0.775 & 0.706 & 0.906 & 0.536 \\
    \quad \emph{Best supervised} (ceiling) & Yes & Manual & --- & --- & --- & --- \\
    \midrule
    \addlinespace[2pt]\multicolumn{7}{l}{\textbf{LLM, self-hosted (abstract-only ablation and full \toolname{})}} \\
    \quad Abstract only (\texttt{metadata\_llm}) & No & None & 0.680 & 0.844 & 0.768 & 0.154 \\
    \addlinespace[4pt]
    \quad \textbf{\toolname{}} (qwen2.5:32b-instruct) & Yes & None & \begin{tabular}[t]{@{}c@{}}$0.795$\\[2pt]{\footnotesize 0.71--0.85}\end{tabular} & \begin{tabular}[t]{@{}c@{}}$0.892$\\[2pt]{\footnotesize 0.83--0.93}\end{tabular} & \begin{tabular}[t]{@{}c@{}}$0.907$\\[2pt]{\footnotesize 0.86--0.94}\end{tabular} & \begin{tabular}[t]{@{}c@{}}$0.536$\\[2pt]{\footnotesize 0.46--0.60}\end{tabular} \\
    \midrule
    \addlinespace[2pt]\multicolumn{7}{l}{\textbf{LLM, served (abstract-only ablation and full \toolname{})}} \\
    \quad Abstract only (\texttt{metadata\_llm}) & No & None & 0.697 & 0.883 & 0.837 & 0.175 \\
    \addlinespace[4pt]
    \quad \textbf{\toolname{}} (gemini-2.5-flash) & Yes & None & \begin{tabular}[t]{@{}c@{}}$0.843$\\[2pt]{\footnotesize 0.78--0.89}\end{tabular} & \begin{tabular}[t]{@{}c@{}}$0.890$\\[2pt]{\footnotesize 0.84--0.93}\end{tabular} & \begin{tabular}[t]{@{}c@{}}$0.934$\\[2pt]{\footnotesize 0.90--0.96}\end{tabular} & \begin{tabular}[t]{@{}c@{}}$0.642$\\[2pt]{\footnotesize 0.57--0.71}\end{tabular} \\
    \addlinespace[4pt]
    \quad \textbf{\toolname{}} (gemini-2.5-pro) & Yes & None & \begin{tabular}[t]{@{}c@{}}$0.877$\\[2pt]{\footnotesize 0.82--0.92}\end{tabular} & \begin{tabular}[t]{@{}c@{}}$0.936$\\[2pt]{\footnotesize 0.89--0.97}\end{tabular} & \begin{tabular}[t]{@{}c@{}}$0.968$\\[2pt]{\footnotesize 0.94--0.98}\end{tabular} & \begin{tabular}[t]{@{}c@{}}$0.778$\\[2pt]{\footnotesize 0.72--0.83}\end{tabular} \\
\bottomrule
\end{tabular}
\footnotetext{Sample-averaged Jaccard similarity $J$ on the $122$ openly licensed evaluation papers.
\textbf{Body?} indicates whether parsed body text is required. \emph{No} generally means title and abstract,
while the majority floor uses no text. \textbf{Task sup.} records the task-specific input used to construct or
fit the method: \emph{None} uses no labelled papers, \emph{Weak} uses a class prior, seeds, or partial paper
labels, and \emph{Manual} uses a fully manually labelled training set. All scores use the same manual labels.
Prototype rows contrast uncalibrated similarity with threshold calibration. BERTopic uses the best topic
count and oracle cluster-to-label alignment. \emph{Best supervised} is the per-task maximum over the
supervised grid. Both are favourable descriptive upper bounds rather than single deployable configurations. The supervised grid is recorded only as a summary over all $214$ labelled papers, so its ceiling is not shown on this subset.
For \toolname{}, the mean is over $R{=}5$ runs at $T{=}0$, with its $95\%$ bootstrap interval over papers beneath it, as in \Cref{tab:metrics_summary_open}. Other rows are
deterministic evaluations or cross-validated estimates. Scores are rounded to three decimals. Paired tests
are reported in \Cref{tab:significance_open}. Full method specifications are given in \Cref{app:baselines}.}
\end{table}

\begin{table}[!htbp]
\centering
\small
\setlength{\tabcolsep}{4pt}
\caption{\textbf{Paired comparisons of \toolname{} (\texttt{gemini-2.5-flash}, $T{=}0$) with selected baselines.}}
\label{tab:significance_open}
\begin{tabular}{lcccc}
\toprule
\textbf{Baseline} & \textbf{Paper type} & \textbf{Geography} & \textbf{Data type} & \textbf{Data access.} \\
\midrule
    Zero-shot NLI & $+0.236$$^{*}$ & $+0.131$$^{*}$ & $+0.920$$^{*}$ & $+0.300$$^{*}$ \\
    TF-IDF + SVM & $+0.093$\phantom{$^{*}$} & $+0.248$$^{*}$ & $+0.040$\phantom{$^{*}$} & $+0.147$$^{*}$ \\
    TF-IDF + SVM (LOO) & $+0.109$$^{*}$ & $+0.235$$^{*}$ & $+0.020$\phantom{$^{*}$} & $+0.128$$^{*}$ \\
    TF-IDF + SVM (full body) & $+0.093$\phantom{$^{*}$} & $+0.231$$^{*}$ & $+0.012$\phantom{$^{*}$} & $+0.068$\phantom{$^{*}$} \\
    SetFit & $+0.072$\phantom{$^{*}$} & $+0.249$$^{*}$ & $+0.039$\phantom{$^{*}$} & $+0.090$\phantom{$^{*}$} \\
    ModernBERT (full body) & $+0.068$\phantom{$^{*}$} & $+0.184$$^{*}$ & $+0.028$\phantom{$^{*}$} & $+0.106$\phantom{$^{*}$} \\
    \texttt{metadata\_llm} & $+0.146$$^{*}$ & $+0.007$\phantom{$^{*}$} & $+0.096$$^{*}$ & $+0.467$$^{*}$ \\
    \midrule
    \addlinespace[2pt]\multicolumn{5}{l}{\textbf{Reference}} \\
    \quad Majority prior (floor) & $+0.228$$^{*}$ & $+0.474$$^{*}$ & $+0.064$$^{*}$ & $+0.167$$^{*}$ \\
\bottomrule
\end{tabular}
\footnotetext{Each cell is the mean per-paper Jaccard difference (\toolname{} minus the baseline) on the $N{=}122$ openly licensed evaluation papers.
A marker ($^{*}$) denotes a Holm-adjusted $p$-value below $0.05$ across eight comparisons per task.
All eight comparisons in the correction are shown.
The majority floor is shown for reference: it predicts the most common label set of the manual labels, which \toolname{} does not see, and its row tests whether $J$ exceeds what label skew alone gives.
It enters the Holm correction, which can only make the other comparisons more conservative, never less.
Tests use $B{=}10{,}000$ paired BCa resamples (\Cref{app:baselines:statistics}).
They hold the manual labels, predictions, and configurations fixed.
The topic-model and \emph{Best supervised} rows of \Cref{tab:baseline_comparison_open} are descriptive because their alignment or configuration was selected using the evaluation papers.
Prototype rows are reported as diagnostics and are not tested here.}
\end{table}

\begin{table}[!htbp]
\centering
\small
\setlength{\tabcolsep}{3pt}
\caption{\textbf{Accuracy of the served backends.}}
\label{tab:metrics_summary_open}
\begin{tabular}{l c | c c c c}
\toprule
& \textbf{Floor} & \multicolumn{2}{c}{\textbf{gemini-2.5-pro}} & \multicolumn{2}{c}{\textbf{gemini-2.5-flash}} \\
\cmidrule(lr){2-2}\cmidrule(lr){3-4}\cmidrule(lr){5-6}
\textbf{Task} & Majority & $T{=}0$ & $T{=}1$ & $T{=}0$ & $T{=}1$ \\
\midrule
Geography          & \textit{0.416} & $0.936_{\,0.89\text{--}0.97}$ & $0.936_{\,0.89\text{--}0.97}$ & $0.890_{\,0.84\text{--}0.93}$ & $0.883_{\,0.83\text{--}0.92}$ \\
Data access.       & \textit{0.474} & $0.778_{\,0.72\text{--}0.83}$ & $0.747_{\,0.69\text{--}0.80}$ & $0.642_{\,0.57\text{--}0.71}$ & $0.657_{\,0.60\text{--}0.71}$ \\
Data type          & \textit{0.870} & $0.968_{\,0.94\text{--}0.98}$ & $0.958_{\,0.92\text{--}0.98}$ & $0.934_{\,0.90\text{--}0.96}$ & $0.931_{\,0.89\text{--}0.96}$ \\
Paper type         & \textit{0.615} & $0.877_{\,0.82\text{--}0.92}$ & $0.885_{\,0.83\text{--}0.92}$ & $0.843_{\,0.78\text{--}0.89}$ & $0.856_{\,0.80\text{--}0.90}$ \\
\bottomrule
\end{tabular}
\footnotetext{Mean Jaccard similarity on the $122$ openly licensed evaluation papers over $R{=}5$ runs per configuration. The subscript is a $95\%$ bootstrap interval over papers (BCa, $B{=}10{,}000$), computed after averaging each paper's score over the runs: it shows how precisely these papers determine the score. Each interval describes one configuration; methods are compared with the paired tests of \Cref{app:baselines:statistics}. Run-to-run standard deviations are in \Cref{app:tab:all_metrics_all_configs_open}.
 The \emph{Floor} column (set off on the left) is the majority-class baseline, a single deterministic value included as a reference point for reading the scores.
 Paper type is a single-label task. The remaining tasks are multi-label.}
\end{table}

\FloatBarrier
\begingroup
\small
\setlength{\tabcolsep}{4pt}
\begin{longtable}{lcccc}
\caption{\textbf{Aggregate agreement metrics by task and configuration.}}\label{app:tab:all_metrics_all_configs_open}\\
\toprule
& \multicolumn{2}{c}{\textbf{gemini-2.5-pro}} & \multicolumn{2}{c}{\textbf{gemini-2.5-flash}} \\
\cmidrule(lr){2-3}\cmidrule(lr){4-5}
\textbf{Metric} & $T{=}0.0$ & $T{=}1.0$ & $T{=}0.0$ & $T{=}1.0$ \\
\midrule
\endfirsthead
\toprule
& \multicolumn{2}{c}{\textbf{gemini-2.5-pro}} & \multicolumn{2}{c}{\textbf{gemini-2.5-flash}} \\
\cmidrule(lr){2-3}\cmidrule(lr){4-5}
\textbf{Metric} & $T{=}0.0$ & $T{=}1.0$ & $T{=}0.0$ & $T{=}1.0$ \\
\midrule
\endhead
\midrule
\multicolumn{5}{r}{\small\itshape (continued on next page)}\\
\endfoot
\bottomrule
\endlastfoot

\addlinespace[2pt]
\multicolumn{5}{l}{\textbf{Geography}}\\
\addlinespace[2pt]
Jaccard          & $0.9355_{\pm 0.0015}$ & $0.9358_{\pm 0.0032}$ & $0.8903_{\pm 0.0092}$ & $0.883_{\pm 0.012}\sdpad$ \\
Micro precision  & $0.8840_{\pm 0.0089}$ & $0.8898_{\pm 0.0089}$ & $0.810_{\pm 0.013}\sdpad$ & $0.774_{\pm 0.029}\sdpad$ \\
Micro recall     & $0.964_{\pm 0.000}\sdpad$ & $0.964_{\pm 0.000}\sdpad$ & $0.9496_{\pm 0.0072}$ & $0.9496_{\pm 0.0051}$ \\
Micro F$_1$      & $0.9223_{\pm 0.0048}$ & $0.9254_{\pm 0.0048}$ & $0.874_{\pm 0.010}\sdpad$ & $0.852_{\pm 0.020}\sdpad$ \\
Macro precision  & $0.760_{\pm 0.010}\sdpad$ & $0.767_{\pm 0.010}\sdpad$ & $0.698_{\pm 0.012}\sdpad$ & $0.658_{\pm 0.028}\sdpad$ \\
Macro recall     & $0.814_{\pm 0.000}\sdpad$ & $0.814_{\pm 0.000}\sdpad$ & $0.783_{\pm 0.016}\sdpad$ & $0.783_{\pm 0.011}\sdpad$ \\
Macro F$_1$      & $0.7807_{\pm 0.0055}$ & $0.7847_{\pm 0.0055}$ & $0.717_{\pm 0.019}\sdpad$ & $0.691_{\pm 0.029}\sdpad$ \\
Subset accuracy  & $0.9082_{\pm 0.0037}$ & $0.9066_{\pm 0.0045}$ & $0.834_{\pm 0.011}\sdpad$ & $0.828_{\pm 0.016}\sdpad$ \\

\addlinespace[4pt]
\multicolumn{5}{l}{\textbf{Data Accessibility}}\\
\addlinespace[2pt]
Jaccard          & $0.778_{\pm 0.028}$ & $0.747_{\pm 0.017}$ & $0.642_{\pm 0.021}\sdpad$ & $0.657_{\pm 0.012}\sdpad$ \\
Micro precision  & $0.808_{\pm 0.022}$ & $0.778_{\pm 0.017}$ & $0.661_{\pm 0.017}\sdpad$ & $0.671_{\pm 0.014}\sdpad$ \\
Micro recall     & $0.800_{\pm 0.036}$ & $0.777_{\pm 0.018}$ & $0.716_{\pm 0.017}\sdpad$ & $0.741_{\pm 0.028}\sdpad$ \\
Micro F$_1$      & $0.804_{\pm 0.029}$ & $0.778_{\pm 0.017}$ & $0.687_{\pm 0.016}\sdpad$ & $0.704_{\pm 0.020}\sdpad$ \\
Macro precision  & $0.721_{\pm 0.022}$ & $0.690_{\pm 0.029}$ & $0.617_{\pm 0.022}\sdpad$ & $0.607_{\pm 0.024}\sdpad$ \\
Macro recall     & $0.743_{\pm 0.035}$ & $0.728_{\pm 0.030}$ & $0.7526_{\pm 0.0079}$ & $0.757_{\pm 0.036}\sdpad$ \\
Macro F$_1$      & $0.716_{\pm 0.024}$ & $0.691_{\pm 0.025}$ & $0.653_{\pm 0.014}\sdpad$ & $0.649_{\pm 0.017}\sdpad$ \\
Subset accuracy  & $0.682_{\pm 0.024}$ & $0.643_{\pm 0.025}$ & $0.507_{\pm 0.022}\sdpad$ & $0.5000_{\pm 0.0058}$ \\

\addlinespace[4pt]
\multicolumn{5}{l}{\textbf{Data Type}}\\
\addlinespace[2pt]
Jaccard          & $0.968_{\pm 0.011}\sdpad$ & $0.9582_{\pm 0.0039}$ & $0.9337_{\pm 0.0059}$ & $0.931_{\pm 0.010}\sdpad$ \\
Micro precision  & $0.963_{\pm 0.010}\sdpad$ & $0.9523_{\pm 0.0054}$ & $0.9059_{\pm 0.0072}$ & $0.910_{\pm 0.014}\sdpad$ \\
Micro recall     & $0.9837_{\pm 0.0091}$ & $0.9769_{\pm 0.0037}$ & $0.9823_{\pm 0.0037}$ & $0.9741_{\pm 0.0089}$ \\
Micro F$_1$      & $0.9731_{\pm 0.0095}$ & $0.9644_{\pm 0.0045}$ & $0.9426_{\pm 0.0053}$ & $0.9409_{\pm 0.0074}$ \\
Macro precision  & $0.687_{\pm 0.067}\sdpad$ & $0.628_{\pm 0.096}\sdpad$ & $0.585_{\pm 0.081}\sdpad$ & $0.516_{\pm 0.094}\sdpad$ \\
Macro recall     & $0.729_{\pm 0.090}\sdpad$ & $0.670_{\pm 0.078}\sdpad$ & $0.684_{\pm 0.038}\sdpad$ & $0.635_{\pm 0.089}\sdpad$ \\
Macro F$_1$      & $0.675_{\pm 0.084}\sdpad$ & $0.618_{\pm 0.080}\sdpad$ & $0.583_{\pm 0.053}\sdpad$ & $0.524_{\pm 0.083}\sdpad$ \\
Subset accuracy  & $0.943_{\pm 0.018}\sdpad$ & $0.9279_{\pm 0.0069}$ & $0.877_{\pm 0.010}\sdpad$ & $0.875_{\pm 0.017}\sdpad$ \\

\addlinespace[4pt]
\multicolumn{5}{l}{\textbf{Paper Type}}\\
\addlinespace[2pt]
Jaccard          & $0.877_{\pm 0.016}$ & $0.885_{\pm 0.020}$ & $0.8426_{\pm 0.0069}$ & $0.856_{\pm 0.015}$ \\
Micro precision  & $0.877_{\pm 0.016}$ & $0.885_{\pm 0.020}$ & $0.8426_{\pm 0.0069}$ & $0.856_{\pm 0.015}$ \\
Micro recall     & $0.877_{\pm 0.016}$ & $0.885_{\pm 0.020}$ & $0.8426_{\pm 0.0069}$ & $0.856_{\pm 0.015}$ \\
Micro F$_1$      & $0.877_{\pm 0.016}$ & $0.885_{\pm 0.020}$ & $0.8426_{\pm 0.0069}$ & $0.856_{\pm 0.015}$ \\
Macro precision  & $0.75_{\pm 0.11}\sdpad$ & $0.84_{\pm 0.11}\sdpad$ & $0.644_{\pm 0.012}\sdpad$ & $0.669_{\pm 0.042}$ \\
Macro recall     & $0.790_{\pm 0.065}$ & $0.891_{\pm 0.090}$ & $0.7247_{\pm 0.0065}$ & $0.745_{\pm 0.098}$ \\
Macro F$_1$      & $0.763_{\pm 0.081}$ & $0.86_{\pm 0.10}\sdpad$ & $0.6404_{\pm 0.0080}$ & $0.667_{\pm 0.061}$ \\
Subset accuracy  & $0.877_{\pm 0.016}$ & $0.885_{\pm 0.020}$ & $0.8426_{\pm 0.0069}$ & $0.856_{\pm 0.015}$ \\
\end{longtable}
{\def\baselinestretch{1}\tablefootnotefont\noindent
Computed on the 122 openly licensed evaluation papers.
Values are mean with the standard deviation as a subscript, across $R{=}5$ repeated runs.\par}
\endgroup

\Cref{app:tab:qwen_metrics_open} adds a \emph{partially local} configuration, in which the local generator retrieves from the Gemini-embedded index used by the served models.
Because the two indices use different encoders, they can retrieve different evidence, so we treat their comparison as a comparison between deployment setups rather than a measure of embedding quality.
On these papers, the fully local configuration is lower than the partially local one in seven of the eight task--temperature pairs, by at most \(0.025\), so removing the remote embedding dependency costs little here.

\begingroup
\scriptsize
\renewcommand{\arraystretch}{0.92}
\setlength{\tabcolsep}{4pt}
\begin{table}[!htbp]
\centering
\caption{\textbf{Full agreement-metric profile of the self-hosted open-weight backend.}}
\label{app:tab:qwen_metrics_open}
\begin{tabular}{lcccc}
\toprule
& \multicolumn{4}{c}{\textbf{Qwen2.5-32B} (open weight, self-hosted generation)} \\
\cmidrule(lr){2-5}
& \multicolumn{2}{c}{\emph{fully local} (bge-m3)} & \multicolumn{2}{c}{\emph{partially local} (Gemini index)} \\
\cmidrule(lr){2-3}\cmidrule(lr){4-5}
\textbf{Metric} & $T{=}0.0$ & $T{=}1.0$ & $T{=}0.0$ & $T{=}1.0$ \\
\midrule
\addlinespace[2pt]
\multicolumn{5}{l}{\textbf{Geography}}\\
\addlinespace[2pt]
Jaccard           & $0.892_{\pm 0.000}$ & $0.894_{\pm 0.017}$ & $0.890_{\pm 0.000}$ & $0.9030_{\pm 0.0093}$ \\
Micro precision   & $0.827_{\pm 0.000}$ & $0.852_{\pm 0.014}$ & $0.815_{\pm 0.000}$ & $0.860_{\pm 0.013}\sdpad$ \\
Micro recall      & $0.928_{\pm 0.000}$ & $0.928_{\pm 0.022}$ & $0.921_{\pm 0.000}$ & $0.919_{\pm 0.021}\sdpad$ \\
Micro F$_1$       & $0.875_{\pm 0.000}$ & $0.888_{\pm 0.014}$ & $0.865_{\pm 0.000}$ & $0.8887_{\pm 0.0047}$ \\
Macro precision   & $0.794_{\pm 0.000}$ & $0.829_{\pm 0.018}$ & $0.792_{\pm 0.000}$ & $0.8392_{\pm 0.0098}$ \\
Macro recall      & $0.886_{\pm 0.000}$ & $0.882_{\pm 0.045}$ & $0.873_{\pm 0.000}$ & $0.861_{\pm 0.040}\sdpad$ \\
Macro F$_1$       & $0.824_{\pm 0.000}$ & $0.839_{\pm 0.033}$ & $0.808_{\pm 0.000}$ & $0.835_{\pm 0.018}\sdpad$ \\
Subset accuracy   & $0.844_{\pm 0.000}$ & $0.852_{\pm 0.016}$ & $0.844_{\pm 0.000}$ & $0.8623_{\pm 0.0090}$ \\
\addlinespace[2pt]
\multicolumn{5}{l}{\textbf{Data access.}}\\
\addlinespace[2pt]
Jaccard           & $0.536_{\pm 0.000}$ & $0.500_{\pm 0.040}$ & $0.553_{\pm 0.000}$ & $0.525_{\pm 0.022}$ \\
Micro precision   & $0.564_{\pm 0.000}$ & $0.538_{\pm 0.028}$ & $0.569_{\pm 0.000}$ & $0.559_{\pm 0.014}$ \\
Micro recall      & $0.630_{\pm 0.000}$ & $0.614_{\pm 0.026}$ & $0.660_{\pm 0.000}$ & $0.632_{\pm 0.021}$ \\
Micro F$_1$       & $0.595_{\pm 0.000}$ & $0.573_{\pm 0.026}$ & $0.611_{\pm 0.000}$ & $0.593_{\pm 0.012}$ \\
Macro precision   & $0.434_{\pm 0.000}$ & $0.422_{\pm 0.019}$ & $0.439_{\pm 0.000}$ & $0.454_{\pm 0.044}$ \\
Macro recall      & $0.511_{\pm 0.000}$ & $0.513_{\pm 0.020}$ & $0.519_{\pm 0.000}$ & $0.533_{\pm 0.049}$ \\
Macro F$_1$       & $0.447_{\pm 0.000}$ & $0.441_{\pm 0.019}$ & $0.438_{\pm 0.000}$ & $0.458_{\pm 0.036}$ \\
Subset accuracy   & $0.352_{\pm 0.000}$ & $0.290_{\pm 0.061}$ & $0.377_{\pm 0.000}$ & $0.320_{\pm 0.038}$ \\
\addlinespace[2pt]
\multicolumn{5}{l}{\textbf{Data type}}\\
\addlinespace[2pt]
Jaccard           & $0.907_{\pm 0.000}$ & $0.9051_{\pm 0.0089}$ & $0.929_{\pm 0.000}$ & $0.927_{\pm 0.010}\sdpad$ \\
Micro precision   & $0.930_{\pm 0.000}$ & $0.921_{\pm 0.011}\sdpad$ & $0.939_{\pm 0.000}$ & $0.945_{\pm 0.011}\sdpad$ \\
Micro recall      & $0.905_{\pm 0.000}$ & $0.909_{\pm 0.010}\sdpad$ & $0.939_{\pm 0.000}$ & $0.928_{\pm 0.010}\sdpad$ \\
Micro F$_1$       & $0.917_{\pm 0.000}$ & $0.9151_{\pm 0.0088}$ & $0.939_{\pm 0.000}$ & $0.9362_{\pm 0.0089}$ \\
Macro precision   & $0.371_{\pm 0.000}$ & $0.413_{\pm 0.053}\sdpad$ & $0.414_{\pm 0.000}$ & $0.401_{\pm 0.021}\sdpad$ \\
Macro recall      & $0.386_{\pm 0.000}$ & $0.484_{\pm 0.066}\sdpad$ & $0.418_{\pm 0.000}$ & $0.396_{\pm 0.028}\sdpad$ \\
Macro F$_1$       & $0.372_{\pm 0.000}$ & $0.418_{\pm 0.049}\sdpad$ & $0.416_{\pm 0.000}$ & $0.395_{\pm 0.024}\sdpad$ \\
Subset accuracy   & $0.844_{\pm 0.000}$ & $0.839_{\pm 0.019}\sdpad$ & $0.877_{\pm 0.000}$ & $0.869_{\pm 0.015}\sdpad$ \\
\addlinespace[2pt]
\multicolumn{5}{l}{\textbf{Paper type}}\\
\addlinespace[2pt]
Jaccard           & $0.795_{\pm 0.000}$ & $0.793_{\pm 0.016}$ & $0.811_{\pm 0.000}$ & $0.800_{\pm 0.016}$ \\
Micro precision   & $0.795_{\pm 0.000}$ & $0.793_{\pm 0.016}$ & $0.811_{\pm 0.000}$ & $0.800_{\pm 0.016}$ \\
Micro recall      & $0.795_{\pm 0.000}$ & $0.793_{\pm 0.016}$ & $0.811_{\pm 0.000}$ & $0.800_{\pm 0.016}$ \\
Micro F$_1$       & $0.795_{\pm 0.000}$ & $0.793_{\pm 0.016}$ & $0.811_{\pm 0.000}$ & $0.800_{\pm 0.016}$ \\
Macro precision   & $0.783_{\pm 0.000}$ & $0.67_{\pm 0.10}\sdpad$ & $0.724_{\pm 0.000}$ & $0.615_{\pm 0.024}$ \\
Macro recall      & $0.790_{\pm 0.000}$ & $0.677_{\pm 0.092}$ & $0.830_{\pm 0.000}$ & $0.711_{\pm 0.015}$ \\
Macro F$_1$       & $0.724_{\pm 0.000}$ & $0.600_{\pm 0.094}$ & $0.715_{\pm 0.000}$ & $0.611_{\pm 0.022}$ \\
Subset accuracy   & $0.795_{\pm 0.000}$ & $0.793_{\pm 0.016}$ & $0.811_{\pm 0.000}$ & $0.800_{\pm 0.016}$ \\
\bottomrule
\end{tabular}
\footnotetext{Mean with the standard deviation across $R{=}5$ repeated runs shown as a subscript, on the
$122$ openly licensed evaluation papers. All four
columns generate with \texttt{qwen2.5:32b-instruct} run locally through Ollama under the
identical workflow, and use schema-constrained decoding. The \emph{fully local} pair embeds
the retrieval index with \texttt{BAAI/bge-m3} on the same machine, so the pipeline makes no
remote call at any stage. The \emph{partially local} pair reuses the Gemini-embedded index
shared with the served-backend runs. Within each pair the columns differ only in sampling
temperature. For the single-label paper-type task, micro-averaged scores coincide with subset
accuracy. All $T{=}0$ standard deviations are exactly zero: the five repeats return identical
label sets for every paper.}
\end{table}
\endgroup

\begin{figure}[!htbp]
    \centering
    \includegraphics[width=\linewidth]{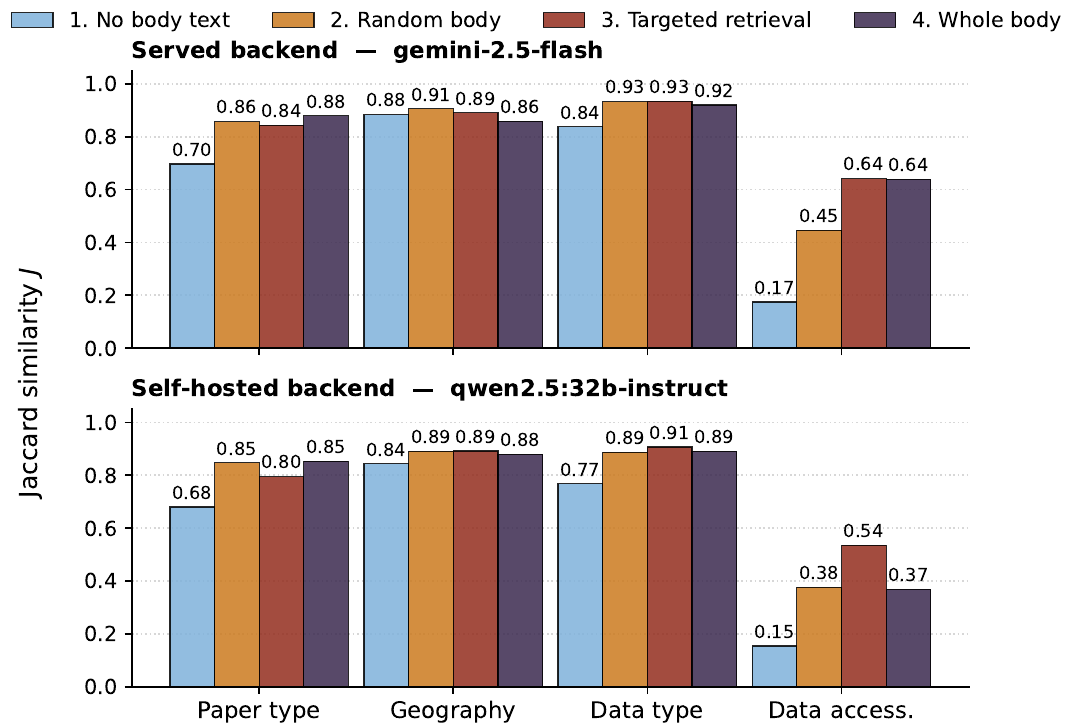}
    \caption{\textbf{Ablation results on the openly licensed papers.}
    The upper panel uses the served \texttt{gemini-2.5-flash} model and the lower panel the self-hosted \texttt{qwen2.5:32b-instruct}, both at $T{=}0$ on the $122$ papers selected by the licence screen of \Cref{app:licences}.
    The conditions are those of \Cref{fig:ablation}.}
    \label{fig:ablation_open}
\end{figure}

\Cref{fig:ablation_open} repeats the ablation of \Cref{sec:ablation} with \texttt{gemini-2.5-flash} and the local workflow on the same \(122\) papers, and \Cref{tab:baseline_comparison_open,tab:significance_open} repeat the comparison with the baselines.

\subsection{Per-label \texorpdfstring{F\textsubscript{1}}{F1}}
\label{app:per_label_f1_all_configs}

\Cref{app:tab:per_label_f1_open} gives the F\textsubscript{1} of each label at $T{=}0$ on the openly licensed papers, with the number of papers that carry it.
On data type, the Jaccard similarity is dominated by \emph{Traditional} ($118$ of the $122$ papers), which every backend labels with an F\textsubscript{1} of at least $0.974$, whereas \emph{Simulated} ranges from $0.829$ for the local workflow to $0.979$ for Pro, and the rarer labels occur in at most two papers.
On data accessibility, no backend exceeds $0.42$ on \emph{Not stated}.

\begin{table}[!htbp]
\centering
\setlength{\tabcolsep}{3pt}
\caption{\textbf{Per-label F$_1$ of \toolname{} against the manual labels.}}
\label{app:tab:per_label_f1_open}
\begin{tabular}{@{}lrccc@{}}
\toprule
\textbf{Label} & $n$ & \textbf{Pro} & \textbf{Flash} & \textbf{Qwen (local)} \\
\midrule
\addlinespace[2pt]
\multicolumn{5}{l}{\textbf{Geography}}\\
\addlinespace[2pt]
Asia & 55 & $0.9752_{\pm 0.0039}$ & $0.9532_{\pm 0.0045}$ & $0.930_{\pm 0.000}$ \\
Africa & 33 & $0.9735_{\pm 0.0065}$ & $0.9402_{\pm 0.0059}$ & $0.925_{\pm 0.000}$ \\
Europe & 18 & $0.8674_{\pm 0.0097}$ & $0.802_{\pm 0.010}\sdpad$ & $0.818_{\pm 0.000}$ \\
North America & 11 & $0.693_{\pm 0.019}\sdpad$ & $0.631_{\pm 0.034}\sdpad$ & $0.750_{\pm 0.000}$ \\
Irrelevant & 8 & $0.857_{\pm 0.000}\sdpad$ & $0.66_{\pm 0.11}\sdpad\sdpad$ & $0.769_{\pm 0.000}$ \\
Oceania & 7 & $0.947_{\pm 0.030}\sdpad$ & $0.960_{\pm 0.037}\sdpad$ & $0.750_{\pm 0.000}$ \\
South America & 7 & $0.933_{\pm 0.000}\sdpad$ & $0.788_{\pm 0.037}\sdpad$ & $0.824_{\pm 0.000}$ \\
\addlinespace[2pt]
\multicolumn{5}{l}{\textbf{Data accessibility}}\\
\addlinespace[2pt]
Reported & 72 & $0.866_{\pm 0.020}$ & $0.733_{\pm 0.013}$ & $0.730_{\pm 0.000}$ \\
Referenced & 41 & $0.812_{\pm 0.042}$ & $0.713_{\pm 0.046}$ & $0.511_{\pm 0.000}$ \\
Open & 32 & $0.818_{\pm 0.040}$ & $0.754_{\pm 0.014}$ & $0.667_{\pm 0.000}$ \\
Upon request & 7 & $0.808_{\pm 0.035}$ & $0.680_{\pm 0.062}$ & $0.308_{\pm 0.000}$ \\
Not stated & 7 & $0.418_{\pm 0.081}$ & $0.286_{\pm 0.037}$ & $0.267_{\pm 0.000}$ \\
Closed & 3 & $0.571_{\pm 0.000}$ & $0.750_{\pm 0.000}$ & $0.200_{\pm 0.000}$ \\
\addlinespace[2pt]
\multicolumn{5}{l}{\textbf{Data type}}\\
\addlinespace[2pt]
Traditional & 118 & $0.9941_{\pm 0.0038}$ & $0.9875_{\pm 0.0029}$ & $0.974_{\pm 0.000}$ \\
Simulated & 23 & $0.979_{\pm 0.000}\sdpad$ & $0.927_{\pm 0.011}\sdpad$ & $0.829_{\pm 0.000}$ \\
Non-trad.\ sentiment & 2 & $0.20_{\pm 0.45}\sdpad\sdpad$ & $0.500_{\pm 0.000}\sdpad$ & $0.800_{\pm 0.000}$ \\
No empirical data & 2 & $0.73_{\pm 0.15}\sdpad\sdpad$ & $0.40_{\pm 0.37}\sdpad\sdpad$ & $0.000_{\pm 0.000}$ \\
Non-trad.\ economic & 1 & $0.93_{\pm 0.15}\sdpad\sdpad$ & $1.000_{\pm 0.000}\sdpad$ & $0.000_{\pm 0.000}$ \\
Non-trad.\ mobility & 1 & $0.61_{\pm 0.24}\sdpad\sdpad$ & $0.264_{\pm 0.020}\sdpad$ & $0.000_{\pm 0.000}$ \\
\addlinespace[2pt]
\multicolumn{5}{l}{\textbf{Paper type}}\\
\addlinespace[2pt]
Empirical & 75 & $0.905_{\pm 0.013}$ & $0.9227_{\pm 0.0078}$ & $0.889_{\pm 0.000}$ \\
Inference & 34 & $0.825_{\pm 0.020}$ & $0.780_{\pm 0.018}\sdpad$ & $0.623_{\pm 0.000}$ \\
Forecast & 9 & $0.935_{\pm 0.047}$ & $0.775_{\pm 0.035}\sdpad$ & $0.364_{\pm 0.000}$ \\
Methodological & 2 & $0.71_{\pm 0.13}\sdpad$ & $0.365_{\pm 0.024}\sdpad$ & $0.800_{\pm 0.000}$ \\
Commentary & 1 & $0.20_{\pm 0.45}\sdpad$ & $0.000_{\pm 0.000}\sdpad$ & $0.667_{\pm 0.000}$ \\
Synthesis & 1 & $1.000_{\pm 0.000}$ & $1.000_{\pm 0.000}\sdpad$ & $1.000_{\pm 0.000}$ \\
\bottomrule
\end{tabular}
\footnotetext{F$_1$ of each label against the manual labels on the $122$ openly licensed evaluation papers at $T{=}0$: the mean over five runs, with the standard deviation as a subscript. $n$ is the number of papers that carry the label in the manual labels. Labels that never occur there are not shown. \textbf{Pro} and \textbf{Flash} are \texttt{gemini-2.5-pro} and \texttt{gemini-2.5-flash}, and \textbf{Qwen (local)} is the fully local \texttt{qwen2.5:32b-instruct} configuration, whose repeats are identical. For labels carried by few papers, F$_1$ rests on a handful of decisions.}
\end{table}

\subsection{Corpus-scale classification}
\label{app:qwen_local_corpus}

\begin{figure}[!htbp]
  \centering
  \includegraphics[width=\textwidth]{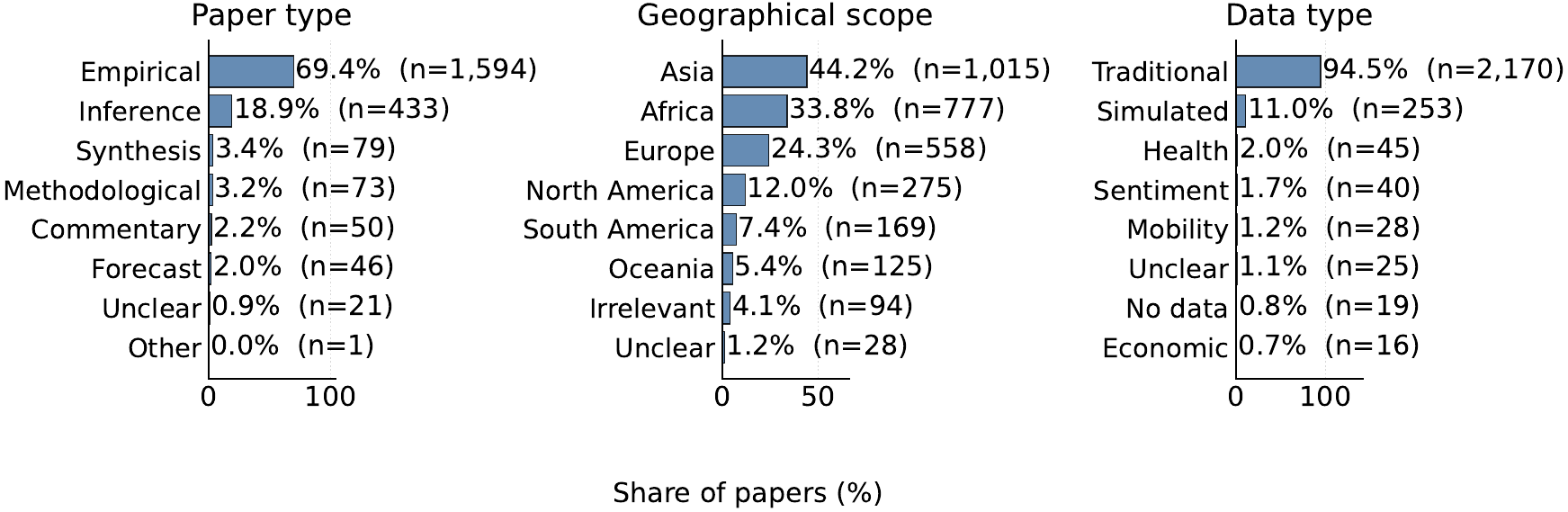}
  \caption{\textbf{Label distributions across all \numpaper{} corpus papers.}
  Panels show paper type, geography, and data type as labelled by the fully local workflow (\texttt{qwen2.5:32b-instruct} and a \texttt{BAAI/bge-m3} index) at \(T{=}0\).
  Bar lengths show percentages, and the adjacent numbers show paper counts.
  Paper type is single-label, whereas geography and data type are multi-label, so their percentages can sum to more than \(100\%\).}
  \label{fig:corpus_scale_distributions_local}
\end{figure}

The corpus-level analyses of \Cref{sec:full_corpus_distributions,sec:provenance} use one fully local run over all \numpaper{} papers at \(T{=}0\) on the three axes used there.
Once the corpus had been assembled, no stage of this classification pass contacted a remote service.
\Cref{fig:corpus_scale_distributions_local} shows the resulting distributions.

The full corpus has no manual labels, but the labelled sample is drawn from the same corpus and shows how the model's shares relate to the annotator's.
On that sample, the manual-label share of \emph{Empirical} papers is \(59.3\%\), against \(70.1\%\) in the local model's corpus-scale run, so the corpus share of \(69.4\%\) is likely an overestimate of roughly the same size.
The same model detects model-based and simulated-data papers less often than the annotator (\Cref{app:provenance}).
The labelled sample therefore helps interpret the direction of aggregate shares, but only the evaluation-set accuracy metrics support model comparisons at paper level.

\subsection{Stability and error diagnostics}
\label{app:error_diagnostics}
\label{app:containment}

Aggregate scores alone do not show whether different models perform poorly on the same papers.
For a fixed task and temperature, we define the \emph{ever-wrong set} \(E\) for a model as the papers for which at least one of the \(R{=}5\) runs differs from the complete manual label set.
This diagnostic is conditional on five repeats: with more repeats, a paper has more opportunities to enter \(E\).
We quantify the overlap and containment of the ever-wrong sets for \texttt{gemini-2.5-pro} and \texttt{gemini-2.5-flash} on the \(122\) openly licensed papers.

\begin{table}[!htbp]
\centering
\small
\setlength{\tabcolsep}{4pt}
\caption{\textbf{Containment of ever-wrong sets at $T{=}0.0$.}}
\label{tab:containment_t0_open}
\begin{tabular}{lrrrcc}
\toprule
Task & $|E_{\text{pro}}|$ & $|E_{\text{flash}}|$ & $|E_{\cap}|$ &
$|E_{\cap}|/|E_{\text{pro}}|$ & $|E_{\cap}|/|E_{\text{flash}}|$ \\
\midrule
Geography          & 12   & 23   & 11   & 91.7\%   & 47.8\%   \\
Data type          & 11   & 18   & 9    & 81.8\%   & 50.0\%   \\
Data accessibility & 57   & 72   & 46   & 80.7\%   & 63.9\%   \\
Paper type         & 24   & 26   & 11   & 45.8\%   & 42.3\%   \\
\bottomrule
\end{tabular}
\footnotetext{For each task, $E$ is the set of papers that are incorrectly labelled in at least one of the $R{=}5$ runs. For multi-label tasks, incorrect means that the predicted label set is not an exact match to the manual label set.
$|E_{\text{pro}}|$ and $|E_{\text{flash}}|$ report ever-wrong set sizes for the two backends.
$|E_{\cap}|$ reports overlap. The last two columns divide it by the corresponding backend's ever-wrong set size.}
\end{table}

\begin{table}[!htbp]
\centering
\small
\setlength{\tabcolsep}{4pt}
\caption{\textbf{Containment of ever-wrong sets at $T{=}1.0$.}}
\label{tab:containment_t1_open}
\begin{tabular}{lrrrcc}
\toprule
Task & $|E_{\text{pro}}|$ & $|E_{\text{flash}}|$ & $|E_{\cap}|$ &
$|E_{\cap}|/|E_{\text{pro}}|$ & $|E_{\cap}|/|E_{\text{flash}}|$ \\
\midrule
Geography          & 14   & 30   & 13   & 92.9\%   & 43.3\%   \\
Data type          & 15   & 23   & 15   & 100.0\%  & 65.2\%   \\
Data accessibility & 71   & 93   & 64   & 90.1\%   & 68.8\%   \\
Paper type         & 24   & 35   & 16   & 66.7\%   & 45.7\%   \\
\bottomrule
\end{tabular}
\footnotetext{Definitions as in \Cref{tab:containment_t0_open}.}
\end{table}

\Cref{tab:containment_t0_open,tab:containment_t1_open} show that the ever-wrong sets overlap substantially.
For \emph{geography}, \emph{data type}, and \emph{data accessibility}, most papers that are ever misclassified by Pro also appear in Flash's ever-wrong set.
The two models therefore tend to share a subset of difficult papers on these axes.
Containment is substantially lower for \emph{paper type}, indicating that disagreement on this task is more model-specific.
These patterns motivate using persistent errors or cross-model disagreement as additional signals for identifying papers that merit targeted retrieval, prompt refinement, or manual inspection.

\section{Comparison baselines}
\label{app:baselines}
\newif\ifshowaudit\showaudittrue

This appendix details the methods, configurations, and statistical comparisons behind the baseline results in \Cref{sec:baselines_summary,tab:baseline_comparison_local}.
The families vary fundamentally in what they require and what they measure, so they are read here not as a homogeneous leaderboard but as a map of what each kind of method can and cannot do on this task.

The bottom line is this: without labelled training data, the fully local \toolname{} leads the tested baselines on geography and paper type and is close to the supervised ones on data type and data accessibility, where served models lead them (\Cref{app:aggregate_metrics_all_configs}).
Body text matters most where the relevant evidence lies outside the title and abstract: the gain over the abstract-only LLM with the same model is largest on data accessibility as well as paper type, and targeted retrieval matters most for data accessibility (\Cref{sec:ablation}).
On geography, whose signal is largely in the title and abstract, that gain is smaller, although the margins over the supervised baselines are the largest of any axis (\Cref{tab:significance_local}).

\subsection{The design space and the methods we compare}
\label{app:baselines:taxonomy}

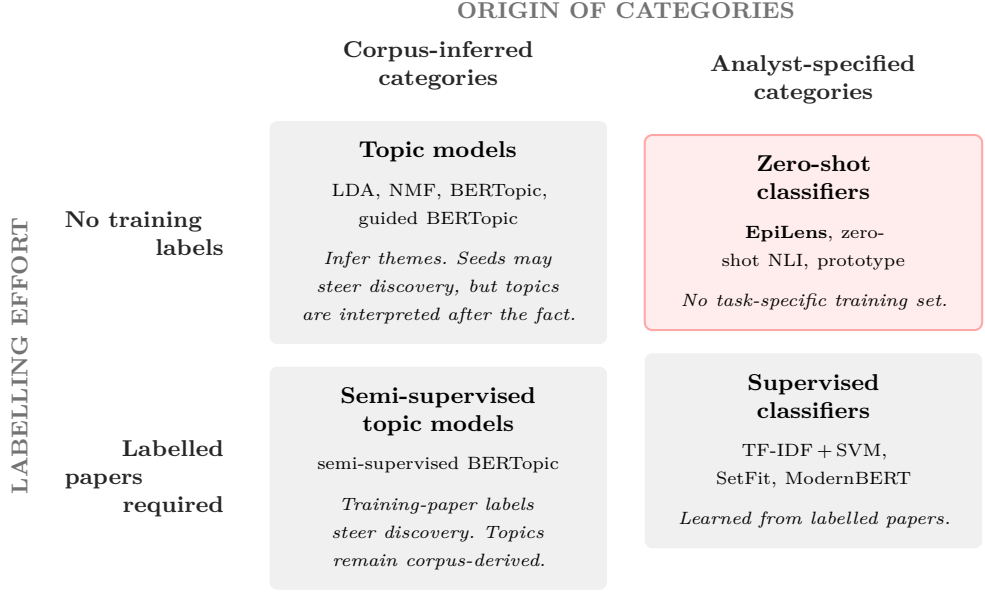
\begin{figure}[!htbp]
  \centering
  \resizebox{\columnwidth}{!}{%
\begin{tikzpicture}[
    node distance=0.3cm and 0.5cm, %
    font=\small,
    cell/.style={
      text width=4.1cm,
      minimum height=2.2cm,
      align=center,
      inner sep=8pt,
      rounded corners=3pt,
      fill=black!6,
      draw=none
    },
    highlight/.style={
      text width=4.1cm,
      minimum height=2.2cm,
      align=center,
      inner sep=8pt,
      rounded corners=3pt,
      fill=red!7,
      draw=red!35,
      line width=0.8pt
    },
    colhead/.style={
      text width=4.1cm,
      align=center,
      font=\small\bfseries,
      text=black!85
    },
    rowhead/.style={
      text width=2.2cm,
      align=right,
      font=\small\bfseries,
      text=black!85
    },
    axistitle/.style={
      font=\small\bfseries\color{black!55},
      align=center
    }
  ]

  \node[cell] (topics) {
    \textbf{Topic models}\\[4pt]
    {\footnotesize LDA, NMF, BERTopic, guided BERTopic}\\[5pt]
    {\footnotesize\itshape Infer themes. Seeds may steer discovery, but topics are interpreted after the fact.}
  };

  \node[highlight] (epilens) [right=of topics] {
    \textbf{Zero-shot}\\\textbf{classifiers}\\[4pt]
    {\footnotesize \textbf{\toolname{}}, zero-shot NLI, prototype}\\[5pt]
    {\footnotesize\itshape No task-specific training set.}
  };

  \node[cell] (guidedTopics) [below=of topics] {
    \textbf{Semi-supervised}\\\textbf{topic models}\\[4pt]
    {\footnotesize semi-supervised BERTopic}\\[5pt]
    {\footnotesize\itshape Training-paper labels steer discovery. Topics remain corpus-derived.}
  };

  \node[cell] (supervised) [below=of epilens] {
    \textbf{Supervised}\\\textbf{classifiers}\\[4pt]
    {\footnotesize TF-IDF\,+\,SVM, \mbox{SetFit}, ModernBERT}\\[5pt]
    {\footnotesize\itshape Learned from labelled papers.}
  };

  \node[colhead] (col1) [above=of topics]   {Corpus-inferred\\categories};
  \node[colhead] (col2) [above=of epilens] {Analyst-specified\\categories};

  \node[rowhead] (row1) [left=of topics]       {No training\\labels};
  \node[rowhead] (row2) [left=of guidedTopics] {Labelled papers\\required};

  \node[axistitle] (xaxis) at ($(col1.north)!0.5!(col2.north)+(0,0.4)$) 
    {ORIGIN OF CATEGORIES};

  \node[axistitle, rotate=90] (yaxis) at ($(row1.west)!0.5!(row2.west)+(-0.5,0)$) 
    {LABELLING EFFORT};

\end{tikzpicture}
}%
  \caption{\textbf{The design space of methods for classifying scientific literature.}
  The map separates two choices: where the categories come from and whether the method requires a labelled training set before use.
  \toolname{} occupies the analyst-specified, annotation-free cell: categories are written as natural-language descriptions and applied without task-specific training labels, and retrieval then grounds those decisions in paper evidence.}
  \label{fig:design_space}
\end{figure}

Classifying a literature into analytically useful categories admits methods that differ along two requirements that are easy to conflate (\Cref{fig:design_space}): whether the category scheme is \emph{specified by the analyst} or \emph{discovered from the corpus}, and whether the method needs \emph{manually labelled training examples} before it can be applied.

\textit{Unsupervised and guided topic models} (latent semantic analysis (LSA)~\citep{deerwester1990indexing}, probabilistic latent semantic analysis (PLSA)~\citep{hofmann1999probabilistic}, latent Dirichlet allocation (LDA)~\citep{Blei2003LDA}, non-negative matrix factorisation (NMF)~\citep{lee1999learning}, standard BERTopic~\citep{Grootendorst2022BERTopic}, and seed-guided variants) need no labelled papers but infer themes from the corpus rather than applying a predefined scheme.
They answer what structure a literature happens to exhibit, not how its papers fall under a fixed scheme.
Semi-supervised topic-model variants occupy a different position: they can use labelled papers to steer topic discovery, but their output remains corpus-inferred topics instead of direct predictions following the analyst's coding scheme.

\textit{Supervised classifiers} learn an analyst-specified scheme from labelled papers by pairing a \emph{representation} of the document with a trained \emph{classifier head}.
Rather than designate a single representative, we name three representative points in that space, plus a calibrated prototype as the most minimal use of labels.
\emph{TF-IDF\,+\,SVM}, a term-frequency--inverse-document-frequency representation~\citep{sparckjones1972statistical} (each word weighted by how distinctive it is to a document) fed to a linear support vector machine~\citep{cortes1995support}, is the classic lexical anchor.
\emph{SetFit}~\citep{tunstall2022efficient}, which fine-tunes a sentence-embedding model on the labelled examples, is the fine-tuned neural representative.
\emph{ModernBERT}~\citep{warner2024modernbert}, an $8{,}192$-token long-context encoder applied to the parsed body text (overflow handled by window pooling), is the long-context body-text representative.
The calibrated prototype tunes only a single decision threshold.
The full representation $\times$ text-scope $\times$ classifier-head grid behind these named points is swept in \Cref{app:baselines:grid}.

\textit{Off-the-shelf zero-shot classifiers} apply analyst-specified category descriptions without task-specific training labels.
In the NLI formulation, the input text is treated as a premise, and each candidate label is verbalised as a hypothesis.
The model then scores whether the premise entails the hypothesis~\citep{yin2019benchmarking}.
In our runs, the premise is the paper title and abstract.
This is consistent with the standard usage mode for zero-shot NLI classifiers.
We do not use the full paper as premise because standard NLI models have a finite input window for scoring premise-hypothesis pairs, which scientific papers often exceed.
We also evaluate a \emph{prototype similarity} approach for zero-shot classification, which represents each label by a single \emph{prototype} vector (the embedding of its description) and assigns the labels whose prototypes are closest to the paper's own embedding.

\toolname{}'s Paper Classifier occupies the position the others leave awkward: analyst-specified categories, revisable as text, applied with no training labels, and grounded in full-document evidence through retrieval.
A \emph{majority prior} (always predicting the most common label set) provides a reference floor.
\Cref{fig:design_space} maps these families and \toolname{}'s place among them.
The next section records how each method is configured and the role it plays.
We then position them by what each requires against the accuracy it reaches.

\subsection{Configuration and role of each method}
\label{app:baselines:descriptions}

Having placed each family in the design space (\Cref{app:baselines:taxonomy}), we now give the configuration and role of each method, grouped by family, so the comparisons that follow can be read against a fully specified set of baselines.
The \emph{majority prior} sits outside these families as a reference floor rather than a method: it always predicts the most frequent label set (estimated leave-one-out to avoid leakage from the target paper's own label), and marks how much apparent performance follows from label skew alone.

\subsubsection{Zero-shot baselines}
\label{app:baselines:zero_shot}

Zero-shot classifier baselines assign papers to an analyst-specified label set without training on manually annotated task instances.
Here we use the term for two retrieval-free, annotation-free classifiers: a prototype-similarity classifier whose label representations are built from the coding protocol, and an off-the-shelf NLI classifier that scores verbalised label hypotheses against the paper metadata.
Both operate on title and abstract only.
Calibrated variants that tune thresholds on manual labels are reported with the supervised baselines.

\paragraph{Prototype similarity.}
Each label is represented by a \emph{prototype} vector: the embedding of its definition text where the coding protocol provides one (paper type, data type, data accessibility), and of the label name otherwise (geography).
At inference, the paper's title and abstract are embedded in the same space and scored by cosine similarity to each prototype.
We use two representations: a sparse TF-IDF space and a dense one from \texttt{allenai-specter}, a sentence transformer trained on scientific-paper pairs.
The choice of representation matters for what ``similarity'' means.
With TF-IDF, the label prototype is essentially its own words, so the method boils down to lexical overlap between the abstract and the label text.
For geography, where the prototype is just the continent name, this is close to detecting whether the abstract literally mentions that region.
The dense SPECTER prototype relaxes this to semantic similarity, which is why it helps on the axes with descriptive definitions more than on the name-only geography axis.
Assignment uses a relative rule rather than a single absolute cutoff: a label is kept if its similarity is at least a fraction $\tau_{\text{rel}}{=}0.92$ of the top-scoring label's similarity, above an absolute floor $\tau{\geq}0.03$, and among the top three.
The \emph{uncalibrated} regime uses these defaults and needs no labels, so it is a genuinely annotation-free zero-shot baseline.
The \emph{calibrated} regime instead selects the absolute floor $\tau$ per fold by 5-fold cross-validation (maximising mean Jaccard on the other folds, over a grid of score quantiles that includes the default), holding the relative rule fixed.
This uses labels, so it counts as a minimal supervised method reported with the supervised baselines.
Because the default is among the candidates, calibration cannot underperform it on the selection folds, though held-out performance may still be lower.
Calibration recovers part of the gap on the multi-label tasks (\Cref{tab:baseline_comparison_local}).
\emph{Paper type}, single-label and decided by arg-max, has no threshold to tune.
Even when calibrated, the prototype stays below discriminative supervised classifiers because a single global acceptance rule cannot capture per-label classification boundaries.

\paragraph{Zero-shot NLI entailment.}
Following the standard zero-shot text-classification recipe~\citep{yin2019benchmarking}, each candidate label is framed as a natural-language-inference hypothesis (``This paper is about \{label\}.''), and an NLI-fine-tuned DeBERTa-v3-base checkpoint (\path{MoritzLaurer/DeBERTa-v3-base-mnli-fever-anli}~\citep{he2023debertav3,laurer_deberta_nli}) scores the entailment of each hypothesis against the paper's title and abstract.
Labels clearing an entailment threshold are kept (the highest-scoring label for the single-label \emph{paper type} task).
This is the most direct off-the-shelf occupant of \toolname{}'s design-space quadrant (analyst-specified yet annotation-free), but, as we run it, it reads only the title and abstract and reasons one label at a time.
Its behaviour varies widely across axes.
We analyse why in \Cref{app:baselines:llm_buys}.
This unevenness does not seem to stem from an obviously weak hypothesis template.
We tested a task-specific template (``This paper uses {label} data''), but it did not recover \emph{Traditional} and increased over-prediction of non-traditional labels.
Replacing label names with one-sentence \emph{definitions} from the coding protocol (\Cref{app:labeling_protocol}) also lowered scores on most tasks, suggesting that the NLI model is sensitive to hypothesis wording and does not reliably apply the full coding protocol from metadata alone.
We therefore report the stronger label-name variant.

\subsubsection{Guided and semi-supervised topic-model baselines}
\label{app:baselines:guided_topics}

\paragraph{Guided and semi-supervised BERTopic.}
BERTopic represents documents with transformer embeddings, clusters them in a reduced space, and labels each cluster with class-based TF--IDF descriptors~\citep{Grootendorst2022BERTopic}.
Its native output is a set of corpus-inferred topics, not assignments to an analyst-defined taxonomy.
In its plain, unlabelled form, it is therefore a pure corpus-structure probe: it asks how far each axis separates from corpus-level vocabulary structure with no label information at all.\ifshowaudit\ In that form, BERTopic is one of the topic models in the corpus-structure audit (\Cref{app:baselines:audit}), which is where we report that recoverability question.\fi

The guided and semi-supervised variants differ in the kind of analyst input they consume.
The \emph{guided} variant supplies seed terms drawn from each label definition, biasing topic discovery toward the target concepts while leaving the discovered topics free to merge, split, or ignore the labels.
It therefore uses analyst guidance but not labelled training papers.
The \emph{semi-supervised} variant instead passes the manual labels of the training papers, using partial document-level supervision to steer the clustering.

For this reason, the two variants occupy different rows of \Cref{fig:design_space}.
Guided BERTopic remains annotation-free with respect to paper labels and belongs with the topic-model family: the analyst supplies seeds, but the emitted topics remain corpus-inferred and must be interpreted or aligned after the fact.
Semi-supervised BERTopic belongs in the labelled-papers row, because it consumes manual labels for training papers, but it still remains in the corpus-inferred column rather than the analyst-specified classifier column: labels steer the clustering, whereas the output is still a set of topics that need not correspond one-to-one to the coding scheme.
The contrast with supervised classifiers is therefore not whether labels can influence the method, but how they enter: a supervised head learns a decision boundary that maps text directly onto the fixed label set, whereas guided or semi-supervised topic modelling uses seeds or labels to bias topic discovery.

These topic-model variants answer a different question from \toolname{}: not whether a fixed coding protocol can be applied directly to each paper, but whether the label axes match separable corpus structure once discovery is nudged toward the target concepts.
We report these guided and semi-supervised topic-model variants in \Cref{tab:baseline_comparison_local}.
To avoid tying the comparison to a single topic granularity, we sweep the requested number of topics over $K \in \{0.5,\,1,\,2,\,4\}\times|\mathcal{L}|$ and report the best-aligned configuration.

\subsubsection{Supervised baselines}
\label{app:baselines:supervised}

Supervised baselines all follow one recipe: pair a \emph{representation} of the text with a trained \emph{classifier head}, fit on the manual labels under cross-validation.
Within that recipe, they differ in one methodologically important way, \emph{where the label signal enters}.
In most of them, the representation is fixed independently of the labels, either unsupervised (the sparse TF-IDF space) or a frozen pretrained encoder whose weights are never updated, and the labels train only the head on top: there is no gradient path from the head's loss back to the representation, which reaches the head as a fixed feature matrix.
SetFit is the sole exception, letting the labels also reshape the representation by contrastively fine-tuning the encoder before fitting a head on the adapted embeddings.
Even there, the encoder and head are fit in separate stages, not by one joint objective, so no baseline here is end-to-end fine-tuned in the sense of gradients flowing through the encoder and classifier together.
Jointly training and tuning a full encoder and classifier across four tasks would be disproportionate to an evaluation with only $214$ manually labelled papers and would also carry a high risk of overfitting.
We surface three named points, and \Cref{app:baselines:grid} sweeps the full representation\,$\times$\,scope\,$\times$\,head grid behind them: \emph{TF-IDF\,+\,SVM} (classic lexical, head-only supervision), \emph{SetFit} (the one representation-adapting encoder), and \emph{ModernBERT} (frozen long-context body-text encoder).
All are trained and evaluated by cross-validation over the $N{=}214$ papers (using iterative multi-label stratification for the set-valued axes where feasible, to balance label prevalence across folds), on the title+abstract scope shared with the other abstract-scoped methods.
\emph{ModernBERT} is the exception, applied to the parsed body text.
We additionally report TF-IDF under leave-one-out as a robustness variant.
Together, they show how far label-supervised text classification can go, as a reference for the contribution of retrieval-grounded reasoning.

\paragraph{\texorpdfstring{TF-IDF\,+\,SVM}{TF-IDF + SVM}.}
The classical anchor and simplest instance of the recipe, and a pure case of head-only supervision: a sparse TF-IDF representation of the title and abstract (each term weighted by how distinctive it is to a document), built from corpus term statistics without ever seeing the labels, is fed to a linear support vector machine that is the only trained component, under stratified 5-fold cross validation with leave-one-out as a robustness variant (\Cref{tab:supervised_grid}).
It is the strongest non-neural reference.

\paragraph{SetFit.}
SetFit~\citep{tunstall2022efficient} is a supervised, label-efficient text-classification method that first fine-tunes a sentence-transformer on pairs of labelled examples with a contrastive objective, then trains a classification head on the resulting embeddings.
It is thus the one supervised baseline here in which the labels reach the representation itself, not only the head, though the contrastive fine-tuning and the head are still fit in separate stages rather than by a single joint objective.
We use \texttt{sentence-transformers/all-mpnet-base-v2} as the base encoder and evaluate it under stratified 5-fold cross-validation across the $N{=}214$ evaluation papers.
We do not use leave-one-out here, since it would need a separate fine-tuning run for each held-out paper.
To cap the number of contrastive pairs drawn from our relatively large cross-validation folds, we set \texttt{num\_iterations}${=}2$ instead of the default sampling strategy.
Text scope is restricted to the title and abstract.
This is also the natural scope for the chosen sentence-transformer: \texttt{all-mpnet-base-v2} is intended for sentence- and short-paragraph embeddings, and its default sentence-transformers configuration truncates inputs longer than a few hundred word pieces.
Applying it to the full body would therefore require either truncation or a separate chunk-level pooling rule to aggregate window embeddings into a paper-level prediction.

\paragraph{ModernBERT (body text, long-context encoder).}
ModernBERT~\citep{warner2024modernbert} is a bidirectional encoder trained with a native $8{,}192$-token sequence length, substantially longer than the few-hundred-token windows of the other dense encoders.
We apply it as a \emph{frozen} feature extractor under the same stratified 5-fold protocol as the other supervised baselines, with its weights never updated (as for the other dense encoders here): each body window is encoded with the encoder held fixed, token embeddings are mean-pooled to obtain a window representation, and these fixed representations are passed to a separately trained MLP classification head.
For papers whose parsed body fits within the $8{,}192$-token window, this gives a single body-level representation.
For longer papers, we encode multiple windows and mean-pool their representations rather than truncating the body (\Cref{app:baselines:grid}).
ModernBERT is therefore the supervised baseline with the widest body-text scope, testing how far a frozen long-context encoder's features, with only the head trained, can go without label-targeted retrieval.

\subsubsection{RAG ablation baselines}
\label{app:baselines:llm}

Three variants of the same workflow (metadata only, random chunks, and the whole body) join full \toolname{} in the four-condition ablation of \Cref{sec:ablation}.

\paragraph{Metadata-only LLM (\texttt{metadata\_llm}).}
Runs the same language model with the same classification prompts as \toolname{}, but replaces retrieved full-text evidence with only the paper's title and abstract.
This directly isolates the contribution of retrieval: any performance gap between this baseline and the full system is attributable to the retrieved body-text passages.
It is given the raw parsed metadata (title and abstract) and never any body text, keeping it a clean ablation of retrieval.
For the $32$ papers with no machine-readable abstract, it therefore sees the title alone, exactly as the other abstract-scoped baselines do.
The paper is not skipped, so we score all $214$ on the same set as the full-text system.

\paragraph{Random-chunk LLM (\texttt{random\_chunk\_llm}).}
Runs the same model and prompts as the full system, but selects $k$ body-text passages by \emph{uniform random sampling} from the per-paper document pool in the vector index rather than by semantic similarity to the template paragraphs.
This holds body-text access fixed while ablating retrieval quality, thereby isolating the contribution of \emph{targeted} retrieval from the contribution of \emph{body-text access} per se.

\paragraph{Whole-body LLM (\texttt{full\_body\_llm}).}
Runs the same model and prompts, with the same per-paper chunk pool, but passes every body chunk of the paper in document order instead of $k$ of them, so the prompt holds the whole parsed body (the longest prompts have $22{,}451$ and $22{,}771$ tokens for the two generators, within their context windows).
It tests whether selecting passages is needed at all.
We ran this condition at $T{=}0$ five times with \texttt{gemini-2.5-flash}, reported on the openly licensed papers only (\Cref{app:licences}), and once with the fully local \texttt{qwen2.5:32b-instruct}, whose full-workflow outputs were identical across five repeats (\Cref{app:qwen_local}).

\subsection{Method requirements and performance trade-offs}
\label{app:baselines:positioning}

This comparison is not a leaderboard, but a map of what each approach requires in order to reach a given level of agreement.
The relevant trade-off is between annotation effort, text scope, and task flexibility: \toolname{} uses analyst-specified categories that can be edited as text, requires no task-specific training labels, and grounds each decision in retrieved body-text evidence.
The baseline scores therefore measure not only modelling accuracy, but also what is lost or gained when one removes retrieval, restricts the text scope, or introduces labelled supervision.

\Cref{tab:baseline_comparison_local} collects the direct comparison results for the fully local workflow on all $214$ papers.
The next two subsections compare methods first at title-and-abstract scope and then with access to body text.

To separate model effects from evidence access, we use \texttt{metadata\_llm} as the pivot comparison.
This baseline keeps \toolname{}'s generator and task instructions but restricts the evidence to paper metadata, i.e.\ the title and abstract.
It therefore answers two questions in sequence: first, at title-and-abstract scope, what does an instruction-following LLM buy relative to classical, neural, and zero-shot baselines? Second, what does targeted body-text retrieval add on top?

\subsubsection{Comparison at title-and-abstract scope}
\label{app:baselines:llm_buys}

At title-and-abstract scope, \texttt{metadata\_llm} is the cleanest comparison because it uses the same instruction-following backend as \toolname{} without targeted body-text retrieval.
With the local model, performance ($0.659$ paper type, $0.807$ geography, $0.741$ data type, $0.213$ data accessibility) shows that the LLM's benefit at fixed evidence scope is task-dependent rather than uniform.

Where the relevant signal is usually present in the abstract, the metadata-only LLM exceeds the standard abstract-scoped baselines.
This is clearest for \emph{geography} ($0.807$ vs.\ TF-IDF\,+\,SVM $0.592$, SetFit $0.599$, NLI $0.709$), and it holds against the strongest tuned abstract-scope configuration of the supervised grid ($0.750$; \Cref{tab:supervised_grid}).
Elsewhere, the labelled baselines do better.
On \emph{paper type} ($0.659$ vs.\ $0.713$ for TF-IDF\,+\,SVM and $0.734$ for SetFit) and \emph{data type} ($0.741$ vs.\ $0.854$ and $0.864$), the supervised abstract classifiers lead, consistent with lexically regular tasks dominated by a few frequent labels.
The clearest failure is \emph{data accessibility}: \texttt{metadata\_llm} reaches only $0.213$, far below the supervised abstract baselines ($0.533$--$0.574$), because the decisive access statement is rarely available in the abstract.
This marks the limit of metadata-only reasoning and identifies where retrieval should matter most.
Thus, \toolname{}'s advantage is not instruction-following at fixed scope: it comes from adding body-text evidence.

The weakest baseline scores should be read as diagnostics of default method--task mismatch, not as ceilings on the corresponding model families.
Zero-shot NLI on \emph{data type} is the clearest case ($J{=}0.018$): it almost never predicts the dominant \emph{Traditional} class.
This seems to reflect a lexical-anchoring asymmetry rather than a residual-category effect: in our protocol \emph{Traditional} is defined positively (standard surveillance case counts, hospital admissions, PCR/antigen-confirmed test line lists), but it is a broad, heterogeneous category that papers signal only implicitly, since a paper reporting national surveillance counts rarely states in so many words that it uses ``traditional'' data.
The non-traditional categories, by contrast, each name a specific kind of source (mobility, health, sentiment, economic) that papers typically do state explicitly, giving an entailment hypothesis built from the label name a much stronger lexical anchor to key on than \emph{Traditional} offers.

Prototype similarity shows a different failure mode: its uncalibrated version is a minimal annotation-free semantic-matching baseline, while threshold calibration uses labels to recover part of the gap.
For example, calibration improves SPECTER on \emph{data type} from $0.52$ to $0.67$, and TF-IDF on \emph{data accessibility} from $0.22$ to $0.25$ (\Cref{tab:baseline_comparison_local}).
We therefore treat the calibrated prototype as a minimal supervised method and the uncalibrated prototype as a zero-shot diagnostic.

\subsubsection{Comparison with body-text access}
\label{app:baselines:retrieval_adds}

Moving from \texttt{metadata\_llm} to full \toolname{} holds the generator and task instructions fixed while replacing metadata-only evidence with targeted body-text retrieval.
The resulting difference estimates the contribution of evidence access and selection and tracks where the discriminative signal lies.
With the local model, body-text retrieval improves every axis, and each paired difference meets the adjusted threshold (\Cref{tab:significance_local}).
The margin is largest where decisive evidence is distributed through the body or concentrated in a local statement: paper type improves from $0.659$ to $0.794$, and data accessibility from $0.213$ to $0.539$.
On \emph{geography}, where the abstract already carries much of the signal, the gain is smaller ($0.807$ to $0.869$).
Against the supervised baselines, the local workflow leads on geography and paper type, while on \emph{data type} it matches them ($0.861$ vs.\ TF-IDF\,+\,SVM $0.854$) and full-document TF-IDF scores higher ($0.881$, \Cref{tab:supervised_grid}), suggesting that this axis is driven more by broad lexical evidence than by label-targeted retrieval.
On \emph{data accessibility}, the local workflow is close to the supervised baselines ($0.539$ vs.\ $0.533$--$0.577$); the served models, which read the same passages with a stronger generator, lead them on the openly licensed papers (\Cref{tab:baseline_comparison_open}).

This body-text access is not an information advantage withheld from the baselines.
Title and abstract are the natural scope for short-window encoders and off-the-shelf zero-shot classifiers, but we also evaluate supervised methods with body-text access in the configuration grid: full-document vectorisation for TF-IDF, chunk-and-pool variants for short-window encoders, and \texttt{ModernBERT}'s $8{,}192$-token long-context window for the dense long-context setting (\Cref{tab:supervised_grid}).
The distinction is therefore not whether baselines are allowed to read the body, but how it is used: the language model applies the coding protocol to body-text passages, and selecting those passages by the labels adds most for data accessibility.
The component ablation (\Cref{sec:ablation}) decomposes this retrieval contribution directly, while the supervised grid and saturation analysis (\Cref{app:baselines:grid}) make an underpowered baseline or small-label artefact unlikely as the main explanation.

\subsection{Statistical methodology}
\label{app:baselines:statistics}

The baseline comparison in \Cref{sec:baselines_summary} uses the mean per-paper Jaccard score defined in \Cref{sec:classification_labeled} on the same $214$ manually labelled papers, and on the $122$ openly licensed ones for the served models.
We use paired resampling to check how sensitive the observed baseline differences and the ablation differences in \Cref{sec:ablation} are to the papers evaluated.
For \toolname{}, each paper's score is averaged over five runs.
The supervised baselines use out-of-fold predictions: within each configuration, each paper is predicted by a model fitted without that paper.

For each comparison, we resample the per-paper Jaccard differences with replacement ($B{=}10{,}000$) and calculate a bias-corrected and accelerated (BCa) interval for their mean.
This keeps both methods' scores paired on each paper; BCa adjusts the interval for bias and skewness~\citep{efron1987better}.
We report pointwise $95\%$ intervals; each two-sided p-value is the smallest significance level at which the corresponding BCa interval excludes zero.
We apply Holm--Bonferroni correction within each task, separately for the baseline comparisons and each generator's ablation comparisons.
Holm's step-down adjustment limits the chance of at least one false rejection within each comparison family~\citep{holm1979simple}.
\Cref{tab:local_main,tab:baseline_comparison_local} give the same BCa interval for one configuration's mean $J$, resampling papers in the same way.
This resampling holds the manual labels, model predictions, and selected configurations fixed.
It does not reproduce the review-stratified sampling design or model fitting, so the intervals describe sensitivity to these papers rather than uncertainty about performance across corpora.

In \Cref{tab:significance_local}, the local workflow's gaps to the three named supervised baselines meet the adjusted threshold on geography for all three and on paper type against TF-IDF\,+\,SVM; on data type and data accessibility none does, and the differences are within $0.04$.
All four gaps to the abstract-only \texttt{metadata\_llm} meet it, so body-text retrieval adds to the same model on every axis.
On the openly licensed papers, \Cref{tab:significance_open} gives the same tests for \texttt{gemini-2.5-flash}.

\begin{table}[!htbp]
\centering
\small
\setlength{\tabcolsep}{4pt}
\caption{\textbf{Paired comparisons of \toolname{} (\texttt{qwen2.5:32b-instruct}, $T{=}0$) with selected baselines.}}
\label{tab:significance_local}
\begin{tabular}{lcccc}
\toprule
\textbf{Baseline} & \textbf{Paper type} & \textbf{Geography} & \textbf{Data type} & \textbf{Data access.} \\
\midrule
    Zero-shot NLI & $+0.229$$^{*}$ & $+0.160$$^{*}$ & $+0.844$$^{*}$ & $+0.184$$^{*}$ \\
    TF-IDF + SVM & $+0.082$$^{*}$ & $+0.277$$^{*}$ & $+0.008$\phantom{$^{*}$} & $+0.006$\phantom{$^{*}$} \\
    TF-IDF + SVM (LOO) & $+0.082$$^{*}$ & $+0.262$$^{*}$ & $-0.010$\phantom{$^{*}$} & $+0.001$\phantom{$^{*}$} \\
    TF-IDF + SVM (full body) & $+0.070$\phantom{$^{*}$} & $+0.254$$^{*}$ & $-0.019$\phantom{$^{*}$} & $-0.051$\phantom{$^{*}$} \\
    SetFit & $+0.061$\phantom{$^{*}$} & $+0.269$$^{*}$ & $-0.002$\phantom{$^{*}$} & $-0.035$\phantom{$^{*}$} \\
    ModernBERT (full body) & $+0.009$\phantom{$^{*}$} & $+0.217$$^{*}$ & $-0.003$\phantom{$^{*}$} & $-0.039$\phantom{$^{*}$} \\
    \texttt{metadata\_llm} & $+0.136$$^{*}$ & $+0.062$$^{*}$ & $+0.121$$^{*}$ & $+0.325$$^{*}$ \\
    \midrule
    \addlinespace[2pt]\multicolumn{5}{l}{\textbf{Reference}} \\
    \quad Majority prior (floor) & $+0.201$$^{*}$ & $+0.516$$^{*}$ & $+0.019$\phantom{$^{*}$} & $+0.067$\phantom{$^{*}$} \\
\bottomrule
\end{tabular}
\footnotetext{Each cell is the mean per-paper Jaccard difference (\toolname{} minus the baseline) on the $N{=}214$ evaluation papers.
A marker ($^{*}$) denotes a Holm-adjusted $p$-value below $0.05$ across eight comparisons per task.
All eight comparisons in the correction are shown.
The majority floor is shown for reference: it predicts the most common label set of the manual labels, which \toolname{} does not see, and its row tests whether $J$ exceeds what label skew alone gives.
It enters the Holm correction, which can only make the other comparisons more conservative, never less.
Tests use $B{=}10{,}000$ paired BCa resamples (\Cref{app:baselines:statistics}).
They hold the manual labels, predictions, and configurations fixed.
The topic-model and \emph{Best supervised} rows of \Cref{tab:baseline_comparison_local} are descriptive because their alignment or configuration was selected using the evaluation papers.
Prototype rows are reported as diagnostics and are not tested here.}
\end{table}

\subsection{Comprehensive supervised configuration grid and saturation}
\label{app:baselines:grid}

To position \toolname{} against a wider range of supervised approaches than the named representatives of \Cref{app:baselines:positioning}, \Cref{tab:supervised_grid} crosses three choices.
The representations are sparse TF-IDF, frozen \texttt{all-mpnet-base-v2}~\citep{song2020mpnet,reimers2019sentence}, frozen \texttt{SPECTER}~\citep{cohan2020specter}, and the frozen 8,192-token long-context \texttt{ModernBERT-base}.
Text scope is either title and abstract or full body, ingested by chunk-and-pool for the short-window encoders and within \texttt{ModernBERT}'s $8{,}192$-token window, with overflow mean-pooled across windows.
The classifier heads are linear SVM, logistic regression, gradient-boosted trees, and MLP.
Each combination follows the same stratified 5-fold protocol over the same $N{=}214$ papers, with leave-one-out additionally reported for the linear heads.
The MPNet embedding uses the \path{sentence-transformers/all-mpnet-base-v2} checkpoint \citep{allmpnet_model_card}.
Every cell is head-only supervision (\Cref{app:baselines:supervised}): the representation is fixed (unsupervised TF-IDF or a frozen encoder) and only the head is trained, so the grid isolates representation family, text scope, and head choice without the confound of end-to-end fine-tuning.
The heads use fixed settings, not tuned in an inner cross-validation loop, and all configurations share one fold partition.
The \emph{Best supervised} row offsets this in the baselines' favour: it picks the best configuration for each axis on the evaluation papers themselves, which overstates what a configuration chosen in advance would reach.
Full-body access and dense encoders both raise the best score reached in the grid, and non-linear heads edge the linear ones on dense representations.
\emph{No} cell reaches the fully local \toolname{} on geography or paper type, but about half of the cells exceed it on data type and data accessibility (\Cref{tab:supervised_grid}).
The long-context \texttt{ModernBERT}, whose $8{,}192$-token window holds the full body of $182$ of the $214$ papers in a single window (the $32$ longer papers, up to ${\sim}16$k tokens, are mean-pooled across a few windows, as the short-window encoders always are), lands in the same band as the other encoders, suggesting that short input windows alone do not explain the differences.

\begin{table}[!htbp]
\centering
\small
\setlength{\tabcolsep}{4pt}
\caption{\textbf{Comprehensive supervised configuration grid.}}
\label{tab:supervised_grid}
\begin{tabular}{lcccc}
\toprule
\textbf{Configuration} & \textbf{Paper type} & \textbf{Geog.} & \textbf{Data type} & \textbf{Data access.} \\
\midrule
    \multicolumn{5}{l}{\textit{TF-IDF (title+abstract)}} \\
    \quad Linear SVM & 0.713\,/\,0.713 & 0.592\,/\,0.607 & 0.854\,/\,0.871 & 0.533\,/\,0.538 \\
    \quad Logistic Reg. & 0.706\,/\,0.696 & 0.593\,/\,0.575 & 0.881\,/\,0.871 & 0.528\,/\,0.514 \\
    \quad GB trees & 0.595\,/\,-- & 0.453\,/\,-- & 0.814\,/\,-- & 0.467\,/\,-- \\
    \quad MLP & 0.696\,/\,-- & 0.599\,/\,-- & 0.856\,/\,-- & 0.558\,/\,-- \\
    \multicolumn{5}{l}{\textit{TF-IDF (full body)}} \\
    \quad Linear SVM & 0.724\,/\,0.741 & 0.615\,/\,0.627 & 0.881\,/\,0.890 & 0.590\,/\,0.583 \\
    \quad Logistic Reg. & 0.694\,/\,0.712 & 0.600\,/\,0.589 & 0.862\,/\,0.850 & 0.555\,/\,0.566 \\
    \quad GB trees & 0.764\,/\,-- & 0.596\,/\,-- & 0.883\,/\,-- & 0.586\,/\,-- \\
    \quad MLP & 0.748\,/\,-- & 0.651\,/\,-- & 0.888\,/\,-- & 0.619\,/\,-- \\
    \multicolumn{5}{l}{\textit{Frozen mpnet (title+abstract)}} \\
    \quad Linear SVM & 0.672\,/\,0.693 & 0.703\,/\,0.686 & 0.822\,/\,0.817 & 0.461\,/\,0.453 \\
    \quad Logistic Reg. & 0.661\,/\,0.681 & 0.685\,/\,0.683 & 0.770\,/\,0.783 & 0.459\,/\,0.464 \\
    \quad GB trees & 0.738\,/\,-- & 0.694\,/\,-- & 0.875\,/\,-- & 0.593\,/\,-- \\
    \quad MLP & 0.780\,/\,-- & 0.750\,/\,-- & 0.873\,/\,-- & 0.588\,/\,-- \\
    \multicolumn{5}{l}{\textit{Frozen mpnet (full body)}} \\
    \quad Linear SVM & 0.661\,/\,0.676 & 0.736\,/\,0.726 & 0.836\,/\,0.815 & 0.539\,/\,0.475 \\
    \quad Logistic Reg. & 0.641\,/\,0.664 & 0.752\,/\,0.734 & 0.787\,/\,0.807 & 0.521\,/\,0.489 \\
    \quad GB trees & 0.750\,/\,-- & 0.724\,/\,-- & 0.898\,/\,-- & 0.612\,/\,-- \\
    \quad MLP & 0.757\,/\,-- & 0.771\,/\,-- & 0.883\,/\,-- & 0.577\,/\,-- \\
    \multicolumn{5}{l}{\textit{Frozen SPECTER (title+abstract)}} \\
    \quad Linear SVM & 0.686\,/\,0.719 & 0.551\,/\,0.545 & 0.822\,/\,0.854 & 0.494\,/\,0.507 \\
    \quad Logistic Reg. & 0.667\,/\,0.681 & 0.526\,/\,0.543 & 0.787\,/\,0.815 & 0.469\,/\,0.488 \\
    \quad GB trees & 0.720\,/\,-- & 0.555\,/\,-- & 0.861\,/\,-- & 0.574\,/\,-- \\
    \quad MLP & 0.742\,/\,-- & 0.633\,/\,-- & 0.881\,/\,-- & 0.558\,/\,-- \\
    \multicolumn{5}{l}{\textit{Frozen SPECTER (full body)}} \\
    \quad Linear SVM & 0.663\,/\,0.720 & 0.622\,/\,0.628 & 0.839\,/\,0.846 & 0.533\,/\,0.530 \\
    \quad Logistic Reg. & 0.663\,/\,0.716 & 0.619\,/\,0.631 & 0.804\,/\,0.835 & 0.512\,/\,0.525 \\
    \quad GB trees & 0.778\,/\,-- & 0.627\,/\,-- & 0.887\,/\,-- & 0.593\,/\,-- \\
    \quad MLP & 0.741\,/\,-- & 0.683\,/\,-- & 0.893\,/\,-- & 0.588\,/\,-- \\
    \multicolumn{5}{l}{\textit{ModernBERT (8k) (title+abstract)}} \\
    \quad Linear SVM & 0.726\,/\,0.693 & 0.507\,/\,0.543 & 0.826\,/\,0.834 & 0.482\,/\,0.511 \\
    \quad Logistic Reg. & 0.688\,/\,0.678 & 0.494\,/\,0.538 & 0.782\,/\,0.827 & 0.482\,/\,0.504 \\
    \quad GB trees & 0.720\,/\,-- & 0.485\,/\,-- & 0.863\,/\,-- & 0.565\,/\,-- \\
    \quad MLP & 0.731\,/\,-- & 0.581\,/\,-- & 0.878\,/\,-- & 0.583\,/\,-- \\
    \multicolumn{5}{l}{\textit{ModernBERT (8k) (full body)}} \\
    \quad Linear SVM & 0.702\,/\,0.706 & 0.591\,/\,0.605 & 0.829\,/\,0.875 & 0.495\,/\,0.524 \\
    \quad Logistic Reg. & 0.700\,/\,0.718 & 0.595\,/\,0.627 & 0.822\,/\,0.840 & 0.472\,/\,0.504 \\
    \quad GB trees & 0.734\,/\,-- & 0.534\,/\,-- & 0.871\,/\,-- & 0.572\,/\,-- \\
    \quad MLP & 0.785\,/\,-- & 0.652\,/\,-- & 0.864\,/\,-- & 0.577\,/\,-- \\
\bottomrule
\end{tabular}
\footnotetext{Sample-averaged Jaccard for every \emph{representation} $\times$ \emph{text scope}
$\times$ \emph{classifier head} combination, all at stratified 5-fold CV over the same
$N{=}214$ papers. Each cell is shown as ``5-fold\,/\,LOO''. The gradient-boosted-tree
and MLP heads show ``--'' for the leave-one-out score, because a LOO sweep would refit
the head once per held-out paper, which is prohibitively expensive for these non-linear
heads, so only the 5-fold score is computed for them. The grid positions \toolname{}
(\Cref{tab:baseline_comparison_local}) against
the full space of reasonable supervised approaches rather than a single representative.
The small 5-fold-to-LOO gaps and the saturation curves (\Cref{fig:saturation}) show the
baselines are near data-saturated at this sample size.}
\end{table}

\begin{figure}[!htbp]
  \centering
  \includegraphics[width=\textwidth]{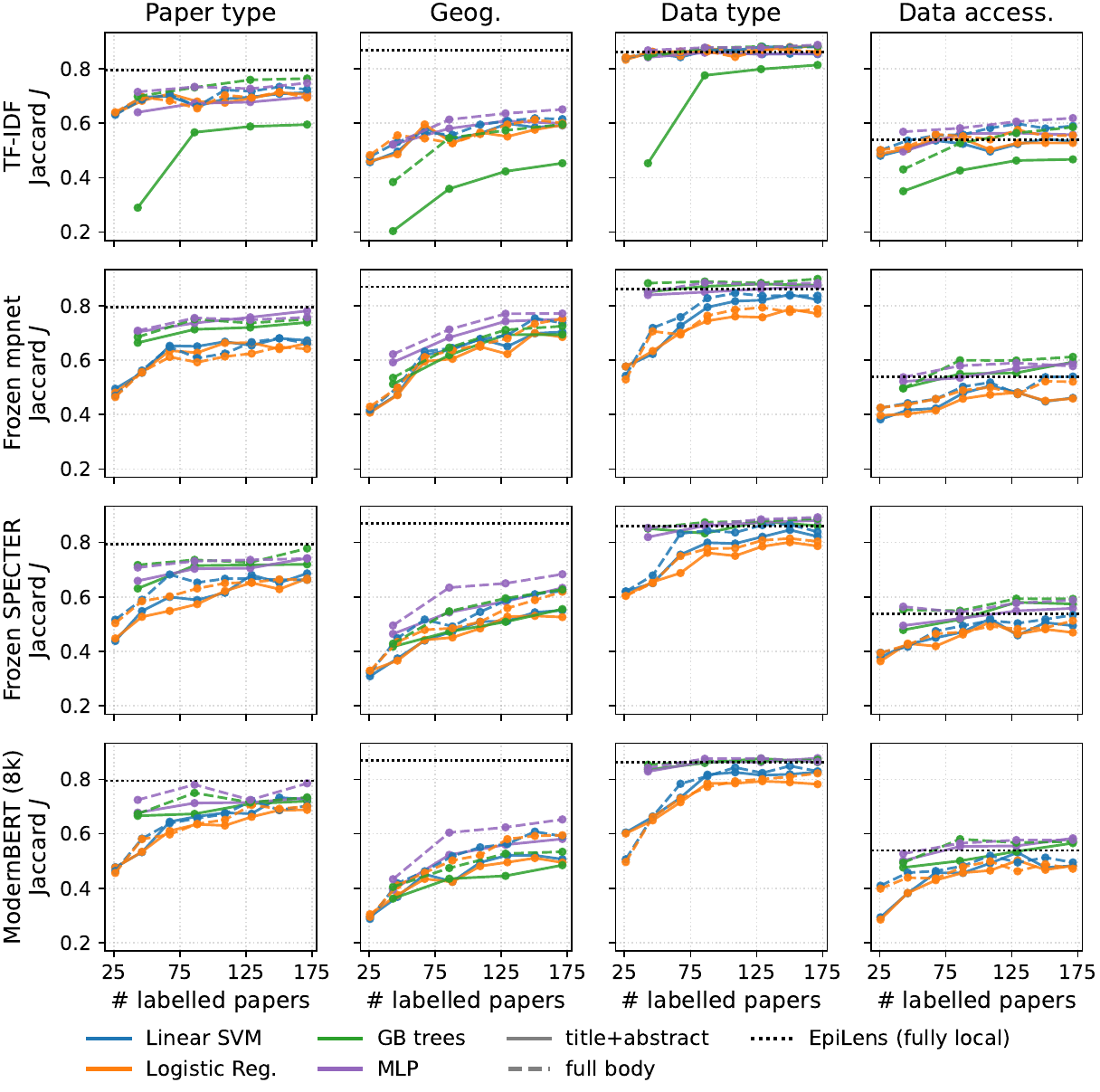}
  \caption{\textbf{Saturation of the supervised configuration grid.}
  Learning curves (held-out Jaccard vs.\ number of labelled training papers) grouped by
  representation (rows) and task (columns).
  Within each panel, colour encodes the
  classifier head (linear SVM, logistic regression, gradient-boosted trees, MLP) and line
  style the text scope (solid = title+abstract, dashed = full body).
  The black dotted line
  is the fully local \toolname{} ($T{=}0$).
  Every curve remains below \toolname{} on geography and paper type inside the observed training range; on data type and data accessibility, many reach it.
  Most curves flatten before the full sample.
  }
  \label{fig:saturation}
\end{figure}

One possible concern is that the supervised baselines have too few training examples.
Inside the observed range, every learning curve in \Cref{fig:saturation} remains below the \toolname{} reference on geography and paper type, while on data type and data accessibility many curves reach it; most flatten before the full sample.
The 5-fold and leave-one-out scores in \Cref{tab:supervised_grid} also differ by only a few points.
This does not establish a ceiling for supervised learning, but it suggests that modest increases in labels alone would not close the gap on geography and paper type under the tested configurations.

\ifshowaudit
\subsection{Corpus-structure audit}
\label{app:baselines:audit}

The topic models of \Cref{app:baselines:taxonomy} are run not as classifiers but as a \emph{corpus-structure audit}: a check of how separable each axis's categories are by vocabulary alone, which bounds how much of the task a purely lexical, retrieval-free method could solve in principle.
\Cref{tab:corpus_audit} reports the best $J$ each model reaches across the $K$ sweep under an oracle cluster-to-label alignment, an \emph{upper bound} on that structural signal.
Even as upper bounds, topic models stay well below the fully local workflow on \emph{geography} and below it on \emph{paper type}, but reach it on \emph{data type} and, with PLSA, exceed it on \emph{data accessibility} ($0.562$ against $0.539$).
On that axis, neither vocabulary structure nor the local generator separates the categories well; the stronger served models gain most there (\Cref{sec:served_reference}).

\begin{table}[!htbp]
\centering
\small
\setlength{\tabcolsep}{3pt}
\caption{\textbf{Corpus-structure audit: how separable each axis is by vocabulary alone.}}
\label{tab:corpus_audit}
\begin{tabular}{lcccc}
\toprule
\textbf{Method} & \textbf{Paper type} & \textbf{Geography} & \textbf{Data type} & \textbf{Data access.} \\
\midrule
    \multicolumn{5}{l}{\textit{Reference}} \\
    \quad Majority prior (floor) & 0.593 & 0.353 & 0.843 & 0.472 \\
    \midrule
    \multicolumn{5}{l}{\textit{Unsupervised topic models (best $K$ across sweep, oracle label alignment)}} \\
    \quad LSA & 0.593 & 0.501 & 0.843 & 0.489 \\
    \quad LDA & 0.757 & 0.489 & 0.880 & 0.484 \\
    \quad NMF & 0.682 & 0.576 & 0.874 & 0.516 \\
    \quad PLSA & 0.748 & 0.539 & 0.875 & 0.562 \\
    \quad BERTopic & 0.626 & 0.581 & 0.843 & 0.477 \\
    \midrule
    \multicolumn{5}{l}{\textit{Reference}} \\
    \quad \toolname{} (fully local, $T{=}0$) & 0.794 & 0.869 & 0.861 & 0.539 \\
\bottomrule
\end{tabular}
\footnotetext{Sample-averaged Jaccard $J$ for unsupervised topic models on the same $214$-paper evaluation set,
reported as the best $J$ across the $K$ sweep ($K \in \{0.5,\,1,\,2,\,4\} \times |\mathcal{L}|$) under an
oracle cluster-to-label alignment, in their original full-text, leave-one-out configuration. These are
\emph{not} classifiers (they discover themes rather than apply the analyst's scheme), and the best-$K$,
oracle-aligned figures are an \emph{upper bound} on what vocabulary structure alone recovers. The majority
floor and \toolname{} are shown for reference. Topic models partly recover the lexically regular axes
(\emph{paper type}, \emph{data type}) but apart from PLSA stay close to the floor on \emph{data accessibility}, the axis whose
signal is a single buried statement.}
\end{table}

\fi

\FloatBarrier
\section{Linking estimates to corpus papers}
\label{app:provenance}
As described in \Cref{sec:corpus}, the corpus was assembled from sources named in the seed study's early extraction sheets \citep{ward2025preprint}.
Here we describe how the estimates in those sheets were linked to corpus papers for the comparisons in \Cref{sec:provenance}, and check how review coverage and classification choices affect the main contrasts.

\subsection{Scope and inputs}

We use the data extraction sheets shared by the seed study authors.
The sheets are not redistributed.
They contain $346$ estimate rows and, after identifiers are resolved, $138$ distinct published sources.
The sheets are a working version and do not exactly match either the first preprint or the published article.
The first preprint reports $342$ estimates \citep{ward2025preprint}.
Five sources behind eight sheet rows do not appear in its references.
The published article \citep{ward2025epidemiology} reports a revised set of $302$ estimates from $154$ articles.
All counts below refer to the early sheets.

We group the parameters into seven categories: reproduction number (basic, effective, or controlled), serial interval, incubation period, latent period, infectious period (including effective infectious periods), dispersion parameter $k$, and fatality risk (case or infection).
These groups combine related quantities for describing their cited papers.

We call a review cited by an estimate a \emph{source review}.
An \emph{expanded source review} is one whose included studies are linked to corpus records, whether via corpus construction or a published study list.

\subsection{Matching cited sources}

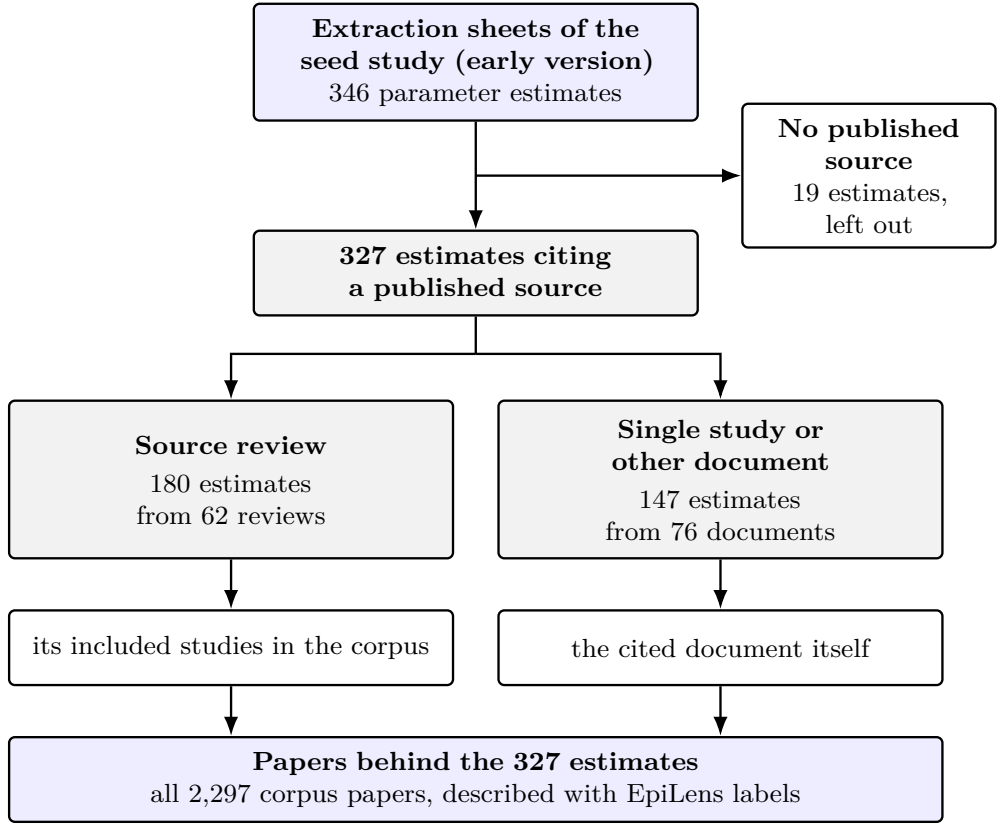
\begin{figure}[!htbp]
  \centering
  \resizebox{\columnwidth}{!}{%
\begin{tikzpicture}[font=\small,
  execute at begin node={\hyphenpenalty=10000\exhyphenpenalty=10000}]

\node[processbox, fill=blue!7, text width=5.0cm] (sheets) {%
  \textbf{Extraction sheets of the seed study (early version)}\\
  346 parameter estimates};
\node[processbox, text width=5.0cm, below=1.3cm of sheets] (cite) {%
  \textbf{327 estimates citing a published source}};

\coordinate (mid) at ($(sheets.south)!0.5!(cite.north)$);
\node[processbox, fill=white, text width=2.7cm, anchor=west] (none) at ($(mid)+(3.2,0)$) {%
  \textbf{No published source}\\19 estimates, left out};

\coordinate (split) at ($(cite.south)+(0,-0.45)$);
\node[processbox, text width=5.0cm, minimum height=1.9cm, anchor=north] (review) at ($(split)+(-2.95,-0.55)$) {%
  \textbf{Source review}\\[2pt]180 estimates\\from 62 reviews};
\node[processbox, text width=5.0cm, minimum height=1.9cm, anchor=north] (study) at ($(split)+(2.95,-0.55)$) {%
  \textbf{Single study or other document}\\[2pt]147 estimates\\from 76 documents};

\node[processbox, fill=white, text width=5.0cm, minimum height=0.9cm, below=0.6cm of review] (incl) {%
  its included studies in the corpus};
\node[processbox, fill=white, text width=5.0cm, minimum height=0.9cm, below=0.6cm of study] (rec) {%
  the cited document itself};

\node[processbox, fill=blue!7, text width=10.9cm, anchor=north] (behind) at ($(incl.south -| split)+(0,-0.6)$) {%
  \textbf{Papers behind the 327 estimates}\\
  all \numpaper{} corpus papers, described with \toolname{} labels};

\draw[flow] (sheets) -- (cite);
\draw[flow] (mid) -- (none);
\draw[thick] (cite.south) -- (split);
\draw[flow] (split) -| (review.north);
\draw[flow] (split) -| (study.north);
\draw[flow] (review) -- (incl);
\draw[flow] (study) -- (rec);
\draw[flow] (incl.south) -- (incl.south |- behind.north);
\draw[flow] (rec.south) -- (rec.south |- behind.north);
\end{tikzpicture}}
  \caption{\textbf{Linking parameter estimates to corpus papers.}
  The $327$ sheet rows citing a published source divide into reviews and single studies or other documents.
  A review principally links to its retrieved included studies in the corpus, while a directly cited document is itself analysed.
  The review and document counts are distinct sources after identifiers are resolved.}
  \label{fig:linkage}
\end{figure}

Nineteen sheet rows have no published source to link, mostly because they relate to the seed study's own incubation-period calculations.
We link the other $327$ to their cited sources, as illustrated in \Cref{fig:linkage}.
For a directly cited study or other document, we analyse that source itself.
A source review links to its included studies present in the corpus.

The corpus-building records group included studies under the source review's first author and year.
We use these identifiers to connect a sheet citation to the appropriate group and check the DOI for reviews sharing an author and year.
We match individual documents chiefly by DOI, with author and year or verified manual matching where needed.
Where an identifier is malformed, takes a different form, or identifies another document, we apply a declared replacement checked against Crossref \citep{hendricks2020crossref}.

\subsection{Associated paper sets and labels}

The corpus records links between its papers and the sources named in the sheets.
We use those links to define the paper set associated with each estimate row $r$, denoted $B_r$.
For a directly cited study or other document, $B_r$ contains that document.
For a source review, $B_r$ contains its included studies that are present in the corpus (\Cref{app:corpus_pipeline}).
These are studies the review identifies as evidence for its analysis, rather than every item in its bibliography.
The sheets cite the review for a particular estimate without naming the contributing studies.
For example, a review reporting both serial-interval and incubation-period estimates may use different subsets of its included studies for each.
Thus $B_r$ reflects the review's included evidence, which may be broader than the studies used for one value.
A paper can also be included in several reviews, even if it was retrieved for only one.
We retain every documented membership.
For each parameter, pathogen, or combination of the two, we take the union of the corresponding $B_r$ sets.
Each paper adds once to a group's paper count and label shares, regardless of how many estimates link to it.
The same paper may appear in several groups, so their paper counts are not additive.

\Cref{tab:underlying_papers} tabulates model-based and simulated-data shares for the paper sets plotted in \Cref{fig:provenance_parameters,fig:provenance_pathogens}.
It also lists the influenza A subtypes combined in the pathogen figure.
These percentages use the labels of the fully local workflow at $T{=}0$ over all associated corpus papers, so they are model outputs without sampling intervals.
The figures do not separate pathogens within each parameter.
\Cref{tab:paper_profile_comparisons} provides that breakdown for the comparisons discussed in \Cref{sec:provenance}.

\begin{table}[!htbp]
  \centering
  \small
  \setlength{\tabcolsep}{4pt}
  \renewcommand{\arraystretch}{0.85}
  \caption{\textbf{Labels of the papers behind the estimates, by parameter and by pathogen.}}
  \label{tab:underlying_papers}
  \begin{tabular}{l r r r}
    \toprule
     & \textbf{Papers} & \begin{tabular}[b]{@{}c@{}}\textbf{Model-based}\\\textbf{(\%)}\end{tabular} & \begin{tabular}[b]{@{}c@{}}\textbf{Simulated}\\\textbf{data (\%)}\end{tabular} \\
    \midrule
    \multicolumn{4}{l}{\emph{By parameter}} \\
    \quad Reproduction number & $629$ & $53$ & $31$ \\
    \quad Fatality risk & $1{,}549$ & $15$ & $8$ \\
    \quad Serial interval & $539$ & $45$ & $23$ \\
    \quad Incubation period & $845$ & $22$ & $11$ \\
    \quad Dispersion $k$ & $137$ & $53$ & $31$ \\
    \quad Infectious period & $186$ & $38$ & $30$ \\
    \quad Latent period & $140$ & $47$ & $37$ \\
    \midrule
    \multicolumn{4}{l}{\emph{By pathogen}} \\
    SARS-CoV-2 & $590$ & $36$ & $15$ \\
    \multicolumn{4}{l}{\textbf{Influenza}} \\
    \quad A/H1N1 & $109$ & $74$ & $45$ \\
    \quad A/H1N1pdm09 & $121$ & $66$ & $37$ \\
    \quad A/H2N2 & $127$ & $62$ & $36$ \\
    \quad A/H3N2 & $121$ & $66$ & $39$ \\
    \quad A/H1N1 and A/H3N2 & $1$ & --- & --- \\
    \quad A/H5N1 & $146$ & $55$ & $31$ \\
    \multicolumn{4}{l}{\textbf{SARS and MERS}} \\
    \quad MERS-CoV & $90$ & $40$ & $16$ \\
    \quad SARS-CoV-1 & $64$ & $33$ & $23$ \\
    \multicolumn{4}{l}{\textbf{Ebola and Marburg}} \\
    \quad Ebola virus & $262$ & $26$ & $18$ \\
    \quad Marburg virus & $59$ & $8$ & $3$ \\
    \multicolumn{4}{l}{\textbf{Bunyavirales}} \\
    \quad CCHF virus & $497$ & $3$ & $2$ \\
    \quad Lassa virus & $125$ & $6$ & $6$ \\
    \quad RVF virus & $214$ & $3$ & $0$ \\
    \addlinespace
    Mpox virus & $153$ & $5$ & $3$ \\
    Nipah virus & $112$ & $1$ & $2$ \\
    Zika virus & $146$ & $33$ & $16$ \\
    \midrule
    All & $\numpaper$ & $21$ & $11$ \\
    \bottomrule
  \end{tabular}
  \footnotetext{Each row counts distinct corpus records associated with estimates of one parameter or pathogen. A record associated with several parameters or pathogens counts for each, so the rows add up to more than All. Model-based and Simulated data: percentage of these papers to which \toolname{} assigns the label with \texttt{qwen2.5:32b-instruct} on local hardware (model outputs, not estimates). Model-based means paper type \texttt{INFERENCE} or \texttt{FORECAST}, and simulated data means data type \texttt{SYNTHETIC}. Shares are not shown for rows with fewer than 10 papers. \Cref{sec:provenance} compares these labels with the manual labels for the comparisons it draws. Parameters are ordered as in \Cref{fig:provenance}. Bold headings name the pathogen groups in the early extraction sheets shared by the authors of \citet{ward2025preprint} that cover several pathogens. A/H1N1: estimates not specific to the 2009 pandemic virus. A/H1N1pdm09: the 2009 pandemic virus. A/H1N1 and A/H3N2: estimates reported for both subtypes together. CCHF: Crimean--Congo haemorrhagic fever. RVF: Rift Valley fever. Fatality risk: case or infection fatality risk. The influenza A subtype sets overlap heavily: reviews covering several influenza pandemics link largely the same model-based papers to every subtype, whereas empirical papers tend to be specific to one subtype. Each subtype set therefore has a higher model-based share than the merged set in \Cref{fig:provenance_pathogens}. Unindented rows are pathogens not grouped in the extraction sheets.}
\end{table}

\begin{table}[!htbp]
  \centering
  \small
  \setlength{\tabcolsep}{4pt}
  \caption{\textbf{Model-based shares in the pathogen--parameter comparisons.}}
  \label{tab:paper_profile_comparisons}
  \begin{tabular}{l l r r}
    \toprule
    \textbf{Pathogen} & \textbf{Parameter} & \textbf{Papers} & \begin{tabular}[b]{@{}c@{}}\textbf{Model-based}\\\textbf{(\%)}\end{tabular} \\
    \midrule
    \multicolumn{4}{l}{\emph{Reproduction number and fatality risk}} \\
    SARS-CoV-2 & Reproduction number & $204$ & $57$ \\
     & Fatality risk & $335$ & $41$ \\
    MERS-CoV & Reproduction number & $55$ & $36$ \\
     & Fatality risk & $67$ & $30$ \\
    Ebola virus & Reproduction number & $117$ & $50$ \\
     & Fatality risk & $237$ & $23$ \\
    Lassa virus & Reproduction number & $22$ & $23$ \\
     & Fatality risk & $124$ & $6$ \\
    Mpox virus & Reproduction number & $61$ & $5$ \\
     & Fatality risk & $55$ & $4$ \\
    \midrule
    \multicolumn{4}{l}{\emph{Other fatality-risk comparisons}} \\
    CCHF virus & Fatality risk & $497$ & $3$ \\
    Nipah virus & Fatality risk & $111$ & $1$ \\
    RVF virus & Fatality risk & $185$ & $3$ \\
    \midrule
    \multicolumn{4}{l}{\emph{Incubation period}} \\
    SARS-CoV-2 & Incubation period & $334$ & $34$ \\
    Mpox virus & Incubation period & $115$ & $1$ \\
    \bottomrule
  \end{tabular}
  \footnotetext{Each row counts distinct corpus papers associated with estimates of the named parameter for that pathogen. A paper can appear in several rows. Model-based is the share labelled \texttt{INFERENCE} or \texttt{FORECAST} by \toolname{} with \texttt{qwen2.5:32b-instruct} on local hardware at $T{=}0$ (model outputs, not parameter estimates). Rows show the comparisons discussed in \Cref{sec:provenance,app:provenance}. The full breakdown is in the reproduction package \citep{epilens_reproducibility}. CCHF: Crimean--Congo haemorrhagic fever. RVF: Rift Valley fever.}
\end{table}

\FloatBarrier

\subsection{Manual-label check}

The shares reported for all associated papers in \Cref{sec:provenance} depend on model-assigned labels.
We therefore check whether the contrast also appears under manual labels among the associated papers in the $214$-paper evaluation sample.
The manual labels show the same ordering, with a $38$-percentage-point difference in model-based shares.
We examine how this difference varies across the cited sources represented in that sample.

The paired bootstrap in \Cref{app:baselines:statistics} resamples individual papers to assess differences in classifier accuracy.
Here, several labelled papers can come from one cited review, so we resample source reviews and directly cited documents as groups.
This asks whether the manual-label contrast depends heavily on a few cited sources.

The resampling pool contains $45$ sources represented by at least one labelled paper behind either parameter.
In each of $4{,}000$ draws, we select $45$ sources with replacement, keep each selected source's labelled papers together, and recalculate the difference between the two manual-label shares.
A source shared by both parameters contributes to both shares in the same draw.
Within each parameter set, a paper linked to $m$ sources receives weight $1/m$ through each source, so that it counts once in the observed share.
The central $95\%$ of resampled differences is $19$--$57$ percentage points, preserving the direction under this source resampling.
This check concerns the manually labelled subset: the papers labelled within each review stay fixed, and the interval does not estimate what manual labels would show across the full associated paper sets.

\subsection{Sensitivity of the paper profiles}

We now check whether the model-assigned reproduction-number--fatality-risk contrast across all associated papers changes with review coverage or the removal of individual source reviews, and how the local labels compare with the manual ones.

\paragraph{Review coverage.}
Incomplete matching of a review's included studies could change the paper profiles.
We linked corpus papers to $62$ cited reviews, of which $60$ report how many studies they included.
The other two are an analytic perspective and a narrative review without a defined set of included studies.
Counting a paper once for each review that includes it, corpus records match $2{,}707$ of $4{,}467$ reported inclusions ($61\%$).
These are shares of reported review inclusions, not shares of the corpus.
Matching varies from $14$ to $100\%$ across reviews and is $47\%$ for those associated with reproduction numbers, against $56\%$ for fatality risks.

Removing reviews below the median matching rate, while retaining directly cited documents and reviews without reported totals, changes the model-based difference from $38$ to $50$ percentage points.
Hence, the direction persists.

\paragraph{Label accuracy and individual reviews.}
On the $214$ labelled papers, the local labels identify model-based papers with sensitivity $0.65$ and specificity $0.97$, and simulated-data papers with sensitivity $0.68$ and specificity $0.98$; they label $70\%$ of papers empirical, against $59\%$ in the manual labels.
The model-based and simulated-data shares across all associated papers are therefore likely underestimates, and the comparisons in \Cref{sec:provenance} rest on the direction of the differences rather than on their levels.

Omitting each of the $42$ source reviews behind either parameter in turn leaves the model-based difference at $34$--$44$ percentage points.
The model-based ordering therefore holds after the removal of any one review.

\paragraph{Pathogens within a parameter.}
Each row of \Cref{fig:provenance_pathogens} pools the papers behind all of a pathogen's estimates, so differences between pathogens partly reflect which parameters the seed study estimated for each.
\Cref{tab:paper_profile_comparisons} therefore also compares pathogens within a single parameter.
Among fatality-risk papers, $41\%$ of those linked to SARS-CoV-2 and $30\%$ of those linked to MERS-CoV are labelled model-based, against $1$--$23\%$ for CCHF, Ebola, Lassa, Nipah, and RVF.
Among incubation-period papers, $61\%$ of the $31$ manually labelled papers behind SARS-CoV-2 estimates are model-based, against $9\%$ of the $11$ behind mpox estimates; in \toolname{}'s outputs, $34\%$ of $334$ against $1\%$ of $115$.
The share is at least $11\%$ in each of seven SARS-CoV-2 source reviews and at most $4\%$ in each of four mpox source reviews.
Many of these papers also lie behind estimates of other parameters for the same pathogen.
Excluding those shared with reproduction-number or fatality-risk estimates leaves $22\%$ of $201$ against $0\%$ of $70$, and keeping only papers behind no other parameter leaves $7\%$ of $58$ against $0\%$ of $69$.
The ordering persists, but most of the contrast comes from papers shared with other parameters, so we do not interpret it further.

\end{appendices}

\FloatBarrier

\bibliography{references}

\end{document}